\documentclass[journal]{IEEEtran}
\usepackage{amssymb}

\usepackage{pifont,calc}
\usepackage{cite}

\usepackage{graphicx}
\ifCLASSINFOpdf
   \usepackage[pdftex]{graphicx}
\else
 \fi
\usepackage[cmex10]{amsmath}
\usepackage{array}

\usepackage{cases} 
\usepackage{bm} 
\usepackage[caption=false]{subfig}
\usepackage{caption,setspace}

\usepackage{mathrsfs}
\usepackage{url}
\usepackage{algorithm}
\usepackage{algpseudocode}

\usepackage{arydshln}
\usepackage{multirow}
\usepackage{multicol}
\makeatletter

\newcommand{\Rmnum}[1]{\expandafter\@slowromancap\romannumeral #1@}
\makeatother

\ifCLASSINFOpdf
\else
\fi
\begin{document}
%
\title{Two Efficient Ridge Solutions  for the Incremental Broad Learning System on Added Inputs}
%
%
%

\author{Hufei~Zhu
\thanks{H. Zhu is with Faculty of Intelligent Manufacturing, Wuyi University, Jiangmen 529020, China (e-mail:
zhuhufei@wyu.edu.cn).}}

%
%

\markboth{Journal of \LaTeX\ Class Files,~Vol.~14, No.~8, August~2015}%
{Shell \MakeLowercase{\textit{et al.}}: Bare Demo of IEEEtran.cls for IEEE Journals}
%

%
%
%
%
%



\maketitle

\begin{abstract}
To improve the existing
 broad learning system (BLS) for new added inputs,
this paper proposes
 the recursive and square-root BLS algorithms
 that
   utilize the inverse and inverse Cholesky factor of
  the Hermitian matrix in the ridge inverse, respectively,
  to update the ridge solution for the output
weights.
 The recursive BLS  updates the inverse
  by the matrix inversion lemma,
     while the square-root BLS  updates the
     upper-triangular
       inverse Cholesky factor
 by multiplying
 it with an upper-triangular intermediate matrix.
When the added $p$ training samples are more than the total
$k$ nodes in the network, i.e., $p>k$,
   the inverse of a sum of matrices is applied to
   take  a smaller  matrix inversion
   or inverse Cholesky factorization.
 For the distributed BLS with data-parallelism, we introduce
  the parallel implementation of the square-root BLS, which is deduced from the parallel implementation of the inverse Cholesky factorization.


The existing BLS based on  the generalized
inverse with the ridge regression
 assumes the ridge parameter $\lambda \to 0$ in the ridge inverse.
When $\lambda \to 0$ is not satisfied,
the   numerical experiments
show that
both the proposed ridge solutions
improve the testing accuracy of the existing BLS, and the improvement becomes more significant as  $\lambda$ is bigger.
For example,  the proposed two ridge solutions and the existing BLS achieve the maximum testing accuracies of
 $90.34\%$ and $90.24\%$ respectively   when $\lambda = {{10}^{-3}}$
 on the NYU object recognition
benchmark  dataset.
  On the other hand, compared to the existing BLS,
  both the proposed BLS algorithms  theoretically require less complexities,
  and are significantly faster in the simulations
on  the Modified National Institute of Standards and Technology
dataset.
The speedups  in  total
training time of  the
recursive and square-root BLS algorithms
  over the existing BLS are $4.41$
and $6.92$ respectively when $p>k$, and are $2.80$
and $1.59$ respectively when $p<k$.

Compared to the
recursive
 BLS,
   the
    square-root BLS requires less complexities when
    $p > 0.82k$, and requires more complexities when
     $p < 0.82k$.
  If nodes are inserted after each increment of inputs,
both the proposed BLS algorithms are  applied with
the efficient ridge solution based on the inverse
Cholesky factor for the BLS on
added nodes,  to obtain the complete ridge solution.
Then the
 recursive BLS always require more complexities than the
  square-root BLS,
  since it
requires the  extra computations
to get the Cholesky factor and multiply the Cholesky factor
with its transpose.  When $\lambda$ is small in the simulations,
the numerical errors caused by
those  extra computations
 reduce
the testing accuracy
 or even make the
 recursive BLS
unworkable.
  On the contrary, the
 square-root BLS does not need any extra computations and basically achieves the
  testing accuracy of the direct ridge solution, since it is based on
  the Cholesky factor just like the efficient  ridge solution for the  BLS on added nodes.

\end{abstract}


%
\IEEEpeerreviewmaketitle

\section{Introduction}

Single layer feedforward neural networks (SLFN) with the universal approximation capability have been
widely applied to
solve the  classification and
regression problems~\cite{BL_Ref_18,BL_Ref_19,BL_Ref_20}.
 SLFNs can utilize
traditional  Gradient-descent-based learning algorithms~\cite{BL_Ref_22,BL_Ref_23}.
However, those Gradient-descent-based algorithms
suffer from the time-consuming training process
and the local minimum trap,
while
their generalization performance is sensitive to the
training parameters,  e.g., learning rate.
Then the random vector functional-link neural network (RVFLNN)
was proposed~\cite{BL_Ref_19}
to eliminate the drawback of long training process,
which trains only the   output weights, and generates the
 input weights and biases randomly.
RVFLNN
 offers the generalization capability in function
approximation~\cite{BL_Ref_20}, and  has been proven to be a
universal approximation for continuous functions on
compact sets.

Based on the RVFLNN model,
a dynamic step-wise updating
algorithm was proposed in \cite{27_ref_BL_trans_paper} to
 model  time-variety data with moderate size.
When a new input is encountered or
the increment of a new node is required,
the dynamic algorithm in \cite{27_ref_BL_trans_paper} only computes the
pseudoinverse of that added input or node,
to update
the output weights easily.
The scheme in \cite{27_ref_BL_trans_paper}
was improved
 into
 Broad Learning
System (BLS) in \cite{BL_trans_paper},
to deal with time-variety big data with
high dimension.
Then in \cite{BL_trans_paperApproximate},
a mathematical proof of the universal approximation capability of BLS
is provided, and several BLS variants were discussed, which include cascade, recurrent, and broad-deep combination structures.


In BLS~\cite{BL_trans_paper,BL_trans_paperApproximate},   the previous scheme~\cite{27_ref_BL_trans_paper} is improved in three aspects.
Firstly, BLS
transforms
 the input data
  into
  the feature nodes
  to reduce the
data dimensions.
 Secondly, BLS can update
the output weights easily
 for any number of new added nodes or
inputs, since it only requires one iteration to compute the
pseudoinverse of those added nodes or inputs.
 Lastly,  to achieve a better generalization performance,
  BLS computes the output weights by the generalized inverse with the ridge regression, which
 assumes the ridge parameter $\lambda \to 0$ in the ridge inverse~\cite{best_ridge_inv_paper213}
  to
  approximate
   the generalized inverse.

To improve the original BLS on added nodes~\cite{BL_trans_paper}, the efficient generalized inverse and ridge solutions~\cite{best_ridge_inv_paper213} were proposed
in \cite{CholBLSdec2020} and \cite{CholBLSnodesSubmitted}, respectively,
which are both based on the Cholesky factor, while the ridge solution based on the ridge inverse was also proposed
in \cite{CholBLSnodesSubmitted}.
To improve the original BLS on added inputs~\cite{BL_trans_paper},  an efficient implementation was proposed  in \cite{my_ppaapper1_on_BL}
to accelerate a step in the generalized inverse of a  partitioned matrix. Specifically,
the inverse of a sum of matrices~\cite{InverseSumofMatrix8312} was utilized in \cite{my_ppaapper1_on_BL}
when the added inputs are more than the total nodes in the network.

 In this paper, we propose two efficient BLS algorithms for added inputs,
  which compute the ridge solution for  the output
weights.
   Then the assumption of  $\lambda \to 0$ in the original BLS~\cite{BL_trans_paper}
   is no longer required, and $\lambda$ can be any positive real number.
The proposed BLS algorithms compute
 the ridge solution
  from the inverse or inverse Cholesky factor of
  the Hermitian matrix in the ridge inverse,
  to
 avoid spending more complexity
 to compute the bigger
  ridge inverse.
In the case of
  more added inputs than the total nodes,
  the inverse of a sum of matrices~\cite{InverseSumofMatrix8312} is also  applied to
  accelerate several steps in the proposed BLS algorithms, as in \cite{my_ppaapper1_on_BL}.

When big data with
high dimension  is processed,  the training processing may exceed the capacity of a single computational node. Accordingly, usually it is required  to distribute computing tasks across multiple computational nodes, that are also called as workers~\cite{SurveyOnDistributedML}.  Specifically,
a distributed implementation is usually necessary when data is inherently distributed or too big to store on a single worker.
It is necessary to choose and implement the algorithms to enable parallel computation in distributed systems.
In this paper, we focus on
 data-parallelism~\cite{SurveyOnDistributedML}, where
  training samples
 are partitioned into multiple workers, and nearly the same algorithms are applied  to different groups of training samples in all workers.
 We will develop the parallel implementation of   the proposed  ridge solution
based on inverse Cholesky factor for the  distributed
BLS with data-parallelism.


  This paper is organized as follows. Section \Rmnum{2} introduces the existing incremental BLS on added inputs based on the generalized inverse
  with the ridge regression.
 In Section \Rmnum{3}, we propose two efficient ridge solutions for the BLS on added inputs.
Then in Section \Rmnum{4},  we compare the expected computational complexities of the
existing BLS and the proposed two ridge solutions,
and evaluate them by numerical experiments.
Finally,
conclusions are given in Section \Rmnum{5}.

%
%


 \section{Existing Incremental BLS on Added Inputs Based on Generalized Inverse Solution}

%
%
%
%
%
%
%
%
%
%
%

In RVFLNN,
the input data  ${{\mathbf{X}}}$  forms the enhancement components
by
$ \xi ({{\mathbf{X}}}{{\mathbf{W}}_{{{h}}}}+{{\mathbf{\beta }}_{{{h}}}})$,
where ${{\mathbf{W}}_{{{h}}}}$ and
        ${{\mathbf{\beta }}_{{{h}}}}$  are random,
        and $\xi$  is the activation function.
Then  the  output
\begin{equation}\label{Y2AW65897}
{\mathbf{\hat{Y}}}={{\mathbf{A}}}{{\mathbf{W}}},
 \end{equation}
 where ${{\mathbf{W}}}$ denotes the output weights, and the expanded input matrix
 ${{\mathbf{A}}}=\left[ {{\mathbf{X}}}| \xi ({{\mathbf{X}}}{{\mathbf{W}}_{{{h}}}}+{{\mathbf{\beta }}_{{{h}}}}) \right]$.  
 The least-square solution~\cite{27_ref_BL_trans_paper} of (\ref{Y2AW65897}) is
  the generalized inverse solution~\cite{best_ridge_inv_paper213}
\begin{equation}\label{W2AinvY989565}
{{\mathbf{W}}}=\mathbf{A}^{+} \mathbf{Y},
\end{equation}
where $\mathbf{Y}$ denotes  the labels, and the generalized inverse
 \begin{equation}\label{Ainv2AtAinvAt9096}
{{\bf{A}}^{+ }}={{(\mathbf{A}^{T}\mathbf{A})}^{-1}}\mathbf{A}^{T}.
 \end{equation}

\subsection{Ridge Regression Approximation of the Generalized Inverse}

The generalized inverse solution 
(\ref{W2AinvY989565})
is aimed to minimize the training errors. But usually it can not
achieve the minimum generalization errors, especially for ill-conditioned problems.
To achieve a better generalization performance,
instead of the generalized inverse solution (\ref{W2AinvY989565}),
an alternative
solution can be utilized, i.e.,
 the ridge solution~\cite{best_ridge_inv_paper213}
\begin{equation}\label{xWbarMN2AbarYYa1341}
{\mathbf{\tilde W}}={{\bf{A}}^{\dagger }}{{\mathbf{Y}}},
 \end{equation}
where
 the ridge inverse~\cite{best_ridge_inv_paper213}
\begin{equation}\label{xAmnAmnTAmnIAmnT231413}
{{\bf{A}}^{\dagger }}={{\left( {{\bf{A}}^{T}}{\bf{A}}+\lambda \mathbf{I} \right)}^{-1}}{{\bf{A}}^{T}}.
\end{equation}
The ridge inverse (\ref{xAmnAmnTAmnIAmnT231413}) degenerates~\cite[Eq. (3)]{BL_trans_paper} into the
 generalized inverse when the ridge parameter $\lambda \to 0$,  i.e.,
 \begin{equation}\label{AinvLimNumda0AAiA1221}
\underset{\lambda \to 0}{\mathop{\lim }}\,\mathbf{A}^{\dagger}=\underset{\lambda \to 0}{\mathop{\lim }}\,{{(\mathbf{A}_{{}}^{T}\mathbf{A}+\lambda \mathbf{I})}^{-1}}\mathbf{A}_{{}}^{T}
=\mathbf{A}_{{}}^{+ },
\end{equation}
which is the ridge regression approximation of the generalized
inverse.
In \cite{BL_trans_paper},
$\mathbf{A}_{{}}^{+ }$ in the output weights
 (\ref{W2AinvY989565})
is computed from (\ref{AinvLimNumda0AAiA1221}) instead of (\ref{Ainv2AtAinvAt9096}),
to improve the generalization performance.


\subsection{Broad Learning Model}

The BLS transfers  the original input data
 $\mathbf{X}$   into the mapped features in the feature nodes,
 and then enhances the feature nodes as the enhancement
nodes. The connections of all the feature
and enhancement nodes are fed into the output finally.

In the BLS,  the input data $\mathbf{X}$
 is projected by
\begin{equation}\label{Z2PhyXWb985498}
{{\mathbf{Z}}_{i}}=\phi (\mathbf{X}{{\mathbf{W}}_{{{e}_{i}}}}+{{\mathbf{\beta }}_{{{e}_{i}}}}),
\end{equation}
to become the $i$-th group of mapped features ${{\mathbf{Z}}_{i}}$,
where  the  weights ${{\mathbf{W}}_{{{e}_{i}}}}$ and the biases ${{\mathbf{\beta }}_{{{e}_{i}}}}$ are
  randomly generated and then fine-tuned by applying the linear inverse problem~\cite{BL_trans_paper}.  All the $n$ groups of mapped features are concatenated into
\begin{equation}\label{Zi2z1zi988689}{{\mathbf{Z}}^{n}}\equiv \left[ \begin{matrix}
   {{\mathbf{Z}}_{1}} & \cdots  & {{\mathbf{Z}}_{n}}  \\
\end{matrix} \right].
\end{equation}
Then all the mapped features ${{\mathbf{Z}}^{n}}$ are  enhanced to become the
$j$-th group of enhancement nodes ${{\mathbf{H}}_{j}}$,
 by
\begin{equation}\label{HjipsenZjWbelta09885}{{\mathbf{H}}_{j}}=\xi ({{\mathbf{Z}}^{n}}{{\mathbf{W}}_{{{h}_{j}}}}+{{\mathbf{\beta }}_{{{h}_{j}}}})
\end{equation}
 where
    ${{\mathbf{W}}_{{{h}_{j}}}}$ and ${{\mathbf{\beta }}_{{{h}_{j}}}}$   are random.
 All the  $m$ groups of enhancement nodes are concatenated into
\begin{equation}\label{Hj2H1Hj9859348}{{\mathbf{H}}^{m}}\equiv \left[ {{\mathbf{H}}_{1}},\cdots ,{{\mathbf{H}}_{m}} \right].
\end{equation}
Finally,
all the feature  and  enhancement nodes are fed into the output by
\begin{equation}\label{Y2ZiHjWj948934}
\mathbf{\hat{Y}}=\left[ {{\mathbf{Z}}^{n}}|{{\mathbf{H}}^{m}} \right]{{\mathbf{W}}_n^{m}}={\bf{A}}_n^m{{\mathbf{W}}_n^{m}},
\end{equation}
where
the expanded input matrix
\begin{equation}\label{Anm2HZ438015}
{\bf{A}}_n^m=\left[ {{\mathbf{Z}}^{n}}|{{\mathbf{H}}^{m}} \right],
\end{equation}
and
 the desired connection weights ${{\mathbf{W}}_n^{m}}$
are computed by (\ref{W2AinvY989565})
from  $({\bf{A}}_n^m)^{+}$,
the generalized inverse with the ridge regression.


\subsection{Incremental Learning for Added Inputs}

The BLS includes the incremental learning for the additional input training samples.
When  encountering new input samples with the corresponding output labels,
the modeled BLS can be remodeled in an
incremental way without a complete retraining process.
It updates the output weights incrementally, without retraining the whole network from the beginning.
%
%
%
%
%


Assume that the expanded input matrix ${\bf{A}}_n^m$
is $l \times k$, where $l$ is the number of training samples, and $k$ is the total number of
feature and enhancement nodes.
Then
we can write ${\bf{A}}_n^m$ and  the corresponding input data
 ${{\mathbf{X}}}$
 as  $\mathbf{A}_{{\bar{l}}}$ and $\mathbf{X}_{{\bar{l}}}$, respectively.
 Similarly,  the additional input
 data can be denoted  as ${{\mathbf{X}}_{ \bar p}}$
with $p$ samples~\footnote{Notice that in this paper,  any subscript
indicating the number of training samples is overlined,
   e.g.,  $\bar{l}$ and $\bar{p}$.}.
In BLS,  ${{\mathbf{X}}_{ \bar p}}$ is projected by (\ref{Z2PhyXWb985498})
 to get  the additional samples for the $i$-th group of feature nodes,
 which are  concatenated into
  \begin{equation}\label{ZxXa94319243}
 \mathbf{Z}_{ \bar p}^{n}=\left[ \phi ({{\mathbf{X}}_{ \bar p}}{{\mathbf{W}}_{{{e}_{1}}}}+{{\mathbf{\beta }}_{{{e}_{1}}}}),\cdots ,\phi ({{\mathbf{X}}_{ \bar p}}{{\mathbf{W}}_{{{e}_{n}}}}+{{\mathbf{\beta }}_{{{e}_{n}}}}) \right]
 \end{equation}
 by (\ref{Zi2z1zi988689}).
 $\mathbf{Z}_{ \bar p}^{n}$ is
  enhanced by (\ref{HjipsenZjWbelta09885}) and then concatenated
  by (\ref{Hj2H1Hj9859348}).
  Accordingly,  the additional samples for the expanded input matrix
  can be written as ${\bf{A}}_{ \bar p} \in {\Re ^{p \times k}}$ satisfying
  \begin{multline}\label{PaperEqu21AxZ23141}\mathbf{A}_{ \bar p}^{{}}=  \\
\left[\mathbf{Z}_{ \bar p}^{n} | \xi (\mathbf{Z}_{ \bar p}^{n}{{\mathbf{W}}_{{{h}_{1}}}}+{{\mathbf{\beta }}_{{{h}_{1}}}}),\cdots ,\xi (\mathbf{Z}_{ \bar p}^{n}{{\mathbf{W}}_{{{h}_{m}}}}+{{\mathbf{\beta }}_{{{h}_{m}}}})\right],
\end{multline}
which is utilized to
update
 $\mathbf{A}_{{\bar{l}}}$  into
\begin{equation}\label{AxInputIncrease31232}
{\bf{A}}_{\bar l + \bar p}= {\left[ {\begin{array}{*{20}{c}}
{{\bf{A}}_{\bar l}^T}&{{\bf{\bar A}}_{\bar p}^T}
\end{array}} \right]^T}.
\end{equation}

In the stepwise updating algorithm in \cite{BL_trans_paper},
 the generalized inverse of  $\mathbf{A}_{{\bar{l}}}$ is updated into that of  ${\bf{A}}_{\bar l + \bar p}$
  by
\begin{equation}\label{xAbar2AbarBtDtBt4132OLD984}
{{\bf{A}}_{\bar l + \bar p}^{+ }}=\left[ \begin{matrix}
   {{\bf{A}}_{\bar l}^{+ }}-{{\mathbf{B}}}{{\mathbf{D}}^{T}} & {{\mathbf{B}}}  \\
\end{matrix} \right],
\end{equation}
where \begin{equation}\label{DtAxAmnInv324141OLD984}
{{\mathbf{D}}^{T}}=\mathbf{A}_{ \bar p}^{{}}{{\bf{A}}_{\bar l}^{+ }},
\end{equation}
 \begin{equation}\label{CAxDtA413124OLD984}
 \mathbf{{C}}=\mathbf{A}_{ \bar p}^{{T}}-{\bf{A}}_{\bar l}^T {{\mathbf{D}}},
 \end{equation}
 and
 \begin{subequations}{\label{BtWaveBest23134OLD984}}
 \begin{numcases}
{  {{\bf{B}}}= }
({{\bf{C}}^+})^T  \   \quad \quad \quad \quad \quad \quad if \ {{\bf{C}} \ne {\bf{0}}}   &  \label{} \\
{  {{\bf{A}}_{\bar l}^{+ }} {{\bf{D}}} {{({\bf{I}} + {{\bf{D}}^T}{\bf{D}})}^{ - 1}} \quad if \ {\bf{C}} = {\bf{0}}}.  &  \label{B2DtDwhenC0arr3}
\end{numcases}
\end{subequations}
Then the output weights $\mathbf{W}_{\bar l}$ is updated into ${\mathbf{W}}_{\bar l + \bar p}$ by
\begin{equation}\label{xWnmBestCompute13434OLD984}
{\mathbf{W}}_{\bar l + \bar p}=\mathbf{W}_{\bar l}+{{\mathbf{B}}}({{\mathbf{Y}}_{ \bar p}}-\mathbf{A}_{ \bar p}^{{}}\mathbf{W}_{\bar l}),
\end{equation}
which is the generalized inverse solution.

Since
 it is  impossible to observe
 any input matrix
 that is rank deficient in practice~\cite[pp. 64]{27_ref_BL_trans_paper},
 we assume the expanded input matrix $\mathbf{A}_{{\bar{l}}}$ is of full rank in this paper. Moreover,
  we focus on the usual case where
  the $l \times k$ matrix
  $\mathbf{A}_{{\bar{l}}}$
    satisfies
      \begin{equation}\label{lbiggerThaNk39w32}
l > k,
 \end{equation}
i.e.,
 the   training samples are more
than  the total
 nodes, as in~\footnote{Equation (3)  in \cite{BL_trans_paper}, i.e., (\ref{AinvLimNumda0AAiA1221}), is
  equal to the left inverse ${{(\mathbf{A}^{T}\mathbf{A})}^{-1}}\mathbf{A}^{T}$,
  which assumes $l \ge k$ for
   ${\bf{A}}\in {\Re ^{l \times k}}$ with full rank. Moreover, usually $l \ge 2k$ is satisfied
   in  Tables \Rmnum{5} and \Rmnum{6} of \cite{BL_trans_paper} for the increment of input pattern.}  \cite{BL_trans_paper, my_ppaapper1_on_BL}.
Then ${\bf{C}} ={\bf{0}}$  can be concluded~\cite[pp. 64]{27_ref_BL_trans_paper}
for ${\bf{C}}$ computed  by  (\ref{CAxDtA413124OLD984})
from $\mathbf{A}_{{\bar{l}}}\in {\Re ^{l \times k}}$ with $l > k$.

%
For the usual case with  ${\bf{C}} ={\bf{0}}$,
  the
   complexity of the BLS on added inputs was reduced in \cite{my_ppaapper1_on_BL} by
modifying  (\ref{BtWaveBest23134OLD984})
      into
 \begin{subequations}{\label{B_Matrix_defSimple3141}}
 \begin{numcases}
{  {{\bf{B}}} = }
{ {{\bf{C}}^ + } \quad \quad \quad \quad \quad \quad \quad \quad \  \,  if \ {\bf{C}} \ne {\bf{0}}}   &  \label{B_Matrix_def2bSimplCne0} \\
{\bf{\bar D}}{{({\bf{I}} + {\bf{A}}_{ \bar p}{\bf{\bar D}})}^{ - 1}} \quad \quad \quad  \ if\  {\bf{C}} ={\bf{0}} \And  p \le k  &   \label{B_Matrix_def2bSimplebbb}\\
{{({\bf{I}} + {\bf{\bar D}} {\bf{A}}_{ \bar p})}^{ - 1}}{\bf{\bar D}} \quad \quad \quad  \ if\ {\bf{C}} ={\bf{0}} \And  p \ge k,   &  \label{B_Matrix_def2bSimpleaaa}
\end{numcases}
\end{subequations}
where
$p$ and $k$  are the size of ${\bf{A}}_{ \bar p} \in {\Re ^{p \times k}}$,
and
\begin{equation}\label{D_bar_def_432ge32}
{\bf{\bar D}} ={{\bf{A}}_{\bar l}^{+ }} {{\bf{D}}}.
\end{equation}
Obviously,
(\ref{B_Matrix_def2bSimplebbb}) for the case of $p \le k$
contains only
  minor changes to
   (\ref{B2DtDwhenC0arr3}).
   Moreover,
 the inverse of a sum of matrices~\cite{InverseSumofMatrix8312} is utilized to deduce
(\ref{B_Matrix_def2bSimpleaaa}) for the case of $ p \ge k$ with the $k \times k$ matrix inverse, which
is usually more efficient than (\ref{B2DtDwhenC0arr3}) with the  $p \times p$ matrix inverse.





\subsection{Construction Model and Learning Procedure of BLS}


In \textbf{Algorithms 1} and \textbf{2},
we   summarize
the construction model and learning procedure of
the existing BLS algorithm on added inputs, which includes the improvement proposed
in \cite{my_ppaapper1_on_BL}.
\textbf{Algorithm 1}  lists the construction model and the procedure to
compute  the initial expanded input matrix $\mathbf{A}_{{\bar{l}}}$,
while \textbf{Algorithm 2}  lists the procedure to compute the
output weights and the incremental learning for added inputs.

\begin{algorithm}
\caption{:~\bf The Broad Learning Algorithm:  Computation of the Initial Expanded Input Matrix}
\begin{algorithmic}[1]
\Require Training samples $\mathbf{X}_{{\bar{l}}}$
\Ensure  Expanded input matrix $\mathbf{A}_{{\bar{l}}}={\bf{A}}_n^m$
\For{$i=1:n$}
\State Fine-tune random ${{\mathbf{W}}_{{{e}_{i}}}}$ and  ${{\bm{\beta }}_{{{e}_{i}}}}$;
\State Compute ${{\mathbf{ Z}}_{i}}=\phi (\mathbf{X}_{{\bar{l}}} {{\mathbf{W}}_{{{e}_{i}}}}+{{\bm{\beta }}_{{{e}_{i}}}})$;
\EndFor
\State Concatenate the feature nodes into ${{\mathbf{ Z}}^{n}}\equiv \left[ \begin{matrix}
   {{\mathbf{ Z}}_{1}} & \cdots  & {{\mathbf{ Z}}_n}  \\
\end{matrix} \right]$;
\For{$j=1:m$}
\State Random   ${{\mathbf{W}}_{{{h}_{j}}}}$ and
    ${{\bm{\beta }}_{{{h}_{j}}}}$;
\State Compute ${{\mathbf{ H}}_{j}}=\xi ({{\mathbf{ Z}}^{n}}{{\mathbf{W}}_{{{h}_{j}}}}+{{\bm{\beta }}_{{{h}_{j}}}})$;
\EndFor
\State   Set the enhancement nodes group 
${{\mathbf{ H}}^{m}}\equiv \left[ {{\mathbf{ H}}_{1}},\cdots,{{\mathbf{ H}}_{m}} \right]$;
\State Set ${{\mathbf{ A}}_n^{m}}=\left[ {{\mathbf{ Z}}^{n}}|{{\mathbf{ H}}^{m}} \right]$, and write ${{\mathbf{ A}}_n^{m}}$ as $\mathbf{A}_{{\bar{l}}}$;
\end{algorithmic}
\end{algorithm}

\begin{algorithm}
\caption{:~\bf The Existing Broad Learning Algorithm:  Computation of Output Weights and Increment of  Inputs}
\begin{algorithmic}[1]
\Require Expanded input matrix $\mathbf{A}_{{\bar{l}}}$, labels $\mathbf{Y}_{{\bar{l}}}$, added inputs ${{\mathbf{X}}_{ \bar p}}$ and  labels ${{\mathbf{Y}}_{ \bar p}}$
\Ensure Output weights $\mathbf{W}$
\State Compute $\mathbf{A}_{{\bar{l}}}^{+}$  and $\mathbf{W}_{{\bar{l}}}$ by (\ref{AinvLimNumda0AAiA1221}) and
(\ref{W2AinvY989565}), respectively;  
\While{\emph{The target training error  is not reached}}
\State Use added inputs ${{\mathbf{X}}_{ \bar p}}$ to get
$\mathbf{A}_{ \bar p}$
by
(\ref{ZxXa94319243}) and
(\ref{PaperEqu21AxZ23141});
\State  Update  ${{\bf{A}}_{\bar l}^{+ }}$ into    ${{\bf{A}}_{\bar l + \bar p}^{+ }}$ by   (\ref{xAbar2AbarBtDtBt4132OLD984})-(\ref{CAxDtA413124OLD984}),
(\ref{B_Matrix_defSimple3141}) and (\ref{D_bar_def_432ge32});
\State  Update $\mathbf{W}_{{\bar{l}}}$ into $\mathbf{W}_{{\bar{l}} + {\bar{p}} }$ by (\ref{xWnmBestCompute13434OLD984});
\State   $ l \Leftarrow l+p $;
\EndWhile
\State Set $\mathbf{W}=\mathbf{W}_{{\bar{l}}}$;
\end{algorithmic}
\end{algorithm}

\begin{table*}[!t]
\scriptsize
\renewcommand{\arraystretch}{1.3}
\newcommand{\tabincell}[2]{\begin{tabular}{@{}#1@{}}#2\end{tabular}}
\caption{Comparison of Flops among the BLS Algorithms} \label{table_example} \centering
\begin{tabular}{c|c|c|c|c|c|c|c|}
& & \multicolumn{2}{c|}{{\bfseries  Existing BLS}}    &  \multicolumn{2}{c|}{{  \bfseries   \tabincell{c}{Proposed Recursive  BLS}   }}  &\multicolumn{2}{c|}{{ \bfseries   \tabincell{c}{Proposed Square-Root BLS} }} \\
   \hline
  \hline
\multirow{3}*{\rotatebox{90}{Initialization \;\,}} & &(\ref{AinvLimNumda0AAiA1221}) &$3k^2 l+k^3$  &(\ref{Qm1AAIdefine23213})   &$k^2 l+k^3$    &(\ref{Q2PiPiT9686954})  &$k^2 l+\frac{2}{3}k^3$  \\
\cdashline{3-8}
& &(\ref{W2AinvY989565}) &$2klc$  &(\ref{WfromQ03223kds423sa})   &$2klc+2k^2 c$    &(\ref{WfromFFFFFF93281329k3sd})  &$2klc+2k^2 c$  \\
\cline{3-8}
& &\multirow{2}*{ Total }   &{\bfseries   \tabincell{c}{$3k^2 l+k^3$\\$+2klc\approx$  }}  &\multirow{2}*{ Total }   &{\bfseries   \tabincell{c}{$k^2 l+k^3+$\\$2klc+2k^2 c\approx$ }} &\multirow{2}*{ Total }  &{\bfseries   \tabincell{c}{$k^2 l+\frac{2}{3}k^3+$\\$2klc+2k^2 c\approx$  }}  \\
 \cdashline{4-4} \cdashline{6-6}  \cdashline{8-8}
& &   &{\bfseries   \tabincell{c}{$3k^2 l+k^3$}}  &   &{\bfseries   \tabincell{c}{$k^2 l+k^3$ }} &  &{\bfseries   \tabincell{c}{$k^2 l+\frac{2}{3}k^3$ }}  \\
  \hline
 \hline
\multirow{9}*{\rotatebox{90}{Each Update for Incremental Learning \quad}} & &(\ref{xWnmBestCompute13434OLD984}) &$4cpk$  &\multirow{2}*{{\tabincell{c}{ ${{\mathbf{Q}}_{\bar l}} \mathbf{A}_{ \bar p}^{T}=$\\$(\mathbf{A}_{ \bar p} {{\mathbf{Q}}_{\bar l}})^{T}$}}}   &\multirow{2}*{$2k^2 p$}    &(\ref{K2AxLm94835})  &$p k^2$  \\
 \cdashline{3-4}  \cdashline{7-8}
& &{\tabincell{c}{(\ref{DtAxAmnInv324141OLD984}),  (\ref{CAxDtA413124OLD984}),(\ref{D_bar_def_432ge32}),  (\ref{xAbar2AbarBtDtBt4132OLD984})}}      &$2pkl \times 4$ &  & &(\ref{Lbig2LLwave59056})   &$k^3/3$ \\
 \cline{2-8}
& \multirow{4}*{\rotatebox{90}{$p < k$}} & \multirow{3}*{ (\ref{B_Matrix_def2bSimplebbb}) } &\multirow{3}*{ $4 p^2 k + p^3$ }  &  (\ref{BmatrixFromMyQ322a})
 &  $3k p^2 + p^3$     &${{\mathbf{S}}}{{(\mathbf{I}+{{\mathbf{S}}^{T}} \mathbf{S})}^{-1}}$ &$p k^2 + p^3+2p^2k$         \\
 \cdashline{5-8}
& &  &  &(\ref{BmatrixFromMyQ322aForQ})   &$k^2 p $ &(\ref{Lwave2IKK40425b}) &$p k^2+k^3/3$   \\
 \cdashline{5-8}
& &  &  &(\ref{xWmnIncreaseCompute86759Before})  &$4cpk$  &(\ref{WfromF3209da}) &$4cpk + c k^2$   \\
\cline{3-8}
& &\multirow{2}*{ Total } &{\bfseries   \tabincell{c}{$8pkl+4 p^2 k $\\ $+p^3+4cpk\approx$  }} &\multirow{2}*{ Total } &{\bfseries   \tabincell{c}{$3 p k^2 +3 p^2k$\\$+p^3+4cpk \approx $ }} &\multirow{2}*{ Total }  &{\bfseries   \tabincell{c}{$\frac{2}{3}k^3+3  p k^2+ 2p^2 k  $\\$+p^3+ck^2+4cpk\approx$ }}   \\
\cdashline{4-4} \cdashline{6-6}  \cdashline{8-8}
& &  & {\bfseries   \tabincell{c}{$8pkl+4 p^2 k +p^3$}}
&  & {\bfseries   \tabincell{c}{$3 p k^2 +3 p^2k+p^3$ }} &   &{\bfseries   \tabincell{c}{$\frac{2}{3}k^3+3  p k^2+ 2p^2 k+p^3$}}
        \\
\cline{2-8}
& \multirow{3}*{\rotatebox{90}{$p \ge k$}}& \multirow{2}*{(\ref{B_Matrix_def2bSimpleaaa})}   &\multirow{2}*{ $4k^2 p + k^3$} &(\ref{BmatrixFromMyQ322bForQ}) & $2k p^2 + 3k^3$ &(\ref{Lwave2IKK40425a})  &$p k^2 + \frac{2}{3}k^3$   \\
 \cdashline{5-8}
 &    & &  &(\ref{xWmnIncreaseCompute86759use231}) &$4cpk + 2c k^2 $ &(\ref{WfromF3209db}) &$4cpk +2 c k^2 $    \\
\cline{3-8}
& &\multirow{2}*{ Total }  & {\bfseries   \tabincell{c}{$8pkl+4k^2 p $\\$+ k^3+4cpk\approx$}}
&\multirow{2}*{ Total }  & {\bfseries   \tabincell{c}{$2k p^2+2 k^2 p + $\\ $3k^3+2c k^2 +4cpk\approx$}} &\multirow{2}*{ Total }   &{\bfseries   \tabincell{c}{$2k^2p + k^3+$\\ $ 2c k^2+4cpk\approx$}}
        \\
\cdashline{4-4} \cdashline{6-6}  \cdashline{8-8}
& &  & {\bfseries   \tabincell{c}{$8pkl+4k^2 p+ k^3$}}
&  & {\bfseries   \tabincell{c}{$2k p^2+2 k^2 p +3k^3$ }} &   &{\bfseries   \tabincell{c}{$2k^2p + k^3$}}
        \\
\hline
\end{tabular}
\end{table*}

\section{Proposed Two Ridge Solutions for Incremental BLS on Added Inputs}

The BLS in \cite{BL_trans_paper} utilizes the ridge regression to approximate the generalized
inverse $\mathbf{A}_{{}}^{+ }$,
 and then
  $\lambda$
must be very small (e.g.,
$\lambda={{10}^{-8}}$) in (\ref{AinvLimNumda0AAiA1221})
 to satisfy the assumption of $\lambda \to 0$.
In this paper, we develop the algorithms based on the ridge inverse (\ref{xAmnAmnTAmnIAmnT231413}) and the ridge solution
(\ref{xWbarMN2AbarYYa1341}). Accordingly,
the assumption of $\lambda \to 0$ for the original BLS \cite{BL_trans_paper} is no longer required,
 and $\lambda$ can be set to any positive real number.

   For the BLS in the usual case of $l>k$,
   we propose a recursive algorithm and a
   square-root algorithm,
   which are based on  the
  $k  \times k$  Hermitian matrix
  ${{\bf{A}}^{T}}{\bf{A}}+\lambda \mathbf{I}$
    in the ridge inverse (\ref{xAmnAmnTAmnIAmnT231413}),
    and
   iteratively
    update   its inverse and inverse Cholesky factor~\footnote{We follow the naming method
   in \cite{my_inv_chol_paper,TransSP2003Blast},  where the recursive algorithm updates the inverse matrix recursively, and the  square-root algorithm
    updates the square-root (including the Cholesky factor) of the inverse matrix.},
  respectively.
To reduce the complexity,
   both the proposed algorithms
   avoid computing the
 $k \times (l+p)$
  ridge inverse ${\bf{A}}_{\bar l + \bar p}^{\dagger}$,
    which is bigger than
   the
  $k  \times k$
   Hermitian matrix.

The proposed recursive BLS algorithm updates   the inverse   recursively by   the matrix inversion lemma~\cite{matrixInversionLemma},
while the proposed  square-root BLS algorithm
 updates the inverse Cholesky factor
 by multiplying it
  with an upper-triangular intermediate matrix.
 When there are more rows than columns in the newly added
 input matrix ${\bf{A}}_{ \bar p} \in {\Re ^{p \times k}}$,
  i.e., $p > k$,
  the inverse of a sum of matrices~\cite{InverseSumofMatrix8312} is utilized in the proposed
  algorithms
  to  compute the intermediate variables by a smaller matrix inversion or inverse Cholesky factorization.

\subsection{Proposed Recursive BLS Algorithm Based on Inverse of the Hermitian Matrix in the Ridge Inverse}

Let us define  the  $k \times k$  inverse matrix
\begin{equation}\label{Qm1AAIdefine23213}
{{\mathbf{Q}}_{\bar l}}={{\left( {{\bf{A}}_{\bar l}^{T}}{\bf{A}}_{\bar l}+\lambda \mathbf{I} \right)}^{-1}}
\end{equation}
to write (\ref{xAmnAmnTAmnIAmnT231413}) as
\begin{equation}\label{xA2QAt13413}
{{\bf{A}}_{\bar l }^{\dagger }}={{\mathbf{Q}}_{\bar l}}{{\bf{A}}_{\bar l }^{T}},
\end{equation}
which can be substituted into (\ref{xWbarMN2AbarYYa1341}) to obtain
\begin{equation}\label{WfromQ03223kds423sa}
{\mathbf{\tilde W}}_{\bar l}={{\mathbf{Q}}_{\bar l}}{{\bf{A}}_{\bar l }^{T}}{{\mathbf{Y}}_{\bar l}}.
\end{equation}
To compute  ${{\mathbf{Q}}_{\bar l+ \bar p}}$  from  ${{\mathbf{Q}}_{\bar l}}$, substitute (\ref{AxInputIncrease31232}) into (\ref{Qm1AAIdefine23213}) to obtain
\begin{equation}\label{Q2AAIAtAinv4133467}
{{\mathbf{Q}}_{\bar l+ \bar p}}={{\left( \left( {{\bf{A}}_{\bar l}^{T}}{\bf{A}}_{\bar l}+\lambda \mathbf{I} \right)-\mathbf{A}_{ \bar p}^{T}(-\mathbf{A}_{ \bar p}^{{}}) \right)}^{-1}},
\end{equation}
into which apply the matrix inversion lemma~\cite[Eq. (1a)]{matrixInversionLemma}
 \begin{displaymath}
{({\bf{A}} - {\bf{UBV}})^{ - 1}} = {{\bf{A}}^{ - 1}} + {{\bf{A}}^{ - 1}}{\bf{U}}{({{\bf{B}}^{ - 1}} - {\bf{V}}{{\bf{A}}^{ - 1}}{\bf{U}})^{ - 1}}{\bf{V}}{{\bf{A}}^{ - 1}},
\end{displaymath}
 to obtain
 \begin{align}
{{\mathbf{Q}}_{\bar l+ \bar p}}&={{\mathbf{Q}}_{\bar l}}+{{\mathbf{Q}}_{\bar l}}\mathbf{A}_{ \bar p}^{T}{{(\mathbf{I}-(-\mathbf{A}_{ \bar p}^{{}}){{\mathbf{Q}}_{\bar l}}
\mathbf{A}_{ \bar p}^{T})}^{-1}}(-\mathbf{A}_{ \bar p}^{{}}){{\mathbf{Q}}_{\bar l}}  \notag \\
&={{\mathbf{Q}}_{\bar l}}-{{\mathbf{Q}}_{\bar l}}\mathbf{A}_{ \bar p}^{T}{{(\mathbf{I}+\mathbf{A}_{ \bar p}^{{}}
{{\mathbf{Q}}_{\bar l}}\mathbf{A}_{ \bar p}^{T})}^{-1}}\mathbf{A}_{ \bar p}^{{}}{{\mathbf{Q}}_{\bar l}}.  \label{Q1toQQAIAQAAQ1314}
\end{align}
Then we  simplify  (\ref{Q1toQQAIAQAAQ1314}) into
 \begin{equation}\label{BmatrixFromMyQ322aForQ}
{{\mathbf{Q}}_{\bar l+ \bar p}}={{\mathbf{Q}}_{\bar l}}-{{\mathbf{\tilde B}}} \mathbf{A}_{ \bar p}^{{}}{{\mathbf{Q}}_{\bar l}},
 \end{equation}
where the intermediate result ${{\mathbf{\tilde B}}}$ is defined by
  \begin{equation}\label{BmatrixFromMyQ322a}
{{\mathbf{\tilde B}}}  = {{\mathbf{Q}}_{\bar l}}\mathbf{A}_{ \bar p}^{T}{{(\mathbf{I}+\mathbf{A}_{ \bar p}^{{}}{{\mathbf{Q}}_{\bar l}}\mathbf{A}_{ \bar p}^{T})}^{-1}}.
 \end{equation}

 The above
   ${{\mathbf{\tilde B}}}$
  can be applied
   to update
${{\bf{A}}_{\bar l}^{\dagger }}$ into ${{\bf{A}}_{\bar l + \bar p}^{\dagger }}$ by
\begin{equation}\label{xA2QAt13413Deduce121a4}
{{\bf{A}}_{\bar l + \bar p}^{\dagger }}= \left[ \begin{matrix}
{{\bf{A}}_{\bar l}^{\dagger }}-{{\mathbf{\tilde B}}} \mathbf{A}_{ \bar p}^{{}}{{\bf{A}}_{\bar l}^{\dagger }}     &  {{\mathbf{\tilde B}}} \\
\end{matrix} \right],
\end{equation}
which is deduced in Appendix A.
Then
(\ref{xA2QAt13413Deduce121a4})
and
${{\mathbf{Y}}_{\bar l + \bar p}}={\left[ {\begin{array}{*{20}{c}}
{{\bf{Y}}_{\bar l}^T}&{{\bf{ Y}}_{\bar p}^T}
\end{array}} \right]^T}$ are substituted
 into
  (\ref{xWbarMN2AbarYYa1341})
   to obtain ${\mathbf{\tilde W}}_{\bar l + \bar p}= \left[ \begin{matrix}
{{\bf{A}}_{\bar l}^{\dagger }}-{{\mathbf{\tilde B}}} \mathbf{A}_{ \bar p}^{{}}{{\bf{A}}_{\bar l}^{\dagger }}     &  {{\mathbf{\tilde B}}} \\
\end{matrix} \right]  {\left[ {\begin{array}{*{20}{c}}
{{\bf{Y}}_{\bar l}^T}&{{\bf{ Y}}_{\bar p}^T}
\end{array}} \right]^T} ={{\bf{A}}_{\bar l}^{\dagger }}\mathbf{Y}_{{\bar{l}}}-{{\mathbf{\tilde B}}}\mathbf{A}_{ \bar p}^{{}}{{\bf{A}}_{\bar l}^{\dagger }} \mathbf{Y}_{{\bar{l}}}+{{\mathbf{\tilde B}}}{{\mathbf{Y}}_{ \bar p}}$, into which substitute (\ref{xWbarMN2AbarYYa1341}) to deduce
\begin{align}
{\mathbf{\tilde W}}_{\bar l + \bar p}&=\mathbf{\tilde W}_{\bar l}-{{\mathbf{\tilde B}}}\mathbf{A}_{ \bar p}^{{}}\mathbf{\tilde W}_{\bar l}+{{\mathbf{\tilde B}}}
{{\mathbf{Y}}_{ \bar p}}  \notag \\
&=\mathbf{\tilde W}_{\bar l}+ {{\mathbf{\tilde B}}}
  \left( {{\mathbf{Y}}_{ \bar p}}-\mathbf{A}_{ \bar p}^{{}}\mathbf{\tilde W}_{\bar l} \right).  \label{xWmnIncreaseCompute86759Before}
\end{align}
\begin{table*}[!t]
\scriptsize
\renewcommand{\arraystretch}{1.3}
\newcommand{\tabincell}[2]{\begin{tabular}{@{}#1@{}}#2\end{tabular}}
\caption{Testing Accuracy of the BLS Algorithms on MNIST Dataset  with $p=10000>k=5100$} \label{table_example} \centering
\begin{tabular}{c|c|c||c|c|c|c|c|c|}
\hline
\multicolumn{3}{c||}{\bfseries  Number of Inputs }   &{{\bfseries   {10000}}}   &{{\bfseries   {$\xrightarrow[\scriptscriptstyle{10000}]{}$ 20000}}}   &{{\bfseries   {$\xrightarrow[\scriptscriptstyle{10000}]{}$ 30000}}}   &{{\bfseries   {$\xrightarrow[\scriptscriptstyle{10000}]{}$ 40000}}}   &{{\bfseries   {$\xrightarrow[\scriptscriptstyle{10000}]{}$ 50000}}}   &{{\bfseries   {$\xrightarrow[\scriptscriptstyle{10000}]{}$ 60000}}}   \\
\hline
\multirow{22}*{\rotatebox{90}{{\bfseries   \tabincell{c}{Testing Accuracy ($\% $) }} }} & \multirow{4}*{\rotatebox{90}{{\bfseries   \tabincell{c}{ $\lambda=$ \\ ${{10}^{-8}}$}} }}    & \textbf{Exst.}   &	 97.57  ($\pm$       0.114  )  &   98.27  ($\pm$       0.069  )  &   98.45  ($\pm$       0.061  )  &   98.56  ($\pm$       0.064  )  &   98.59  ($\pm$       0.058  )  &  \textbf{98.64}  ($\pm$       0.058  )	 \\
\cline{3-9}
& &  \textbf{Recur.}    & 97.57  ($\pm$       0.114  )  &   98.27  ($\pm$       0.072  )  &   98.45  ($\pm$       0.060  )  &   98.56  ($\pm$       0.063  )  &   98.58  ($\pm$       0.060  )  &   \textbf{98.64}  ($\pm$       0.056  )	 \\
\cdashline{3-9}
& &  \textbf{Sqrt.}     & 97.57  ($\pm$       0.114  )  &   98.27  ($\pm$       0.069  )  &   98.45  ($\pm$       0.061  )  &   98.56  ($\pm$       0.064  )  &   98.59  ($\pm$       0.061  )  &   \textbf{98.64}  ($\pm$       0.058  )		 \\
\cdashline{3-9}
& &  \textbf{D-Rdg}    &97.57  ($\pm$       0.114  )  &   98.27  ($\pm$       0.069  )  &   98.45  ($\pm$       0.061  )  &   98.56  ($\pm$       0.064  )  &   98.59  ($\pm$       0.061  )  &   \textbf{98.64}  ($\pm$       0.058  )  		\\
\cline{2-9}
 & \multirow{3}*{\rotatebox{90}{{\bfseries   \tabincell{c}{ $\lambda=$ \\ ${{10}^{-7}}$}} }}    & \textbf{Exst.}    &	97.76  ($\pm$       0.085  )  &   98.27  ($\pm$       0.070  )  &   98.43  ($\pm$       0.065  )  &   98.53  ($\pm$       0.070  )  &   98.55  ($\pm$       0.059  )  &   98.61  ($\pm$       0.053  )		 \\
\cline{3-9}
& &  \textbf{Recur.}    & 97.76  ($\pm$       0.085  )  &   98.28  ($\pm$       0.066  )  &   98.44  ($\pm$       0.057  )  &   98.55  ($\pm$       0.063  )  &   98.57  ($\pm$       0.055  )  &   98.63  ($\pm$       0.054  )	 	 \\
\cdashline{3-9}
& &  \textbf{Sqrt., D-Rdg}    &	 97.76  ($\pm$       0.085  )  &   98.28  ($\pm$       0.067  )  &   98.44  ($\pm$       0.057  )  &   98.55  ($\pm$       0.064  )  &   98.57  ($\pm$       0.055  )  &   98.63  ($\pm$       0.055  )   		\\
\cline{2-9}
& \multirow{3}*{\rotatebox{90}{{\bfseries   \tabincell{c}{ $\lambda=$ \\ ${{10}^{-6}}$}} }}    & \textbf{Exst.}    &	97.76  ($\pm$       0.081  )  &   98.13  ($\pm$       0.094  )  &   98.28  ($\pm$       0.092  )  &   98.39  ($\pm$       0.072  )  &   98.41  ($\pm$       0.083  )  &   98.46  ($\pm$       0.079  )		 \\
\cline{3-9}
& &  \textbf{Recur.}    &  97.76  ($\pm$       0.081  )  &   98.19  ($\pm$       0.078  )  &   98.36  ($\pm$       0.079  )  &   98.47  ($\pm$       0.066  )  &   98.51  ($\pm$       0.059  )  &   98.56  ($\pm$       0.059  )	 \\
\cdashline{3-9}
& &  \textbf{Sqrt., D-Rdg}    &	97.76  ($\pm$       0.081  )  &   98.19  ($\pm$       0.077  )  &   98.36  ($\pm$       0.079  )  &   98.47  ($\pm$       0.066  )  &   98.51  ($\pm$       0.059  )  &   98.56  ($\pm$       0.058  )   		\\
\cline{2-9}
 & \multirow{2}*{\rotatebox{90}{{\bfseries   \tabincell{c}{ $\lambda=$ \\ ${{10}^{-5}}$}} }}    & \textbf{Exst.}   &97.49  ($\pm$       0.105  )  &   97.82  ($\pm$       0.120  )  &   97.96  ($\pm$       0.122  )  &   98.07  ($\pm$       0.113  )  &   98.09  ($\pm$       0.125  )  &   98.13  ($\pm$       0.113  )	 		 \\
\cline{3-9}
& &  \textbf{Recur., Sqrt., D-Rdg}    &	   97.49  ($\pm$       0.105  )  &   97.93  ($\pm$       0.103  )  &   98.11  ($\pm$       0.104  )  &   98.24  ($\pm$       0.090  )  &   98.27  ($\pm$       0.093  )  &   98.33  ($\pm$       0.097  )   		 \\
\cline{2-9}
 & \multirow{2}*{\rotatebox{90}{{\bfseries   \tabincell{c}{ $\lambda=$ \\ ${{10}^{-4}}$}} }}    & \textbf{Exst.}   &	96.96  ($\pm$       0.172  )  &   97.21  ($\pm$       0.190  )  &   97.35  ($\pm$       0.190  )  &   97.45  ($\pm$       0.187  )  &   97.46  ($\pm$       0.170  )  &   97.52  ($\pm$       0.170  )   		 \\
\cline{3-9}
& &  \textbf{Recur., Sqrt., D-Rdg}    &	96.96  ($\pm$       0.172  )  &   97.38  ($\pm$       0.180  )  &   97.60  ($\pm$       0.161  )  &   97.74  ($\pm$       0.165  )  &   97.77  ($\pm$       0.155  )  &   97.85  ($\pm$       0.143  )   		 \\
\cline{2-9}
 & \multirow{2}*{\rotatebox{90}{{\bfseries   \tabincell{c}{ $\lambda=$ \\ ${{10}^{-3}}$}} }}    & \textbf{Exst.}   &  96.03  ($\pm$       0.260  )  &   96.24  ($\pm$       0.259  )  &   96.37  ($\pm$       0.260  )  &   96.48  ($\pm$       0.251  )  &   96.50  ($\pm$       0.251  )  &   96.58  ($\pm$       0.253  )   		 \\
\cline{3-9}
& &  \textbf{Recur., Sqrt., D-Rdg}    &	96.03  ($\pm$       0.260  )  &   96.46  ($\pm$       0.238  )  &   96.69  ($\pm$       0.231  )  &   96.87  ($\pm$       0.228  )  &   96.96  ($\pm$       0.218  )  &   97.08  ($\pm$       0.214  ) 		 \\
\cline{2-9}
 & \multirow{2}*{\rotatebox{90}{{\bfseries   \tabincell{c}{ $\lambda=$ \\ ${{10}^{-2}}$}} }}    & \textbf{Exst.}   & 94.55  ($\pm$       0.423  )  &   94.81  ($\pm$       0.380  )  &   94.94  ($\pm$       0.373  )  &   95.05  ($\pm$       0.369  )  &   95.12  ($\pm$       0.346  )  &   95.20  ($\pm$       0.347  ) 		 \\
\cline{3-9}
& &  \textbf{Recur., Sqrt., D-Rdg}    & 94.55  ($\pm$       0.423  )  &   95.15  ($\pm$       0.344  )  &   95.44  ($\pm$       0.312  )  &   95.65  ($\pm$       0.306  )  &   95.77  ($\pm$       0.290  )  &   95.92  ($\pm$       0.281  )  		 \\
\cline{2-9}
 & \multirow{2}*{\rotatebox{90}{{\bfseries   \tabincell{c}{ $\lambda=$ \\ ${{10}^{-1}}$}} }}    & \textbf{Exst.}   & 92.31  ($\pm$       0.535  )  &   92.65  ($\pm$       0.506  )  &   92.80  ($\pm$       0.486  )  &   92.94  ($\pm$       0.491  )  &   93.05  ($\pm$       0.483  )  &   93.13  ($\pm$       0.482  )   		 \\
\cline{3-9}
& &  \textbf{Recur., Sqrt., D-Rdg}    & 92.31  ($\pm$       0.535  )  &   93.15  ($\pm$       0.468  )  &   93.52  ($\pm$       0.463  )  &   93.83  ($\pm$       0.450  )  &   94.07  ($\pm$       0.433  )  &   94.23  ($\pm$       0.417  )  	 		 \\
\hline
\end{tabular}
\end{table*}

When  the added training samples are more than the
total nodes, i.e., $p > k$,
 the $p \times p$  matrix inverse  in   (\ref{BmatrixFromMyQ322a})
 for computing  ${{\mathbf{\tilde B}}}$    can be  simplified into
 a smaller $k \times k$
  inverse,
 by
  applying the inverse of a sum of matrices~\cite[Eq. (20)]{InverseSumofMatrix8312}
 \begin{equation}\label{SumOfMatricesEqu31431}
{{(\mathbf{I}+\mathbf{P} {{\mathbf{Q}}})}^{-1}}\mathbf{P}=\mathbf{P}{{(\mathbf{I}+ {{\mathbf{Q}}}\mathbf{P})}^{-1}}
\end{equation}
 to
write
(\ref{BmatrixFromMyQ322a})
 as
  \begin{equation}\label{BmatrixFromMyQ322b}
{{\mathbf{\tilde B}}}  = {{(\mathbf{I}+{{\mathbf{Q}}_{\bar l}}\mathbf{A}_{ \bar p}^{T} \mathbf{A}_{ \bar p}^{{}})}^{-1}}{{\mathbf{Q}}_{\bar l}}\mathbf{A}_{ \bar p}^{T}.
 \end{equation}
 Fortunately,
we can substitute  (\ref{BmatrixFromMyQ322b}) into (\ref{BmatrixFromMyQ322aForQ}) to obtain
\begin{align}
{{\mathbf{Q}}_{\bar l+ \bar p}}&={{\mathbf{Q}}_{\bar l}}-{{(\mathbf{I}+{{\mathbf{Q}}_{\bar l}}\mathbf{A}_{ \bar p}^{T} \mathbf{A}_{ \bar p}^{{}})}^{-1}}{{\mathbf{Q}}_{\bar l}}\mathbf{A}_{ \bar p}^{T}\mathbf{A}_{ \bar p}^{{}}  {{\mathbf{Q}}_{\bar l}}  \notag \\
&={{\mathbf{Q}}_{\bar l}}- {{(\mathbf{I}+{{\mathbf{Q}}_{\bar l}}\mathbf{A}_{ \bar p}^{T} \mathbf{A}_{ \bar p}^{{}})}^{-1}}
(\mathbf{I} + {{\mathbf{Q}}_{\bar l}}\mathbf{A}_{ \bar p}^{T}\mathbf{A}_{ \bar p}^{{}} - \mathbf{I}) {{\mathbf{Q}}_{\bar l}}  \notag \\
&={{\mathbf{Q}}_{\bar l}}-\left({\mathbf{I}}-{{(\mathbf{I}+{{\mathbf{Q}}_{\bar l}}\mathbf{A}_{ \bar p}^{T} \mathbf{A}_{ \bar p}^{{}})}^{-1}}\right)  {{\mathbf{Q}}_{\bar l}}  \notag  \\
&={{(\mathbf{I}+{{\mathbf{Q}}_{\bar l}}\mathbf{A}_{ \bar p}^{T} \mathbf{A}_{ \bar p}^{{}})}^{-1}}{{\mathbf{Q}}_{\bar l}},   \label{BmatrixFromMyQ322bForQ}
\end{align}
which is substituted into   (\ref{BmatrixFromMyQ322b}) to obtain
\begin{equation}\label{B2QmqAxt9954535}
{{\mathbf{\tilde B}}}= {{\mathbf{Q}}_{\bar l+ \bar p}} \mathbf{A}_{ \bar p}^{T}.
\end{equation}
Then we can compute  (\ref{BmatrixFromMyQ322bForQ}) and   (\ref{B2QmqAxt9954535})
instead of (\ref{BmatrixFromMyQ322b}),
to obtain ${{\mathbf{Q}}_{\bar l+ \bar p}}$  by (\ref{BmatrixFromMyQ322bForQ}) as a byproduct.
Moreover, we can avoid storing  ${{\mathbf{\tilde B}}}$ by
 substituting   (\ref{B2QmqAxt9954535}) into  (\ref{xWmnIncreaseCompute86759Before}) to deduce
\begin{equation}\label{xWmnIncreaseCompute86759use231}
{\mathbf{\tilde W}}_{\bar l + \bar p}=\mathbf{\tilde W}_{\bar l}+ {{\mathbf{Q}}_{\bar l+ \bar p}} \mathbf{A}_{ \bar p}^{T}
  \left({{\mathbf{Y}}_{ \bar p}}-\mathbf{A}_{ \bar p}^{{}}\mathbf{\tilde W}_{\bar l} \right),
\end{equation}
which computes ${\mathbf{\tilde W}}_{\bar l + \bar p}$ from ${{\mathbf{Q}}_{\bar l+ \bar p}}$ directly.

We summarize
 (\ref{BmatrixFromMyQ322a}),
(\ref{BmatrixFromMyQ322aForQ}),
(\ref{xWmnIncreaseCompute86759Before}),
(\ref{BmatrixFromMyQ322bForQ}) and (\ref{xWmnIncreaseCompute86759use231})
in \textbf{Algorithm 3},
where we choose a smaller matrix inverse according to the size of  ${\bf{A}}_{ \bar p}\in {\Re ^{p \times k}}$.
Notice that when $p=k$, we choose (\ref{BmatrixFromMyQ322bForQ}) and (\ref{xWmnIncreaseCompute86759use231})
with lower computational complexity.
Then for the proposed recursive BLS algorithm, we  list the procedure to compute the
output weights and the incremental learning for added inputs in \textbf{Algorithm 4},
where  the function ${\psi _1}\left(\bullet \right)$
is defined by the above \textbf{Algorithm 3}.


 \begin{algorithm}
\caption{The Proposed Algorithm to Update ${{\mathbf{Q}}_{\bar l}}$ and  ${\mathbf{\tilde W}}_{\bar l}$}\label{euclid}
\begin{algorithmic}[0]
\Function{${\psi _1}$}{$\mathbf{\tilde W}_{\bar l}$, ${{\mathbf{Q}}_{\bar l}}$, $\mathbf{A}_{ \bar p}$, ${{\mathbf{Y}}_{ \bar p}}$}
\State Let $p$ and $k$ denote the dimensions of ${\bf{A}}_{ \bar p}\in {\Re ^{p \times k}}$;
\If{\emph{$p \ge k$}}
\State ${{\mathbf{Q}}_{\bar l+ \bar p}}={{(\mathbf{I}+{{\mathbf{Q}}_{\bar l}}\mathbf{A}_{ \bar p}^{T} \mathbf{A}_{ \bar p}^{{}})}^{-1}}{{\mathbf{Q}}_{\bar l}}$;
\State ${\mathbf{\tilde W}}_{\bar l + \bar p}=\mathbf{\tilde W}_{\bar l}+ {{\mathbf{Q}}_{\bar l+ \bar p}} \mathbf{A}_{ \bar p}^{T}
  \left({{\mathbf{Y}}_{ \bar p}}-\mathbf{A}_{ \bar p}^{{}}\mathbf{\tilde W}_{\bar l} \right)$;
\Else[\emph{$p < k$}]
\State ${{\mathbf{\tilde B}}}  = {{\mathbf{Q}}_{\bar l}}\mathbf{A}_{ \bar p}^{T}{{(\mathbf{I}+\mathbf{A}_{ \bar p}^{{}}{{\mathbf{Q}}_{\bar l}}\mathbf{A}_{ \bar p}^{T})}^{-1}}$;
\State ${{\mathbf{Q}}_{\bar l+ \bar p}}={{\mathbf{Q}}_{\bar l}}-{{\mathbf{\tilde B}}} \mathbf{A}_{ \bar p}^{{}}{{\mathbf{Q}}_{\bar l}}$;
\State  ${\mathbf{\tilde W}}_{\bar l + \bar p}=\mathbf{\tilde W}_{\bar l}+ {{\mathbf{\tilde B}}}
  \left( {{\mathbf{Y}}_{ \bar p}}-\mathbf{A}_{ \bar p}^{{}}\mathbf{\tilde W}_{\bar l} \right)$;
\EndIf
\State \textbf{return} ${{\mathbf{Q}}_{\bar l+ \bar p}}$, ${\mathbf{\tilde W}}_{\bar l + \bar p}$
\EndFunction
\end{algorithmic}
\end{algorithm}

\begin{algorithm}
\caption{:~\bf The Proposed Recursive BLS Algorithm Based on Inverse of the Hermitian Matrix}   
\begin{algorithmic}[1]
\Require $\mathbf{A}_{{\bar{l}}}$,  $\mathbf{Y}_{{\bar{l}}}$,  ${{\mathbf{X}}_{ \bar p}}$ and  ${{\mathbf{Y}}_{ \bar p}}$
\Ensure Output weights $\mathbf{W}$
\State Compute ${{\mathbf{Q}}_{\bar l}}$ and ${\mathbf{\tilde W}}_{\bar l}$ by
(\ref{Qm1AAIdefine23213}) and  (\ref{WfromQ03223kds423sa}), respectively;
\While{\emph{The target training error  is not reached}}
\State  Use  ${{\mathbf{X}}_{ \bar p}}$ to get $\mathbf{A}_{ \bar p}$ by (\ref{ZxXa94319243}) and (\ref{PaperEqu21AxZ23141});
\State Compute ${{\mathbf{Q}}_{\bar l+ \bar p}}$ and ${\mathbf{\tilde W}}_{\bar l + \bar p}$  by
 ${\psi _1}( \mathbf{\tilde W}_{\bar l}, {{\mathbf{Q}}_{\bar l}}, \mathbf{A}_{ \bar p}, {{\mathbf{Y}}_{ \bar p}})$;
\State   $ l \Leftarrow l+p $;
\EndWhile
\State Set $\mathbf{W}=\mathbf{W}_{{\bar{l}}}$;
\end{algorithmic}
\end{algorithm}
\subsection{Proposed  Square-Root BLS Algorithm Based on Inverse Cholesky Factor of the Hermitian Matrix in the Ridge Inverse}
Obviously
${{{{\bf{A}}_{\bar l}^{T}}{\bf{A}}_{\bar l}+\lambda \mathbf{I}}}$
is positive definite for $\lambda >0$.
 Then we can assume that the inverse Cholesky factor~\cite{BL_trans_paper} of
 ${{{{\bf{A}}_{\bar l}^{T}}{\bf{A}}_{\bar l}+\lambda \mathbf{I}}}$
  is the upper-triangular  ${{\mathbf{F}}_{\bar l}}$
 satisfying
\begin{equation}\label{Q2PiPiT9686954}
{\mathbf{F}}_{\bar l}{{\mathbf{F}}_{\bar l}^{T}}={({{{{\bf{A}}_{\bar l}^{T}}{\bf{A}}_{\bar l}+\lambda \mathbf{I}}})^{-1}}={{\mathbf{Q}}_{\bar l}},
\end{equation}
which is substituted into (\ref{WfromQ03223kds423sa}) (i.e., ${\mathbf{\tilde W}}_{\bar l}={{\mathbf{Q}}_{\bar l}}{{\bf{A}}_{\bar l }^{T}}{{\mathbf{Y}}_{\bar l}}$) to obtain
\begin{equation}\label{WfromFFFFFF93281329k3sd}
{\mathbf{\tilde W}}_{\bar l}={\mathbf{F}}_{\bar l}{{\mathbf{F}}_{\bar l}^{T}}{{\bf{A}}_{\bar l }^{T}}{{\mathbf{Y}}_{\bar l}}.
\end{equation}

To deduce the algorithm that updates ${\mathbf{F}}_{\bar l}$
into ${{\mathbf{F}}_{{\bar{l}}+\bar{p}}}$,
substitute (\ref{Q2PiPiT9686954})
into  (\ref{Q1toQQAIAQAAQ1314})
 to obtain
\begin{align}
{{\mathbf{F}}_{{\bar{l}}+\bar{p}}}\mathbf{F}_{{\bar{l}}+\bar{p}}^{T} &
={{\mathbf{F}}_{{{\bar{l}}}}}\mathbf{F}_{{{\bar{l}}}}^{T}-
 {{\mathbf{F}}_{{{\bar{l}}}}}\mathbf{F}_{{{\bar{l}}}}^{T}\mathbf{A}_{ \bar p}^{T}{(\mathbf{I}+\mathbf{A}_{ \bar p}^{{}}
{{\mathbf{F}}_{{{\bar{l}}}}}\mathbf{F}_{{{\bar{l}}}}^{T}\mathbf{A}_{ \bar p}^{T})^{-1}}\mathbf{A}_{ \bar p}^{{}}
{{\mathbf{F}}_{{{\bar{l}}}}}\mathbf{F}_{{{\bar{l}}}}^{T}   \notag \\
& = {{\mathbf{F}}_{{{\bar{l}}}}}\left( \mathbf{I}-\mathbf{F}_{{{\bar{l}}}}^{T}\mathbf{A}_{ \bar p}^{T} {{(\mathbf{I}+ \mathbf{A}_{ \bar p} \mathbf{F}_{{{\bar{l}}}} \mathbf{F}_{{{\bar{l}}}}^{T}\mathbf{A}_{ \bar p}^{T})}^{-1}}  \mathbf{A}_{ \bar p} \mathbf{F}_{{{\bar{l}}}} \right)\mathbf{F}_{{{\bar{l}}}}^{T}   \notag \\
& ={{\mathbf{F}}_{{{\bar{l}}}}}\left( \mathbf{I}-{{\mathbf{S}}}{{(\mathbf{I}+{{\mathbf{S}}^{T}}\mathbf{S})}^{-1}}{{\mathbf{S}}^{T}} \right)\mathbf{F}_{{{\bar{l}}}}^{T},  \label{FF2FISISSSF9320ds23}
\end{align}
where  the intermediate matrix $\mathbf{S} \in {\Re ^{k  \times p}} $
is defined
 by
\begin{equation}\label{K2AxLm94835}
\mathbf{S}=\mathbf{F}_{{{\bar{l}}}}^{T}\mathbf{A}_{ \bar p}^{T}.
\end{equation}
We can write (\ref{FF2FISISSSF9320ds23})  as
\begin{equation}\label{FF2FVVF329090ds32}
{{\mathbf{F}}_{{\bar{l}}+\bar{p}}}\mathbf{F}_{{\bar{l}}+\bar{p}}^{T} ={{\mathbf{F}}_{{{\bar{l}}}}}{\mathbf{V}} {\mathbf{V}}^{T}\mathbf{F}_{{{\bar{l}}}}^{T},
 \end{equation}
where
the upper-triangular matrix $\mathbf{V}$ is defined by
 \begin{equation}\label{Lwave2IKK40425bBefore}
 {\mathbf{V}} {\mathbf{V}}^{T} =
\mathbf{I}- \mathbf{S}{{(\mathbf{I}+{{\mathbf{S}}^{T}} \mathbf{S})}^{-1}} {{\mathbf{S}}^{T}}.
\end{equation}
Then from
(\ref{FF2FVVF329090ds32}),
 we  deduce the algorithm to update ${\mathbf{F}}_{\bar l}$
 by
\begin{equation}\label{Lbig2LLwave59056}
{{\mathbf{F}}_{{\bar{l}}+\bar{p}}}={{\mathbf{F}}_{{{\bar{l}}}}}\mathbf{V},
\end{equation}
where  ${{\mathbf{F}}_{{\bar{l}}+\bar{p}}}$
 must  be upper triangular, since both ${{\mathbf{F}}_{{{\bar{l}}}}}$ and $\mathbf{V}$ are upper triangular.
 Moreover, to use  the inverse Cholesky factor to update $\mathbf{\tilde W}_{\bar l}$ into ${\mathbf{\tilde W}}_{\bar l + \bar p}$,
 substitute  (\ref{Q2PiPiT9686954}) into (\ref{BmatrixFromMyQ322a})
    to get
 \begin{equation}\label{BwaveFromF1st3232a}
{{\mathbf{\tilde B}}} = {\mathbf{F}}_{\bar l}{{\mathbf{F}}_{\bar l}^{T}}\mathbf{A}_{ \bar p}^{T}{{(\mathbf{I}+\mathbf{A}_{ \bar p}^{{}}{\mathbf{F}}_{\bar l}{{\mathbf{F}}_{\bar l}^{T}}\mathbf{A}_{ \bar p}^{T})}^{-1}},
 \end{equation}
and then substitute  (\ref{K2AxLm94835}) (i.e., $\mathbf{S}=\mathbf{F}_{{{\bar{l}}}}^{T}\mathbf{A}_{ \bar p}^{T}$) into (\ref{BwaveFromF1st3232a})
to get
\begin{equation}\label{BwaveFromF1st3232a2sd2322}
{{\mathbf{\tilde B}}} = {\mathbf{F}}_{\bar l} \mathbf{S} {{(\mathbf{I}+\mathbf{S}^{T}\mathbf{S} )}^{-1}},
\end{equation}
which is substituted
 into (\ref{xWmnIncreaseCompute86759Before}) finally to obtain
    \begin{equation}\label{WfromF3209daBefore}
{\mathbf{\tilde W}}_{\bar l + \bar p} = {{{\bf{\tilde W}}}_{\bar l}} + {\mathbf{F}}_{\bar l} \mathbf{S} {{(\mathbf{I}+\mathbf{S}^{T}\mathbf{S} )}^{-1}} ({{{\bf{Y}}_{ \bar p}} - {\bf{A}}_{ \bar p}^{}{{{\bf{\tilde W}}}_{\bar l}}}).
 \end{equation}

  When  the added training samples are no less than the
total nodes, i.e., $p \ge k$,
 we   utilize (\ref{SumOfMatricesEqu31431}) (i.e., the inverse of a sum of matrices~\cite{InverseSumofMatrix8312})
 to write (\ref{Lwave2IKK40425bBefore})
 as
\begin{align}
\mathbf{V}{{\mathbf{V}}^{T}} & =\mathbf{I}-{{(\mathbf{I}+\mathbf{S}{{\mathbf{S}}^{T}})}^{-1}}
\mathbf{S}{{\mathbf{S}}^{T}}=\mathbf{I}-\mathbf{I}+{{(\mathbf{I}+\mathbf{S}{{\mathbf{S}}^{T}})}^{-1}}  \notag \\
& ={{(\mathbf{I}+\mathbf{S}{{\mathbf{S}}^{T}})}^{-1}},    \label{Lwave2IKK40425aBefore}
\end{align}
which uses a smaller $k \times k$
  inverse Cholesky factorization instead of the $p \times p$  matrix inverse  in   (\ref{Lwave2IKK40425bBefore}).
Moreover, we also substitute  (\ref{Q2PiPiT9686954}) into (\ref{xWmnIncreaseCompute86759use231})
 to 
 update $\mathbf{\tilde W}_{\bar l}$ into ${\mathbf{\tilde W}}_{\bar l + \bar p}$ by
    \begin{equation}\label{WfromF3209dbBefore932}
   {\mathbf{\tilde W}}_{\bar l + \bar p} = {{{\bf{\tilde W}}}_{\bar l}} + {{\bf{F}}_{\bar l + \bar p}}{\bf{F}}_{\bar l + \bar p}^T {\bf{A}}_{ \bar p}^T ({{{\bf{Y}}_{ \bar p}} - {\bf{A}}_{ \bar p}^{}{{{\bf{\tilde W}}}_{\bar l}}}).
 \end{equation}



From (\ref{Lwave2IKK40425bBefore}),
(\ref{WfromF3209daBefore}),
(\ref{Lwave2IKK40425aBefore}) and
(\ref{WfromF3209dbBefore932}), it can be seen that we can
 compute the upper-triangular
$\mathbf{V}$
 and update ${\mathbf{\tilde W}}$
by
\begin{subnumcases}{\label{VWpSmallThaNk32230}}
{\mathbf{V}} {\mathbf{V}}^{T}= \mathbf{I}- \mathbf{S} {{(\mathbf{I}+{{\mathbf{S}}^{T}}\mathbf{S})}^{-1}}{{\mathbf{S}}^{T}} &  \label{Lwave2IKK40425b}\\
{\mathbf{\tilde W}}_{\bar l + \bar p} = {{{\bf{\tilde W}}}_{\bar l}} + {\mathbf{F}}_{\bar l} \mathbf{S} {{(\mathbf{I}+\mathbf{S}^{T}\mathbf{S} )}^{-1}} ({{{\bf{Y}}_{ \bar p}} - {\bf{A}}_{ \bar p}^{}{{{\bf{\tilde W}}}_{\bar l}}})   &  \label{WfromF3209da}
\end{subnumcases}
when  $p<k$,  or by
\begin{subnumcases}{\label{VWpBigThaNk02kds32s}}
{\mathbf{V}} {\mathbf{V}}^{T} ={{(\mathbf{I}+\mathbf{S} {{\mathbf{S}}^{T}})}^{-1}} &  \label{Lwave2IKK40425a}\\
{\mathbf{\tilde W}}_{\bar l + \bar p} = {{{\bf{\tilde W}}}_{\bar l}} + {{\bf{F}}_{\bar l + \bar p}}{\bf{F}}_{\bar l + \bar p}^T {\bf{A}}_{ \bar p}^T ({{{\bf{Y}}_{ \bar p}} - {\bf{A}}_{ \bar p}^{}{{{\bf{\tilde W}}}_{\bar l}}})  &  \label{WfromF3209db}
\end{subnumcases}
when $p \ge k$.
Notice that  $\mathbf{V}$ computed by (\ref{Lwave2IKK40425b}) is
the upper-triangular Cholesky factor~\footnote{A method has been introduced to transfer the upper-triangular Cholesky factor into the traditional lower-triangular Cholesky factor by permuting rows and columns,
on Page 45 of \cite{my_inv_chol_paper} (in the paragraph beginning on the $23$-rd line of the first column). In Matlab simulations,
$\mathbf{V}$ can be computed by $\mathbf{V}=fliplr(flipud(chol(fliplr(flipud(\mathbf{I}- \mathbf{S} {{(\mathbf{I}+{{\mathbf{S}}^{T}}\mathbf{S})}^{-1}}{{\mathbf{S}}^{T}})), 'lower')))$.},
  which
 is different from the traditional lower-triangular Cholesky factor~\cite{Matrix_Computations_book}.
 On the other hand, to compute $\mathbf{V}$ by  (\ref{Lwave2IKK40425a}),
 we can invert and transpose the traditional lower-triangular Cholesky
factor
of $\mathbf{I}+\mathbf{S} {{\mathbf{S}}^{T}}$,
 or use
  the inverse Cholesky factorization~\cite{my_inv_chol_paper}.

We summarize (\ref{K2AxLm94835}),
   (\ref{Lbig2LLwave59056}),  (\ref{VWpSmallThaNk32230})   and
    (\ref{VWpBigThaNk02kds32s})  in  the following \textbf{Algorithm 5}.
    Then for the proposed square-root BLS algorithm, we  list the procedure to compute the
output weights and the incremental learning for added inputs in \textbf{Algorithm 6},
where  the function ${\psi _2}\left(\bullet \right)$
is defined by the above \textbf{Algorithm 5}.

  \begin{algorithm}
\caption{The Proposed Algorithm to Update  ${{\mathbf{F}}_{\bar l}}$ and ${\mathbf{\tilde W}}_{\bar l}$}\label{euclid}
\begin{algorithmic}[0]
\Function{${\psi _2}$}{$\mathbf{\tilde W}_{\bar l}$, ${{\mathbf{F}}_{\bar l}}$, $\mathbf{A}_{ \bar p}$, ${{\mathbf{Y}}_{ \bar p}}$}
\State Let $p$ and $k$ denote the dimensions of ${\bf{A}}_{ \bar p}\in {\Re ^{p \times k}}$;
\State       $\mathbf{S}=\mathbf{F}_{{{\bar{l}}}}^{T}\mathbf{A}_{ \bar p}^{T}$;
\If{\emph{$p \ge k$}}
\State  Get $\mathbf{V}$ by  ${\mathbf{V}} {\mathbf{V}}^{T} = {{(\mathbf{I}+\mathbf{S} {{\mathbf{S}}^{T}})}^{-1}}$;
\State ${{\mathbf{F}}_{{\bar{l}}+\bar{p}}}={{\mathbf{F}}_{{{\bar{l}}}}}\mathbf{V}$;
\State ${\mathbf{\tilde W}}_{\bar l + \bar p} =\mathbf{\tilde W}_{\bar l}+ {\mathbf{F}}_{\bar l+ \bar p}{{\mathbf{F}}_{\bar l+ \bar p}^{T}} \mathbf{A}_{ \bar p}^{T} \left({{\mathbf{Y}}_{ \bar p}}-\mathbf{A}_{ \bar p}^{{}}\mathbf{\tilde W}_{\bar l} \right)$;
\Else[\emph{$p < k$}]
\State Get $\mathbf{V}$ by  ${\mathbf{V}} {\mathbf{V}}^{T} = \mathbf{I}- \mathbf{S} {{(\mathbf{I}+{{\mathbf{S}}^{T}}\mathbf{S})}^{-1}}{{\mathbf{S}}^{T}}$;
\State ${{\mathbf{F}}_{{\bar{l}}+\bar{p}}}={{\mathbf{F}}_{{{\bar{l}}}}}\mathbf{V}$;
\State   ${\mathbf{\tilde W}}_{\bar l + \bar p} = {{{\bf{\tilde W}}}_{\bar l}} + {\mathbf{F}}_{\bar l} \mathbf{S} {{(\mathbf{I}+\mathbf{S}^{T}\mathbf{S} )}^{-1}} \left( {{{\bf{Y}}_{ \bar p}} - {\bf{A}}_{ \bar p}^{}{{{\bf{\tilde W}}}_{\bar l}}} \right)$;
\EndIf
\State \textbf{return} ${{\mathbf{F}}_{\bar l+ \bar p}}$, ${\mathbf{\tilde W}}_{\bar l + \bar p}$
\EndFunction
\end{algorithmic}
\end{algorithm}


\begin{algorithm}
\caption{:~\bf The Proposed  Square-Root BLS Algorithm  by Inverse Cholesky Factorization}
\begin{algorithmic}[1]
\Require $\mathbf{A}_{{\bar{l}}}$,  $\mathbf{Y}_{{\bar{l}}}$,  ${{\mathbf{X}}_{ \bar p}}$ and  ${{\mathbf{Y}}_{ \bar p}}$
\Ensure Output weights $\mathbf{W}$
\State Compute ${\mathbf{F}}_{\bar l}$  and ${\mathbf{\tilde W}}_{\bar l}$ by
(\ref{Q2PiPiT9686954}) and  (\ref{WfromFFFFFF93281329k3sd}), respectively;
\While{\emph{The target training error  is not reached}}
\State  Use  ${{\mathbf{X}}_{ \bar p}}$ to get $\mathbf{A}_{ \bar p}$ by (\ref{ZxXa94319243}) and (\ref{PaperEqu21AxZ23141});
\State Compute ${{\mathbf{F}}_{\bar l+ \bar p}}$ and ${\mathbf{\tilde W}}_{\bar l + \bar p}$  by
 ${\psi _2}( \mathbf{\tilde W}_{\bar l}, {{\mathbf{F}}_{\bar l}}, \mathbf{A}_{ \bar p}, {{\mathbf{Y}}_{ \bar p}})$;
\State   $ l \Leftarrow l+p $;
\EndWhile
\State Set $\mathbf{W}=\mathbf{W}_{{\bar{l}}}$;
\end{algorithmic}
\end{algorithm}

\begin{table*}[!t]
\scriptsize
\renewcommand{\arraystretch}{1.3}
\newcommand{\tabincell}[2]{\begin{tabular}{@{}#1@{}}#2\end{tabular}}
\caption{Testing Accuracy of the BLS Algorithms  on MNIST Dataset  with $p=4000<k=15110$} \label{table_example} \centering
\begin{tabular}{c|c|c||c|c|c|c|c|c|}
\hline
\multicolumn{3}{c||}{\bfseries  Number of Inputs }   &{{\bfseries   {40000}}}   &{{\bfseries   {$\xrightarrow[\scriptscriptstyle{4000}]{}$ 44000}}}   &{{\bfseries   {$\xrightarrow[\scriptscriptstyle{4000}]{}$ 48000}}}   &{{\bfseries   {$\xrightarrow[\scriptscriptstyle{4000}]{}$ 52000}}}   &{{\bfseries   {$\xrightarrow[\scriptscriptstyle{4000}]{}$ 56000}}}   &{{\bfseries   {$\xrightarrow[\scriptscriptstyle{4000}]{}$ 60000}}}   \\
\hline
\multirow{22}*{\rotatebox{90}{{\bfseries   \tabincell{c}{Testing Accuracy ($\% $) }} }} & \multirow{4}*{\rotatebox{90}{{\bfseries   \tabincell{c}{ $\lambda=$ \\ ${{10}^{-8}}$}} }}    & \textbf{Exst.}   &98.74  ($\pm$    0.062) &   98.81  ($\pm$    0.064) &   98.86  ($\pm$    0.055) &   98.88  ($\pm$    0.055) &   98.92  ($\pm$    0.057) &   \textbf{98.92}  ($\pm$    0.059)	 \\
\cline{3-9}
& &  \textbf{Recur.}    &98.74  ($\pm$    0.062) &   98.81  ($\pm$    0.064) &   98.86  ($\pm$    0.058) &   98.88  ($\pm$    0.055) &   98.92  ($\pm$    0.052) &   \textbf{98.93}  ($\pm$    0.057)	 	 \\
\cdashline{3-9}
& &  \textbf{Sqrt.}     &98.74  ($\pm$    0.062) &   98.81  ($\pm$    0.064) &   98.86  ($\pm$    0.058) &   98.88  ($\pm$    0.056) &   98.92  ($\pm$    0.053) &   \textbf{98.93}  ($\pm$    0.057)	 \\
\cdashline{3-9}
& &  \textbf{D-Rdg}    &98.74  ($\pm$    0.062) &   98.81  ($\pm$    0.063) &   98.86  ($\pm$    0.059) &   98.88  ($\pm$    0.056) &   98.92  ($\pm$    0.054) &   \textbf{98.93}  ($\pm$    0.057) 		\\
\cline{2-9}
  & \multirow{4}*{\rotatebox{90}{{\bfseries   \tabincell{c}{ $\lambda=$ \\ ${{10}^{-7}}$}} }}    & \textbf{Exst.}   &98.86  ($\pm$       0.054  )  &   98.89  ($\pm$       0.050  )  &   98.87  ($\pm$       0.044  )  &   98.88  ($\pm$       0.042  )  &   98.91  ($\pm$       0.046  )  &   98.92  ($\pm$       0.041  )	 \\
\cline{3-9}
& &  \textbf{Recur.}    &	  98.86  ($\pm$       0.054  )  &   98.89  ($\pm$       0.053  )  &   98.88  ($\pm$       0.046  )  &   98.88  ($\pm$       0.046  )  &   98.91  ($\pm$       0.047  )  &   98.92  ($\pm$       0.041  )	 	 \\
\cdashline{3-9}
& &  \textbf{Sqrt.}     &98.86  ($\pm$       0.054  )  &   98.89  ($\pm$       0.053  )  &   98.87  ($\pm$       0.046  )  &   98.88  ($\pm$       0.046  )  &   98.91  ($\pm$       0.047  )  &   98.92  ($\pm$       0.042  ) 		 \\
\cdashline{3-9}
& &  \textbf{D-Rdg}    &98.86  ($\pm$       0.054  )  &   98.89  ($\pm$       0.053  )  &   98.87  ($\pm$       0.046  )  &   98.88  ($\pm$       0.046  )  &   98.91  ($\pm$       0.047  )  &   98.92  ($\pm$       0.042  )		\\
\cline{2-9}
   & \multirow{4}*{\rotatebox{90}{{\bfseries   \tabincell{c}{ $\lambda=$ \\ ${{10}^{-6}}$}} }}    & \textbf{Exst.}  &98.75  ($\pm$       0.055  )  &   98.77  ($\pm$       0.056  )  &   98.79  ($\pm$       0.056  )  &   98.78  ($\pm$       0.055  )  &   98.81  ($\pm$       0.058  )  &   98.82  ($\pm$       0.060  )	 \\
\cline{3-9}
& &  \textbf{Recur.}    & 98.75  ($\pm$       0.055  )  &   98.78  ($\pm$       0.056  )  &   98.79  ($\pm$       0.055  )  &   98.80  ($\pm$       0.055  )  &   98.82  ($\pm$       0.061  )  &   98.84  ($\pm$       0.057  )	 \\
\cdashline{3-9}
& &  \textbf{Sqrt.}     & 98.75  ($\pm$       0.055  )  &   98.78  ($\pm$       0.056  )  &   98.79  ($\pm$       0.055  )  &   98.80  ($\pm$       0.054  )  &   98.82  ($\pm$       0.060  )  &   98.84  ($\pm$       0.057  )		 \\
\cdashline{3-9}
& &  \textbf{D-Rdg}    &98.75  ($\pm$       0.055  )  &   98.78  ($\pm$       0.056  )  &   98.79  ($\pm$       0.055  )  &   98.80  ($\pm$       0.054  )  &   98.82  ($\pm$       0.061  )  &   98.84  ($\pm$       0.057  )   		\\
\cline{2-9}
 & \multirow{2}*{\rotatebox{90}{{\bfseries   \tabincell{c}{ $\lambda=$ \\ ${{10}^{-5}}$}} }}    & \textbf{Exst.}   &	 98.49  ($\pm$       0.078  )  &   98.52  ($\pm$       0.078  )  &   98.53  ($\pm$       0.069  )  &   98.52  ($\pm$       0.071  )  &   98.56  ($\pm$       0.080  )  &   98.57  ($\pm$       0.081  ) 		 \\
\cline{3-9}
& &  \textbf{Recur., Sqrt., D-Rdg}    &	 98.49  ($\pm$       0.078  )  &   98.52  ($\pm$       0.075  )  &   98.54  ($\pm$       0.066  )  &   98.54  ($\pm$       0.071  )  &   98.60  ($\pm$       0.080  )  &   98.61  ($\pm$       0.080  )  		 \\
\cline{2-9}
 & \multirow{2}*{\rotatebox{90}{{\bfseries   \tabincell{c}{ $\lambda=$ \\ ${{10}^{-4}}$}} }}    & \textbf{Exst.}   & 98.08  ($\pm$       0.107  )  &   98.07  ($\pm$       0.109  )  &   98.08  ($\pm$       0.116  )  &   98.09  ($\pm$       0.116  )  &   98.11  ($\pm$       0.109  )  &   98.12  ($\pm$       0.111  ) 		 \\
\cline{3-9}
& &  \textbf{Recur., Sqrt., D-Rdg}    & 98.08  ($\pm$       0.107  )  &   98.08  ($\pm$       0.108  )  &   98.11  ($\pm$       0.112  )  &   98.12  ($\pm$       0.115  )  &   98.16  ($\pm$       0.110  )  &   98.18  ($\pm$       0.109  )  	 		 \\
\cline{2-9}
 & \multirow{2}*{\rotatebox{90}{{\bfseries   \tabincell{c}{ $\lambda=$ \\ ${{10}^{-3}}$}} }}    & \textbf{Exst.}   &	 97.32  ($\pm$       0.174  )  &   97.34  ($\pm$       0.178  )  &   97.36  ($\pm$       0.168  )  &   97.37  ($\pm$       0.175  )  &   97.39  ($\pm$       0.176  )  &   97.42  ($\pm$       0.173  ) 		 \\
\cline{3-9}
& &  \textbf{Recur., Sqrt., D-Rdg}    &	 97.32  ($\pm$       0.174  )  &   97.37  ($\pm$       0.180  )  &   97.40  ($\pm$       0.168  )  &   97.43  ($\pm$       0.172  )  &   97.47  ($\pm$       0.173  )  &   97.51  ($\pm$       0.162  )   		 \\
\cline{2-9}
 & \multirow{2}*{\rotatebox{90}{{\bfseries   \tabincell{c}{ $\lambda=$ \\ ${{10}^{-2}}$}} }}    & \textbf{Exst.}   &	 96.17  ($\pm$       0.224  )  &   96.19  ($\pm$       0.221  )  &   96.21  ($\pm$       0.216  )  &   96.22  ($\pm$       0.219  )  &   96.25  ($\pm$       0.226  )  &   96.28  ($\pm$       0.229  ) 		 \\
\cline{3-9}
& &  \textbf{Recur., Sqrt., D-Rdg}    &	 96.17  ($\pm$       0.224  )  &   96.22  ($\pm$       0.224  )  &   96.26  ($\pm$       0.220  )  &   96.30  ($\pm$       0.224  )  &   96.36  ($\pm$       0.220  )  &   96.41  ($\pm$       0.217  )   		 \\
\cline{2-9}
 & \multirow{2}*{\rotatebox{90}{{\bfseries   \tabincell{c}{ $\lambda=$ \\ ${{10}^{-1}}$}} }}    & \textbf{Exst.}   & 94.57  ($\pm$       0.300  )  &   94.61  ($\pm$       0.289  )  &   94.65  ($\pm$       0.293  )  &   94.68  ($\pm$       0.287  )  &   94.69  ($\pm$       0.288  )  &   94.72  ($\pm$       0.300  ) 		 \\
\cline{3-9}
& &  \textbf{Recur., Sqrt., D-Rdg}    &94.57  ($\pm$       0.300  )  &   94.65  ($\pm$       0.287  )  &   94.74  ($\pm$       0.289  )  &   94.81  ($\pm$       0.283  )  &   94.85  ($\pm$       0.287  )  &   94.92  ($\pm$       0.292  ) 		 \\
\hline
\end{tabular}
\end{table*}

In Appendix B, we develop the parallel implementation of
the above  square-root BLS algorithm based on inverse Cholesky factor,
for the distributed
BLS with data-parallelism.  When deducing the parallel implementation of
the  square-root BLS, we apply the parallel implementation of the inverse Cholesky factorization introduced in \cite{CholBLSnodesSubmitted}.

\subsection{Comparison of  Ridge Inverse and Generalized Inverse}

By comparing the
 algorithm to update the generalized inverse (i.e., (\ref{xAbar2AbarBtDtBt4132OLD984})-(\ref{BtWaveBest23134OLD984}))
and the proposed algorithm to update the ridge inverse (i.e., (\ref{BmatrixFromMyQ322a}) and
(\ref{xA2QAt13413Deduce121a4})),  it can be seen that
 the only difference
 lies
between  ${{\mathbf{B}}}$ and ${{\mathbf{\tilde B}}}$ computed by (\ref{BtWaveBest23134OLD984}) and
  (\ref{BmatrixFromMyQ322a}), respectively.
In the usual case with  ${\bf{C}} ={\bf{0}}$,
we only need to consider (\ref{B2DtDwhenC0arr3})  in (\ref{BtWaveBest23134OLD984}),
into which
  substitute (\ref{DtAxAmnInv324141OLD984})
  to obtain
\begin{align}
{{\mathbf{B}}}&={{\bf{A}}_{\bar l}^{+ }} (\mathbf{A}_{ \bar p}^{{}}{{\bf{A}}_{\bar l}^{+ }})^T {{\left({\bf{I}} + \mathbf{A}_{ \bar p}{{\bf{A}}_{\bar l}^{+ }}
(\mathbf{A}_{ \bar p}^{{}}{{\bf{A}}_{\bar l}^{+ }})^T\right)}^{ - 1}}  \notag \\
&={{\bf{A}}_{\bar l}^{+ }} ({{\bf{A}}_{\bar l}^{+ }})^T\mathbf{A}_{ \bar p}^{{T}} {{\left({\bf{I}} + \mathbf{A}_{ \bar p}{{\bf{A}}_{\bar l}^{+ }}
({{\bf{A}}_{\bar l}^{+ }})^T \mathbf{A}_{ \bar p}^{{T}} \right)}^{ - 1}}.  \label{B2CinvLimCtCCtI3421}
\end{align}
We can utilize (\ref{AinvLimNumda0AAiA1221}) to write the entry ${{\bf{A}}_{\bar l}^{+ }} ({{\bf{A}}_{\bar l}^{+ }})^T$
in (\ref{B2CinvLimCtCCtI3421}) as
\begin{align}
{{\bf{A}}_{\bar l}^{+ }} ({{\bf{A}}_{\bar l}^{+ }})^T&=
\underset{\lambda \to 0}{\mathop{\lim }}\,{{(\mathbf{A}_{{\bar l}}^{T}\mathbf{A}_{{\bar l}}+\lambda \mathbf{I})}^{-1}}\mathbf{A}_{{\bar l}}^{T}
\mathbf{A}_{{\bar l}}{{(\mathbf{A}_{{\bar l}}^{T}\mathbf{A}_{{\bar l}}+\lambda \mathbf{I})}^{-1}}  \notag \\
&=\underset{\lambda \to 0}{\mathop{\lim }}\,{{\left(\mathbf{I}+\lambda {{(\mathbf{A}_{{\bar l}}^{T}\mathbf{A}_{{\bar l}}+\lambda \mathbf{I})}^{-1}} \right)}}{{(\mathbf{A}_{{\bar l}}^{T}\mathbf{A}_{{\bar l}}+\lambda \mathbf{I})}^{-1}}  \notag \\
&= \underset{\lambda \to 0}{\mathop{\lim }}\,{({{\mathbf{Q}}_{\bar l}}+\lambda {{\mathbf{Q}}_{\bar l}} {{\mathbf{Q}}_{\bar l}})},  \label{AAt2IandQQnumda14367}
\end{align}
 where (\ref{Qm1AAIdefine23213}) is applied.
Then we  
 substitute (\ref{AAt2IandQQnumda14367}) into (\ref{B2CinvLimCtCCtI3421}) to get
\begin{multline}\label{B2CinvLimCtCCtI4Q496443}
{{\mathbf{B}}}=\underset{\lambda \to 0}{\mathop{\lim }}\,{({{\mathbf{Q}}_{\bar l}}+\lambda {{\mathbf{Q}}_{\bar l}} {{\mathbf{Q}}_{\bar l}})}
  \mathbf{A}_{ \bar p}^{{T}} \times \\
{{\left({\bf{I}} + \mathbf{A}_{ \bar p}\underset{\lambda \to 0}{\mathop{\lim }}\,{({{\mathbf{Q}}_{\bar l}}+\lambda {{\mathbf{Q}}_{\bar l}} {{\mathbf{Q}}_{\bar l}})} \mathbf{A}_{ \bar p}^{{T}} \right)}^{ - 1}}.
\end{multline}

Obviously, ${{\mathbf{\tilde B}}}$ computed by (\ref{BmatrixFromMyQ322a})
 is equal to
${{\mathbf{B}}}$ computed by
 (\ref{B2CinvLimCtCCtI4Q496443}) (i.e., (\ref{B2DtDwhenC0arr3})) when $\lambda \to 0$,
while (\ref{BmatrixFromMyQ322a})  is different from
(\ref{B2CinvLimCtCCtI4Q496443})  when
 $\lambda \to 0$ is not satisfied,
since usually  $\lambda {{\mathbf{Q}}_{\bar l}} {{\mathbf{Q}}_{\bar l}}$ in (\ref{B2CinvLimCtCCtI4Q496443})
cannot be neglected
in this case.
Thus  the ridge inverse updated by the proposed
 (\ref{BmatrixFromMyQ322a})
 and  (\ref{xA2QAt13413Deduce121a4})
 is equal to the ridge regression
  of the generalized
inverse updated by the existing
(\ref{xAbar2AbarBtDtBt4132OLD984}),
(\ref{DtAxAmnInv324141OLD984})
and
 (\ref{B2DtDwhenC0arr3})
  when $\lambda \to 0$,
while usually the former is
different from
the latter
when $\lambda \to 0$ is not satisfied.

\begin{table*}[!t]
  \scriptsize
\renewcommand{\arraystretch}{1.3}
\newcommand{\tabincell}[2]{\begin{tabular}{@{}#1@{}}#2\end{tabular}}
\caption{Training Time of the BLS Algorithms and the Corresponding Speedups  $(p=10000>k=5100)$} \label{table_example} \centering
\begin{tabular}{|c||c c c| c c |c c c|c c|}
\hline
 {\bfseries   \tabincell{c}{Number of\\  Input Patterns}}
      &\multicolumn{3}{c|}{{\bfseries   \tabincell{c}{Each Additional \\ Training Times (s) }}}
      &\multicolumn{2}{c|}{{\bfseries   \tabincell{c}{Speedups in \\ Each Additional \\ Training Time }}}
       &\multicolumn{3}{c|}{{\bfseries   \tabincell{c}{Accumulative \\ Training Times (s) }}}
      &\multicolumn{2}{c|}{{\bfseries   \tabincell{c}{Speedups in \\ Accumulative \\ Training Time }}}  \\
     &Exst.        &Recur.    &Sqrt.    &Recur.    &Sqrt.   &Exst.        &Recur.    &Sqrt.   &Recur.    &Sqrt.      \\
\hline
  \bfseries 10000   &5.32   &    4.36   &    3.65   &     &       &    5.32   &    4.36   &    3.65   &     &       \\
 \bfseries 10000 $\xrightarrow[\scriptscriptstyle{10000}]{}$ 20000  &20.55   &   10.35   &    6.39   &    1.99   &    3.22   &   25.87   &   14.70   &   10.03   &    1.76   &    2.58    \\
 \bfseries 20000 $\xrightarrow[\scriptscriptstyle{10000}]{}$ 30000  &34.98   &   10.46   &    6.48   &    3.34   &    5.40   &   60.85   &   25.17   &   16.51   &    2.42   &    3.68  \\
 \bfseries 30000 $\xrightarrow[\scriptscriptstyle{10000}]{}$ 40000    &48.67   &   10.48   &    6.48   &    4.64   &    7.51   &  109.52   &   35.65   &   22.99   &    3.07   &    4.76  \\
 \bfseries 40000 $\xrightarrow[\scriptscriptstyle{10000}]{}$ 50000  &62.42   &   10.43   &    6.46   &    5.99   &    9.66   &  171.94   &   46.08   &   29.45   &    3.73   &    5.84  \\
 \bfseries 50000 $\xrightarrow[\scriptscriptstyle{10000}]{}$ 60000  &76.66   &   10.36   &    6.46   &    7.40   &   11.87   &  248.60   &   56.44   &   35.91   &    4.41   &    6.92  \\
\hline
\end{tabular}
\end{table*}

\begin{table*}[!t]
\scriptsize
\renewcommand{\arraystretch}{1.3}
\newcommand{\tabincell}[2]{\begin{tabular}{@{}#1@{}}#2\end{tabular}}
\caption{Training Time of the BLS Algorithms and the Corresponding Speedups  $(p=4000<k=15110)$} \label{table_example} \centering
\begin{tabular}{|c||c c c| c c |c c c|c c|}
\hline
 {\bfseries   \tabincell{c}{Number of\\  Input Patterns}}
      &\multicolumn{3}{c|}{{\bfseries   \tabincell{c}{Each Additional \\ Training Times (s) }}}
      &\multicolumn{2}{c|}{{\bfseries   \tabincell{c}{Speedups in \\ Each Additional \\ Training Time }}}
       &\multicolumn{3}{c|}{{\bfseries   \tabincell{c}{Accumulative \\ Training Times (s) }}}
      &\multicolumn{2}{c|}{{\bfseries   \tabincell{c}{Speedups in \\ Accumulative \\ Training Time }}}  \\
     &Exst.        &Recur.    &Sqrt.    &Recur.    &Sqrt.   &Exst.        &Recur.    &Sqrt.   &Recur.    &Sqrt.      \\
\hline
  \bfseries 40000   &134.22   &   89.12   &   73.28   &       &       &  134.22   &   89.12   &   73.28   &      &   	  \\
 \bfseries 40000 $\xrightarrow[\scriptscriptstyle{4000}]{}$ 44000  &71.72   &   21.87   &   55.16   &    3.28   &    1.30   &  205.94   &  110.98   &  128.43   &    1.86   &    1.60	  \\
 \bfseries 44000 $\xrightarrow[\scriptscriptstyle{4000}]{}$ 48000  & 78.73   &   22.05   &   55.64   &    3.57   &    1.41   &  284.67   &  133.03   &  184.07   &    2.14   &    1.55	 \\
 \bfseries 48000 $\xrightarrow[\scriptscriptstyle{4000}]{}$ 52000  &83.94   &   22.21   &   55.50   &    3.78   &    1.51   &  368.61   &  155.24   &  239.57   &    2.37   &    1.54	  \\
 \bfseries 52000 $\xrightarrow[\scriptscriptstyle{4000}]{}$ 56000  &91.73   &   21.95   &   55.49   &    4.18   &    1.65   &  460.33   &  177.18   &  295.06   &    2.60   &    1.56  \\
 \bfseries 56000 $\xrightarrow[\scriptscriptstyle{4000}]{}$ 60000  &96.91   &   22.15   &   55.49   &    4.37   &    1.75   &  557.24   &  199.34   &  350.55   &    2.80   &    1.59	  \\
\hline
\end{tabular}
\end{table*}

\section{Complexity Comparison and Numerical Experiments}

To compare the learning speed and testing accuracy of  the existing BLS
on added inputs~\cite{BL_trans_paper,my_ppaapper1_on_BL} (i.e., \textbf{Algorithm 2}),
the
 proposed recursive BLS (i.e., \textbf{Algorithms 3} and \textbf{4})
and the proposed square-root BLS (i.e., \textbf{Algorithms 5} and \textbf{6}),
 we calculate  the expected flops (floating-point operations)
and conduct numerical experiments in this section.
The experiments are carried out
 on MATLAB software platform under a Microsoft-Windows Server with  $128$ GB of RAM.
 For the
enhancement nodes,
 the tansig function is chosen,
 and the weights
    ${{\mathbf{W}}_{{{h}_{j}}}}$ and
   the biases
    ${{\mathbf{\beta }}_{{{h}_{j}}}}$ ($j=1,2,\cdots, m$)
    are drawn from the
standard uniform distributions on the interval $\left[ {\begin{array}{*{20}{c}}
{{{ - }}1}&1
\end{array}} \right]$.

\subsection{Complexity  Comparison}

This subsection computes the expected flops
 of the existing BLS algorithm~\cite{BL_trans_paper,my_ppaapper1_on_BL} and
the proposed BLS algorithms.
It can  be seen that
 $l p (2 k - 1) \approx 2 l  k p$ flops are required to multiply
 a $l \times k$ matrix by a $k \times p$ matrix. To sum two matrices in size $l \times k$, only $l  k=0(l  k p)$  flops are required,
 which will be neglected  for simplicity in what follows.
 In Matlab,
  the  inv function~\cite{Matlab_inv_function_introduce} requires $\frac{1}{3} k^3$ flops~\cite{Matrix_Computations_book} to compute the ${\mathbf{LDL}}^T$ factors of the $k \times k$ Hermitian matrix $\mathbf{X}$,
 and  $\frac{2}{3} k^3$ flops
  to
 invert the factors and multiply the inverses.
 Thus it totally requires $k^3$ flops  to compute the inverse of the Hermitian matrix $\mathbf{X}$, while it
 totally requires $2 k^3$ flops  to compute the inverse of the non-Hermitian matrix $\mathbf{X}$ by the $\mathbf{LU}$ factorization.
   Moreover,  the inverse Cholesky factorization of a $k \times k$ matrix requires $k^3/3$ multiplications and additions~\cite{my_inv_chol_paper}, i.e., $\frac{2}{3}k^3$ flops.

 Table \Rmnum{1}  lists
the flops of
 the existing BLS
on added inputs,
the
 proposed recursive BLS 
and the proposed square-root BLS.
 We include the flops of each equation and
the  total flops, for the initialization of $\mathbf{W}_{{\bar{l}}}$
and the incremental learning to update $\mathbf{W}_{{\bar{l}}}$ into $\mathbf{W}_{{\bar{l}} + {\bar{p}} }$.
In  Table \Rmnum{1},
   the entry ${{\mathbf{Q}}_{\bar l}} \mathbf{A}_{ \bar p}^{T}=(\mathbf{A}_{ \bar p} {{\mathbf{Q}}_{\bar l}})^{T}$
is utilized in  (\ref{BmatrixFromMyQ322a}),
(\ref{BmatrixFromMyQ322aForQ})
and (\ref{BmatrixFromMyQ322bForQ}),
while  the entry ${{\mathbf{S}}}{{(\mathbf{I}+{{\mathbf{S}}^{T}} \mathbf{S})}^{-1}}$ is utilized in  (\ref{Lwave2IKK40425b}) and (\ref{WfromF3209da}).
Moreover, notice that when computing
    (\ref{BmatrixFromMyQ322aForQ}), we only need to
 obtain about half entries in the Hermitian
${{\mathbf{\tilde B}}} ({{\mathbf{Q}}_{\bar l}}\mathbf{A}_{ \bar p}^{T})^T$.
To obtain the dominant total flops of the BLS algorithms in Table \Rmnum{1},
we assume the usual case where the output nodes are much less than
the training samples and the total feature and
enhancement nodes, i.e.,
    \begin{equation}\label{cTo0l0k}
c=0(l)=0(k).
 \end{equation}






From the dominant total flops in Table \Rmnum{1}, it can be seen
that when computing the initial $\mathbf{W}_{{\bar{l}}}$,
 the proposed recursive and square-root BLS algorithms only
require~\footnote{Here we utilize $\frac{k}{l}<1$, i.e., (\ref{lbiggerThaNk39w32}).}
  \begin{equation}\label{FlopRatioRecur2Orig930ds}
\frac{k^2 l+k^3}{3k^2 l+k^3}= \frac{l+k}{3l+k}= \frac{1}{2/(1+\frac{k}{l})+1}<\frac{1}{2}
 \end{equation}
 and
   \begin{equation}\label{FlopRatioSqrt2Orig930ds}
\frac{k^2 l+\frac{2}{3}k^3}{3k^2 l+k^3}= \frac{l+\frac{2}{3}k}{3l+k}=\frac{2}{3+27/(9+6 \frac{k}{l})} <\frac{5}{12}
 \end{equation}
 of dominant flops, respectively, compared to the existing BLS.
When updating $\mathbf{W}_{{\bar{l}}}$ into $\mathbf{W}_{{\bar{l}} + {\bar{p}} }$,
it can be seen from Table \Rmnum{1} that
the proposed recursive and square-root BLS algorithms  both
require less flops than the existing BLS.
Moreover, it can be seen from Table \Rmnum{1} that compared with the proposed square-root BLS,
the proposed recursive  BLS
requires
  \begin{equation}\label{FlopRatioRecur2Sqrt2ewvew3}
\frac{2k p^2+2 k^2 p + 3k^3}{2k^2p + k^3}=\frac{p}{k}+0.5 + \frac{5}{4}\frac{1}{\frac{p}{k}+0.5} \ge  \frac{7}{3}
 \end{equation}
 times of flops  if $p \ge k$,
and requires the extra
 \begin{multline}\label{FlopRatioRecur2Sqrt1a3d21ds}
(3 p k^2 +3 p^2k+p^3)-(\frac{2}{3}k^3+3  p k^2+ 2p^2 k+p^3) \\ = p^2 k -\frac{2}{3}k^3
 \end{multline}
 flops if  $p < k$.
 Since (\ref{FlopRatioRecur2Sqrt1a3d21ds}) $>0$
 if $p > \sqrt {\frac{2}{3}} k \approx 0.82k$,
 we can combine (\ref{FlopRatioRecur2Sqrt2ewvew3}) and
(\ref{FlopRatioRecur2Sqrt1a3d21ds}) to conclude that
compared to the
proposed
recursive
 BLS,
   the
  proposed
    square-root BLS requires less flops when
    $p > 0.82k$, and requires more flops when
     $p < 0.82k$.


After adding  new inputs to the network, it may be required to
insert
new nodes when the training error threshold is not satisfied~\cite[Alg. 3]{BL_trans_paper}.
To insert nodes,
 we can use any of the efficient BLS algorithms
 based on the Cholesky factor in \cite{CholBLSdec2020} and \cite{CholBLSnodesSubmitted}, which are the generalized inverse and ridge solutions, respectively.
 In this case,
     the proposed recursive BLS
       requires additional $\frac{1}{3} k^3$ flops~\cite{Matrix_Computations_book}
     to compute ${{\mathbf{F}}_{\bar l}}$ from  ${{\mathbf{Q}}_{\bar l}}$ by the Cholesky factorization in (\ref{Q2PiPiT9686954}).
     Then after the Cholesky factor  ${{\mathbf{F}}}$ is updated to insert $i$ nodes by
     an efficient BLS algorithm
      on added nodes,
      additional $\frac{1}{3} (k+i)^3$ flops are  required to compute ${{\mathbf{Q}}}$ from  ${\mathbf{F}}$ by (\ref{Q2PiPiT9686954})  (i.e., ${{\mathbf{Q}}}={{\mathbf{F}}}{{\mathbf{F}}}^T$),
     when the proposed recursive BLS
      is utilized again to add new inputs.
      Then if nodes are inserted after each increment of inputs,
        the difference in flops between the proposed two BLS algorithms
      represented by
      (\ref{FlopRatioRecur2Sqrt1a3d21ds})  should be modified into
  \begin{equation}\label{PxiaoyukDifFlopsSpecialCase}
p^2 k -\frac{2}{3}k^3 + \frac{1}{3} k^3 + \frac{1}{3} (k+i)^3 >p^2 k,
 \end{equation}
 which shows that the proposed recursive BLS  always requires more flops than the proposed square-root BLS.

\begin{table*}[!t]
\scriptsize
\renewcommand{\arraystretch}{1.3}
\newcommand{\tabincell}[2]{\begin{tabular}{@{}#1@{}}#2\end{tabular}}
\caption{{Testing Accuracy of the BLS Algorithms on MNIST Dataset for Increment of Inputs and Nodes}} \label{table_example} \centering
\begin{tabular}{c|c|c|c|c|c|c |c | c | c| c|c |c | c| c |}
\hline
\multicolumn{4}{c|}{{\bfseries \tabincell{c}{ Number of  }}}    &\multirow{2}*{60}   &\multirow{2}*{60}    &     $\downarrow \scriptscriptstyle{10}$    &\multirow{2}*{70}          &    $\downarrow \scriptscriptstyle{10}$     & \multirow{2}*{80}      &   $\downarrow \scriptscriptstyle{10}$   &  \multirow{2}*{90}       &   $\downarrow \scriptscriptstyle{10}$     &  \multirow{2}*{100}       &    $\downarrow \scriptscriptstyle{10}$         \\
\multicolumn{4}{c|}{{\bfseries \tabincell{c}{Feature   Nodes}}}     &   &     &     70     &          &    80     &        &   90    &         &    100     &          &    110         \\
 \hline
\multicolumn{4}{c|}{{\bfseries \tabincell{c}{Number of }}}    &\multirow{2}*{11000}   &\multirow{2}*{11000}    &     $\downarrow \scriptscriptstyle{800}$     &\multirow{2}*{11800}        &    $\downarrow \scriptscriptstyle{800}$     &\multirow{2}*{12600}         &   $\downarrow \scriptscriptstyle{800} $    & \multirow{2}*{13400}        &    $\downarrow \scriptscriptstyle{800}$     &\multirow{2}*{14200}          &    $\downarrow \scriptscriptstyle{800}$        \\
\multicolumn{4}{c|}{{\bfseries \tabincell{c}{Enhancement   Nodes}}}     &   &     &     11800     &          &    12600     &        &   13400    &         &    14200     &          &    15000        \\
\hline
\multicolumn{4}{c|}{{\bfseries \tabincell{c}{ Number of  }}}    &\multirow{2}*{40000}   &  $\downarrow \scriptscriptstyle{4000}$   & \multirow{2}*{44000}        &    $\downarrow \scriptscriptstyle{4000}$      &\multirow{2}*{48000}          &   $\downarrow \scriptscriptstyle{4000}$     &  \multirow{2}*{52000}      &    $\downarrow \scriptscriptstyle{4000}$     & \multirow{2}*{56000}    &    $\downarrow \scriptscriptstyle{4000}$      &  \multirow{2}*{60000}          \\
\multicolumn{4}{c|}{{\bfseries \tabincell{c}{Input   Patterns}}}     &   &  44000    &          &    48000      &         &   52000     &       &    56000     &         &    60000      &             \\
\hline
\hline
 \multirow{46}*{\rotatebox{90}{{\bfseries   \tabincell{c}{Testing Accuracy ($\% $)   }} }} &  \multirow{6}*{\rotatebox{90}{{\bfseries   \tabincell{c}{  $\lambda= {{10}^{-8}}$  }} }}     & \multirow{2}*{\textbf{Exst.}}    &Mean      & 98.72  &   98.75  &   98.78  &   98.82  &   98.83  &   98.85  &   98.84  &   98.88  &   98.87  &   98.91  &   98.90         \\
\cdashline{4-15}[0.8pt/2pt]
   &  &       &  Std      &0.052  &    0.049  &    0.058  &    0.054  &    0.056  &    0.058  &    0.059  &    0.059  &    0.061  &    0.055  &    0.052 \\
\cline{3-15}
    &    & \multirow{2}*{\textbf{Sqrt.}}    &Mean      & 98.72  &   98.76  &   98.78  &   98.82  &   98.84  &   98.85  &   98.86  &   98.89  &   98.89  &   98.91  &   98.92    \\
 \cdashline{4-15}[0.8pt/2pt]
   &     &    &  Std    & 0.052  &    0.049  &    0.056  &    0.054  &    0.057  &    0.050  &    0.045  &    0.055  &    0.055  &    0.050  &    0.048         \\
\cdashline{3-15}
   &     & \multirow{2}*{\textbf{D-Rdg}}     &Mean      &98.72  &   98.76  &   98.78  &   98.82  &   98.84  &   98.85  &   98.86  &   98.89  &   98.89  &   98.92  &   98.92         \\
  \cdashline{4-15}[0.8pt/2pt]
    &    &     &  Std    &  0.052  &    0.048  &    0.056  &    0.054  &    0.056  &    0.051  &    0.046  &    0.055  &    0.052  &    0.050  &    0.049     \\
\cline{2-15}
\cline{2-15}
    &  \multirow{6}*{\rotatebox{90}{{\bfseries   \tabincell{c}{  $\lambda= {{10}^{-7}}$  }} }}     & \multirow{2}*{\textbf{Exst.}}    &Mean      &98.72  &   98.76  &   98.79  &   98.81  &   98.83  &   98.84  &   98.84  &   98.84  &   98.85  &   98.87  &   98.87       \\
\cdashline{4-15}[0.8pt/2pt]
   &  &       &  Std    &0.063  &    0.057  &    0.045  &    0.054  &    0.050  &    0.050  &    0.061  &    0.044  &    0.050  &    0.045  &    0.050       \\
\cline{3-15}
    &    & \multirow{2}*{\textbf{Sqrt.}}    &Mean      &  98.72  &   98.76  &   98.80  &   98.82  &   98.84  &   98.86  &   98.87  &   98.87  &   98.88  &   98.90  &   98.91        \\
 \cdashline{4-15}[0.8pt/2pt]
   &     &    &  Std    & 0.063  &    0.056  &    0.047  &    0.051  &    0.055  &    0.046  &    0.049  &    0.049  &    0.040  &    0.043  &    0.039          \\
\cdashline{3-15}
   &     & \multirow{2}*{\textbf{D-Rdg}}     &Mean      &  98.72  &   98.76  &   98.80  &   98.82  &   98.84  &   98.86  &   98.87  &   98.87  &   98.88  &   98.90  &   98.91         \\
  \cdashline{4-15}[0.8pt/2pt]
    &    &     &  Std    & 0.063  &    0.056  &    0.047  &    0.051  &    0.055  &    0.046  &    0.047  &    0.049  &    0.044  &    0.045  &    0.037       \\
\cline{2-15}
\cline{2-15}
    &  \multirow{6}*{\rotatebox{90}{{\bfseries   \tabincell{c}{  $\lambda= {{10}^{-6}}$  }} }}     & \multirow{2}*{\textbf{Exst.}}    &Mean      &98.54  &   98.57  &   98.62  &   98.67  &   98.71  &   98.72  &   98.75  &   98.76  &   98.77  &   98.79  &   98.79          \\
\cdashline{4-15}[0.8pt/2pt]
   &  &       &  Std    &0.064  &    0.071  &    0.063  &    0.055  &    0.053  &    0.059  &    0.058  &    0.054  &    0.058  &    0.055  &    0.066        \\
\cline{3-15}
    &    & \multirow{2}*{\textbf{Sqrt.}}    &Mean      & 98.54  &   98.58  &   98.62  &   98.68  &   98.71  &   98.73  &   98.77  &   98.77  &   98.79  &   98.80  &   98.82       \\
 \cdashline{4-15}[0.8pt/2pt]
   &     &    &  Std    & 0.064  &    0.074  &    0.065  &    0.058  &    0.057  &    0.057  &    0.052  &    0.054  &    0.055  &    0.050  &    0.049         \\
\cdashline{3-15}
   &     & \multirow{2}*{\textbf{D-Rdg}}     &Mean      &98.54  &   98.58  &   98.62  &   98.68  &   98.71  &   98.73  &   98.77  &   98.77  &   98.80  &   98.80  &   98.82        \\
  \cdashline{4-15}[0.8pt/2pt]
    &    &     &  Std    &0.064  &    0.074  &    0.065  &    0.058  &    0.057  &    0.057  &    0.052  &    0.054  &    0.055  &    0.050  &    0.049       \\
\cline{2-15}
\cline{2-15}
    &  \multirow{4}*{\rotatebox{90}{{\bfseries   \tabincell{c}{  $\lambda= {{10}^{-5}}$  }} }}     & \multirow{2}*{\textbf{Exst.}}    &Mean      & 98.27  &   98.28  &   98.35  &   98.40  &   98.45  &   98.46  &   98.51  &   98.52  &   98.55  &   98.57  &   98.59          \\
\cdashline{4-15}[0.8pt/2pt]
   &  &       &  Std    &0.087  &    0.080  &    0.087  &    0.074  &    0.075  &    0.072  &    0.073  &    0.066  &    0.069  &    0.066  &    0.058         \\
\cline{3-15}
   &     & \multirow{2}*{\textbf{Sqrt., D-Rdg}}     &Mean      &98.27  &   98.29  &   98.35  &   98.40  &   98.46  &   98.47  &   98.52  &   98.52  &   98.57  &   98.58  &   98.62        \\
  \cdashline{4-15}[0.8pt/2pt]
    &    &     &  Std    &0.087  &    0.081  &    0.080  &    0.081  &    0.072  &    0.072  &    0.069  &    0.069  &    0.060  &    0.064  &    0.052     \\
\cline{2-15}
\cline{2-15}
    &  \multirow{6}*{\rotatebox{90}{{\bfseries   \tabincell{c}{  $\lambda= {{10}^{-4}}$  }} }}     & \multirow{2}*{\textbf{Exst.}}    &Mean      &97.76  &   97.78  &   97.89  &   97.93  &   98.00  &   98.02  &   98.09  &   98.10  &   98.15  &   98.18  &   98.20         \\
\cdashline{4-15}[0.8pt/2pt]
   &  &       &  Std    &0.099  &    0.095  &    0.081  &    0.082  &    0.078  &    0.079  &    0.068  &    0.066  &    0.072  &    0.063  &    0.069          \\
\cline{3-15}
    &    & \multirow{2}*{\textbf{Sqrt., D-Rdg}}    &Mean      &97.76  &   97.80  &   97.89  &   97.95  &   98.02  &   98.03  &   98.11  &   98.12  &   98.17  &   98.20  &   \textbf{98.25}        \\
 \cdashline{4-15}[0.8pt/2pt]
   &     &    &  Std    &0.099  &    0.093  &    0.086  &    0.084  &    0.078  &    0.075  &    0.065  &    0.065  &    0.059  &    0.051  &    0.050           \\
\cdashline{3-15}
   &     & \multirow{2}*{\textbf{Recur.}}     &Mean      &97.76  &   97.80  &   97.39  &   97.53  &   96.78  &   97.09  &   96.02  &   96.67  &   95.32  &   96.27  &   \textbf{94.71}         \\
  \cdashline{4-15}[0.8pt/2pt]
    &    &     &  Std    &0.099  &    0.093  &    0.126  &    0.112  &    0.271  &    0.203  &    0.472  &    0.303  &    0.621  &    0.394  &    0.758       \\
\cline{2-15}
\cline{2-15}
    &  \multirow{6}*{\rotatebox{90}{{\bfseries   \tabincell{c}{  $\lambda= {{10}^{-3}}$  }} }}     & \multirow{2}*{\textbf{Exst.}}    &Mean      &97.05  &   97.06  &   97.22  &   97.24  &   97.36  &   97.37  &   97.46  &   97.47  &   97.50  &   97.56  &   97.58          \\
\cdashline{4-15}[0.8pt/2pt]
   &  &       &  Std    &0.174  &    0.166  &    0.143  &    0.140  &    0.116  &    0.123  &    0.114  &    0.103  &    0.115  &    0.111  &    0.107         \\
\cline{3-15}
    &    & \multirow{2}*{\textbf{Sqrt., D-Rdg}}    &Mean      &97.05  &   97.08  &   97.22  &   97.25  &   97.36  &   97.40  &   97.48  &   97.49  &   97.56  &   97.59  &   \textbf{97.65}      \\
 \cdashline{4-15}[0.8pt/2pt]
   &     &    &  Std    &0.174  &    0.163  &    0.140  &    0.144  &    0.123  &    0.121  &    0.098  &    0.093  &    0.096  &    0.100  &    0.092        \\
\cdashline{3-15}
   &     & \multirow{2}*{\textbf{Recur.}}     &Mean      &97.05  &   97.08  &   97.21  &   97.24  &   97.35  &   97.39  &   97.46  &   97.47  &   97.53  &   97.57  &   \textbf{97.62}       \\
  \cdashline{4-15}[0.8pt/2pt]
    &    &     &  Std    &0.174  &    0.163  &    0.144  &    0.143  &    0.122  &    0.122  &    0.106  &    0.105  &    0.111  &    0.108  &    0.102      \\
\cline{2-15}
\cline{2-15}
    &  \multirow{6}*{\rotatebox{90}{{\bfseries   \tabincell{c}{  $\lambda= {{10}^{-2}}$  }} }}     & \multirow{2}*{\textbf{Exst.}}    &Mean      &95.80  &   95.83  &   96.06  &   96.08  &   96.26  &   96.26  &   96.37  &   96.40  &   96.48  &   96.55  &   96.58        \\
\cdashline{4-15}[0.8pt/2pt]
   &  &       &  Std    &0.230  &    0.228  &    0.170  &    0.168  &    0.136  &    0.135  &    0.153  &    0.140  &    0.162  &    0.140  &    0.159          \\
\cline{3-15}
    &    & \multirow{2}*{\textbf{Sqrt., D-Rdg}}    &Mean      &95.80  &   95.87  &   96.07  &   96.12  &   96.26  &   96.28  &   96.41  &   96.46  &   96.56  &   96.60  &   96.69     \\
 \cdashline{4-15}[0.8pt/2pt]
   &     &    &  Std    & 0.230  &    0.216  &    0.159  &    0.171  &    0.133  &    0.139  &    0.124  &    0.121  &    0.110  &    0.118  &    0.113        \\
\cdashline{3-15}
   &     & \multirow{2}*{\textbf{Recur.}}     &Mean      &95.80  &   95.87  &   96.07  &   96.12  &   96.26  &   96.28  &   96.41  &   96.46  &   96.56  &   96.60  &   96.69      \\
  \cdashline{4-15}[0.8pt/2pt]
    &    &     &  Std    &0.230  &    0.216  &    0.159  &    0.171  &    0.133  &    0.139  &    0.125  &    0.123  &    0.110  &    0.119  &    0.114   \\
\cline{2-15}
\cline{2-15}
    &  \multirow{6}*{\rotatebox{90}{{\bfseries   \tabincell{c}{  $\lambda= {{10}^{-1}}$  }} }}     & \multirow{2}*{\textbf{Exst.}}    &Mean      &94.18  &   94.22  &   94.54  &   94.56  &   94.79  &   94.80  &   94.94  &   94.98  &   95.05  &   95.12  &   95.14      \\
\cdashline{4-15}[0.8pt/2pt]
   &  &       &  Std    &0.234  &    0.237  &    0.214  &    0.209  &    0.169  &    0.166  &    0.166  &    0.162  &    0.161  &    0.152  &    0.162         \\
\cline{3-15}
    &    & \multirow{2}*{\textbf{Sqrt., D-Rdg}}    &Mean      &94.18  &   94.28  &   94.57  &   94.63  &   94.83  &   94.88  &   95.02  &   95.08  &   95.21  &   95.25  &   95.38       \\
 \cdashline{4-15}[0.8pt/2pt]
   &     &    &  Std    &0.234  &    0.236  &    0.207  &    0.210  &    0.167  &    0.168  &    0.143  &    0.140  &    0.147  &    0.145  &    0.122         \\
\cdashline{3-15}
   &     & \multirow{2}*{\textbf{Recur.}}     &Mean      &94.18  &   94.28  &   94.57  &   94.63  &   94.83  &   94.88  &   95.02  &   95.08  &   95.21  &   95.25  &   95.38     \\
  \cdashline{4-15}[0.8pt/2pt]
    &    &     &  Std    &0.234  &    0.236  &    0.207  &    0.210  &    0.168  &    0.168  &    0.142  &    0.140  &    0.147  &    0.145  &    0.121      \\
\hline
\end{tabular}
\end{table*}

\subsection{Numerical Experiments on  MNIST and NORB Datasets}

In this subsection,
  the experimental results
on the Modified National Institute of Standards and Technology (MNIST)
dataset~\cite{61_dataSet} and the
NYU
object recognition
benchmark (NORB) dataset~\cite{Norb_dataSet} will be given in Tables  \Rmnum{2}-\Rmnum{6}
and Table  \Rmnum{7}, respectively, where  \textbf{Exst.}, \textbf{Recur.}, \textbf{Sqrt.} and \textbf{D-Rdg} denote the abbreviations
of  the existing BLS on added inputs,
 the proposed recursive BLS, the proposed square-root BLS   and
the direct ridge  solution (by (\ref{xWbarMN2AbarYYa1341})  and
(\ref{xAmnAmnTAmnIAmnT231413})),
 respectively.
  The MNIST handwritten digital images
include $60000$ training samples and $10000$ testing samples.
On the other hand, the NORB dataset
 contains $48600$ images
  belonging to five distinct categories: (1) animals; (2) humans; (3) airplanes; (4) trucks; and (5) cars.
 Of these images, half are the training samples,
  and the other half are the test samples.

\begin{table*}[!t]
\scriptsize
\renewcommand{\arraystretch}{1.3}
\newcommand{\tabincell}[2]{\begin{tabular}{@{}#1@{}}#2\end{tabular}}
\caption{Testing Accuracy of the BLS Algorithms   on NORB Dataset with $p=1600<k=15000$} \label{table_example} \centering
\begin{tabular}{c|c|c||c|c|c|c|c|c|}
\hline
\multicolumn{3}{c||}{\bfseries  Number of Inputs }   &{{\bfseries   {16300}}}   &{{\bfseries   {$\xrightarrow[\scriptscriptstyle{1600}]{}$ 17900}}}   &{{\bfseries   {$\xrightarrow[\scriptscriptstyle{1600}]{}$ 19500}}}   &{{\bfseries   {$\xrightarrow[\scriptscriptstyle{1600}]{}$ 21100}}}   &{{\bfseries   {$\xrightarrow[\scriptscriptstyle{1600}]{}$ 22700}}}   &{{\bfseries   {$\xrightarrow[\scriptscriptstyle{1600}]{}$ 24300}}}   \\
\hline
\multirow{24}*{\rotatebox{90}{{\bfseries   \tabincell{c}{Testing Accuracy ($\% $) }} }} & \multirow{4}*{\rotatebox{90}{{\bfseries   \tabincell{c}{ $\lambda=$ \\ ${{10}^{-8}}$}} }}    & \textbf{Exst.}   &85.00 ($\pm$      0.890) &    85.88 ($\pm$      0.721) &    86.68 ($\pm$      0.717) &    87.04 ($\pm$      0.691) &    87.45 ($\pm$      0.646) &    87.70 ($\pm$      0.592)	 \\
\cline{3-9}
& &  \textbf{Recur.}    &  85.00 ($\pm$      0.890) &    85.54 ($\pm$      0.819) &    86.32 ($\pm$      0.770) &    86.72 ($\pm$      0.675) &    87.18 ($\pm$      0.624) &    87.49 ($\pm$      0.593)	 	 \\
\cdashline{3-9}
& &  \textbf{Sqrt.}     &  85.00 ($\pm$      0.890) &    85.54 ($\pm$      0.814) &    86.32 ($\pm$      0.769) &    86.72 ($\pm$      0.678) &    87.18 ($\pm$      0.627) &    87.49 ($\pm$      0.592)	 \\
\cdashline{3-9}
& &  \textbf{D-Rdg}    & 85.00 ($\pm$      0.890) &    85.54 ($\pm$      0.814) &    86.32 ($\pm$      0.769) &    86.72 ($\pm$      0.678) &    87.18 ($\pm$      0.626) &    87.49 ($\pm$      0.592) 		\\
\cline{2-9}
  & \multirow{4}*{\rotatebox{90}{{\bfseries   \tabincell{c}{ $\lambda=$ \\ ${{10}^{-7}}$}} }}    &\textbf{Exst.}   & 87.84 ($\pm$      0.654) &    88.11 ($\pm$      0.603) &    88.39 ($\pm$      0.536) &    88.45 ($\pm$      0.532) &    88.58 ($\pm$      0.506) &    88.66 ($\pm$      0.507) 	 \\
\cline{3-9}
& &  \textbf{Recur.}    & 87.84 ($\pm$      0.654) &    88.00 ($\pm$      0.615) &    88.24 ($\pm$      0.570) &    88.20 ($\pm$      0.593) &    88.36 ($\pm$      0.553) &    88.37 ($\pm$      0.564)	 	 \\
\cdashline{3-9}
& &  \textbf{Sqrt.}     & 87.84 ($\pm$      0.654) &    88.00 ($\pm$      0.614) &    88.25 ($\pm$      0.570) &    88.20 ($\pm$      0.593) &    88.36 ($\pm$      0.553) &    88.37 ($\pm$      0.563)		 \\
\cdashline{3-9}
& &  \textbf{D-Rdg}    & 87.84 ($\pm$      0.654) &    88.00 ($\pm$      0.614) &    88.25 ($\pm$      0.570) &    88.20 ($\pm$      0.593) &    88.36 ($\pm$      0.552) &    88.37 ($\pm$      0.564)		\\
\cline{2-9}
   & \multirow{4}*{\rotatebox{90}{{\bfseries   \tabincell{c}{ $\lambda=$ \\ ${{10}^{-6}}$}} }}    & \textbf{Exst.}  & 89.05 ($\pm$      0.405) &    89.11 ($\pm$      0.395) &    89.23 ($\pm$      0.377) &    89.25 ($\pm$      0.362) &    89.27 ($\pm$      0.387) &    89.28 ($\pm$      0.379)	 \\
\cline{3-9}
& &  \textbf{Recur.}    & 89.05 ($\pm$      0.405) &    89.10 ($\pm$      0.395) &    89.27 ($\pm$      0.376) &    89.26 ($\pm$      0.386) &    89.28 ($\pm$      0.411) &    89.28 ($\pm$      0.397) 	 \\
\cdashline{3-9}
& &  \textbf{Sqrt.}     &89.05 ($\pm$      0.405) &    89.10 ($\pm$      0.395) &    89.27 ($\pm$      0.376) &    89.26 ($\pm$      0.386) &    89.28 ($\pm$      0.411) &    89.28 ($\pm$      0.397) 		 \\
\cdashline{3-9}
& &  \textbf{D-Rdg}    &89.05 ($\pm$      0.405) &    89.10 ($\pm$      0.395) &    89.27 ($\pm$      0.377) &    89.26 ($\pm$      0.386) &    89.28 ($\pm$      0.411) &    89.28 ($\pm$      0.397) 		\\
\cline{2-9}
 & \multirow{3}*{\rotatebox{90}{{\bfseries   \tabincell{c}{ $\lambda=$ \\ ${{10}^{-5}}$}} }}    & \textbf{Exst.}   &89.46 ($\pm$      0.323) &    89.53 ($\pm$      0.301) &    89.61 ($\pm$      0.285) &    89.63 ($\pm$      0.298) &    89.63 ($\pm$      0.296) &    89.62 ($\pm$      0.313)		 \\
\cline{3-9}
& &  \textbf{Recur.}    & 89.46 ($\pm$      0.323) &    89.54 ($\pm$      0.292) &    89.68 ($\pm$      0.292) &    89.69 ($\pm$      0.282) &    89.70 ($\pm$      0.296) &    89.67 ($\pm$      0.325)	 		 \\
\cdashline{3-9}
& &  \textbf{Sqrt., D-Rdg}    &89.46 ($\pm$      0.323) &    89.54 ($\pm$      0.292) &    89.68 ($\pm$      0.292) &    89.69 ($\pm$      0.282) &    89.70 ($\pm$      0.296) &    89.67 ($\pm$      0.325) 	  		 \\
\cline{2-9}
 & \multirow{3}*{\rotatebox{90}{{\bfseries   \tabincell{c}{ $\lambda=$ \\ ${{10}^{-4}}$}} }}    & \textbf{Exst.}   &89.76 ($\pm$      0.237) &    89.83 ($\pm$      0.247) &    89.83 ($\pm$      0.238) &    89.84 ($\pm$      0.247) &    89.85 ($\pm$      0.258) &    89.82 ($\pm$      0.256)  		 \\
\cline{3-9}
& &  \textbf{Recur.}    & 89.76 ($\pm$      0.237) &    89.85 ($\pm$      0.247) &    89.90 ($\pm$      0.231) &    89.91 ($\pm$      0.246) &    89.95 ($\pm$      0.250) &    89.92 ($\pm$      0.253) 	 		 \\
\cdashline{3-9}
& &  \textbf{Sqrt., D-Rdg}    &89.76 ($\pm$      0.237) &    89.85 ($\pm$      0.247) &    89.90 ($\pm$      0.231) &    89.91 ($\pm$      0.246) &    89.95 ($\pm$      0.250) &    89.92 ($\pm$      0.253) 	 		 \\
\cline{2-9}
 & \multirow{2}*{\rotatebox{90}{{\bfseries   \tabincell{c}{ $\lambda=$ \\ ${{10}^{-3}}$}} }}    & \textbf{Exst.}   & 90.17 ($\pm$      0.206) &    90.22 ($\pm$      0.195) &    90.24 ($\pm$      0.219) &    90.23 ($\pm$      0.230) &    90.24 ($\pm$      0.232) &    \textbf{90.24} ($\pm$      0.223)		 \\
\cline{3-9}
& &  \textbf{Recur., Sqrt., D-Rdg}    &90.17 ($\pm$      0.206) &    90.25 ($\pm$      0.193) &    90.29 ($\pm$      0.208) &    90.29 ($\pm$      0.231) &   90.34 ($\pm$      0.250) &    \textbf{90.34} ($\pm$      0.236) 	   		 \\
\cline{2-9}
 & \multirow{2}*{\rotatebox{90}{{\bfseries   \tabincell{c}{ $\lambda=$ \\ ${{10}^{-2}}$}} }}    & \textbf{Exst.}   & 89.66 ($\pm$      0.282) &    89.66 ($\pm$      0.277) &    89.70 ($\pm$      0.266) &    89.69 ($\pm$      0.274) &    89.71 ($\pm$      0.263) &    89.67 ($\pm$      0.268)		 \\
\cline{3-9}
& &  \textbf{Recur., Sqrt., D-Rdg}    &89.66 ($\pm$      0.282) &    89.74 ($\pm$      0.267) &    89.85 ($\pm$      0.250) &    89.87 ($\pm$      0.248) &    89.92 ($\pm$      0.258) &    89.92 ($\pm$      0.263) 	   		 \\
\cline{2-9}
 & \multirow{2}*{\rotatebox{90}{{\bfseries   \tabincell{c}{ $\lambda=$ \\ ${{10}^{-1}}$}} }}    & \textbf{Exst.}   & 87.45 ($\pm$      0.605) &    87.49 ($\pm$      0.592) &    87.53 ($\pm$      0.588) &    87.53 ($\pm$      0.594) &    87.56 ($\pm$      0.581) &    87.49 ($\pm$      0.590) 		 \\
\cline{3-9}
& &  \textbf{Recur., Sqrt., D-Rdg}    & 87.45 ($\pm$      0.605) &    87.58 ($\pm$      0.581) &    87.73 ($\pm$      0.579) &    87.80 ($\pm$      0.577) &    87.90 ($\pm$      0.554) &    87.88 ($\pm$      0.540) 		 \\
\hline
\end{tabular}
\end{table*}

\begin{figure*}[!t]
    \centering
    \subfloat[MNIST
dataset.]{
        \hspace*{-.1in}
        \includegraphics[scale=0.56]{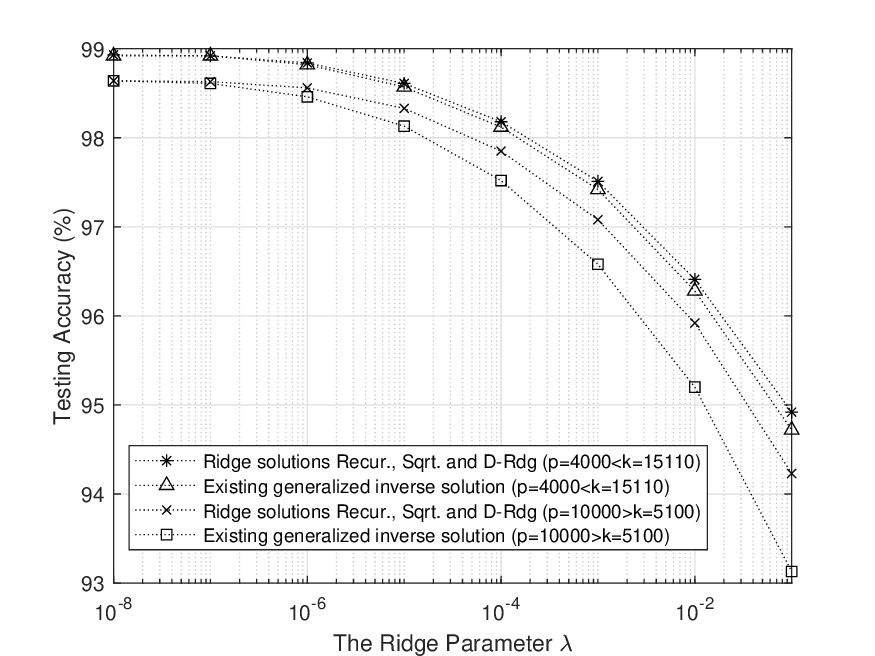}
        \label{p_D_req_100}
    }
        \subfloat[NORB dataset.]{
        \hspace*{-.1in}
        \includegraphics[scale=0.56]{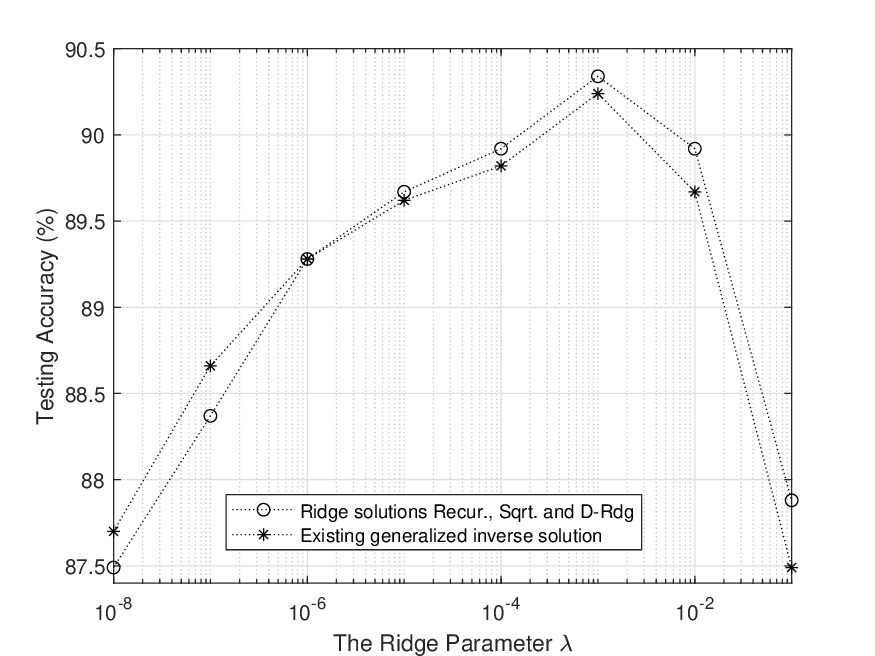}
        \label{p_D_req_100}
    }
\caption{The maximum testing accuracies achieved under different ridge parameters on MNIST and NORB datasets.}
\label{zhftry1}
\end{figure*}






Regarding  the simulations on the MNIST dataset for Tables  \Rmnum{2}-\Rmnum{6},
 the relevant networks and increments of inputs are as follows.
The simulations for Table  \Rmnum{5} of \cite{BL_trans_paper} are followed
 by those for
  Tables  \Rmnum{2} and \Rmnum{4},
 where
 the network is constructed by
  $k=5100$ nodes including  $10 \times 10$ feature nodes and $5000$ enhancement nodes,
and  $p=10000>k$
training samples are added
 in each of the 5 update.
In our simulations  for Tables  \Rmnum{3} and  \Rmnum{5},
we construct the network by
 $k=15110$ nodes including  $10 \times 11$ feature nodes and $15000$ enhancement nodes,
and add $p=4000<k$
training samples
 in each of the 5 update.
Moreover,
as Table  \Rmnum{6} in \cite{BL_trans_paper},
Table  \Rmnum{6} gives the  experimental results for the increment
of inputs and nodes. In the simulations for Table  \Rmnum{6},
 we utilize  40000 training samples to train
 the initial network with $10 \times 6$
feature nodes and $11000$ enhancement nodes,
and in each of the 5 update, we increase 4000 inputs,
and then increase 810 nodes including $10$ feature nodes, $300$ enhancement nodes only corresponding to the added feature nodes,
and  $500$ extra enhancement nodes.

Tables \Rmnum{2} and \Rmnum{3} give the testing accuracies~\footnote{In all the tables of this paper, the testing accuracies  are the mean and standard deviation of 100 simulations.} of the presented BLS algorithms,
 for
 the ridge parameter $\lambda={{10}^{-8}},{{10}^{-7}},\cdots,{{10}^{-1}}$.
 The simulations for Table \Rmnum{2} show that
\textbf{Sqrt.}
and
\textbf{Recur.}
always achieve the same  testing accuracy
  as \textbf{D-Rdg} when
   $\lambda \ge {{10}^{-7}}$
   and
   $\lambda \ge {{10}^{-5}}$,
  respectively,
 and
 the simulations for Table \Rmnum{3} show that
 \textbf{Sqrt.},  \textbf{Recur.} and \textbf{D-Rdg} always achieve
 the same  testing accuracy when  $\lambda \ge {{10}^{-5}}$.
Then for simplicity, the same testing accuracy achieved by  multiple
 BLS algorithms is listed only once in Tables \Rmnum{2} and \Rmnum{3}.
 As observed from Tables \Rmnum{2} and \Rmnum{3},
 both the proposed ridge solutions for BLS basically achieve the testing accuracy of the direct ridge solution,
 and improve the testing accuracy of the existing BLS
   when $\lambda \ge 10^{-7}$ in Table \Rmnum{2} and $\lambda \ge 10^{-6}$ in Table \Rmnum{3}.
  The above-mentioned improvement becomes more significant as  $\lambda$ is bigger.

Tables \Rmnum{4} and \Rmnum{5}  show the training time of the existing BLS and the proposed recursive and square-root algorithms,
which is the average value of 500 simulations.  In Tables \Rmnum{4} and \Rmnum{5},
 the speedups are
 ${T_{existing}}/{T_{proposed}}$, i.e.,
the  ratio between the training time of the existing BLS   and that of the proposed algorithm.
 Table \Rmnum{4} shows that when $p>k$,  the speedups of \textbf{Recur.} and  \textbf{Sqrt.}
   over \textbf{Exst.}
in  each additional  training time
   are $1.99 \sim 7.40$
and $3.22 \sim 11.87$, respectively, and the speedups in total
training time
are $4.41$
and $6.92$, respectively.  On the other hand,
 Table \Rmnum{5} shows that when $p<k$,  the speedups of \textbf{Recur.} and  \textbf{Sqrt.}
   over \textbf{Exst.}
 in  each additional  training time
  are $3.28 \sim 4.37$
and $1.30 \sim 1.75$, respectively,
 and the speedups in total
training time
are $2.80$
and $1.59$, respectively.
Obviously both the proposed BLS algorithms significantly accelerate the
  existing BLS, while compared to the proposed recursive BLS,
    the proposed square-root BLS    is faster  when $p>k$, and is slower when $p<k$.

   To increase inputs and nodes in our simulations for Table \Rmnum{7},  \textbf{Exst.}
  is applied with the original BLS on added nodes~\cite[Alg. 3]{BL_trans_paper},
  while both the proposed  ridge solutions
    are applied with
   the efficient ridge solution for the BLS on
added nodes~\cite{CholBLSnodesSubmitted} to obtain the complete ridge solution. As mentioned in the last subsection,
when \textbf{Recur.} is applied with
the efficient BLS in  \cite{CholBLSnodesSubmitted} that is based on the inverse
Cholesky factor,
(\ref{Q2PiPiT9686954}) (i.e., ${{\mathbf{Q}}}={{\mathbf{F}}}{{\mathbf{F}}}^T$)
is utilized
 to
compute ${{\mathbf{F}}}$ from  ${{\mathbf{Q}}}$ by the Cholesky factorization and compute ${{\mathbf{Q}}}$ from  ${\mathbf{F}}$,
which causes extra numerical errors. When $\lambda \le 10^{-5}$,
  those extra numerical errors make
${{\mathbf{Q}}}$  no longer
positive definite  and causes  the  Cholesky
factor of ${{\mathbf{Q}}}$ to be unavailable.
Then    Table  \Rmnum{6} has not given any testing accuracy for  \textbf{Recur.}
when $\lambda \le 10^{-5}$. Moreover,
 Table  \Rmnum{6} shows that when $\lambda=10^{-4}$  or  $\lambda=10^{-3}$,
\textbf{Recur.}  achieves worse testing accuracies than \textbf{Sqrt.}, which can also be explained by the above-mentioned
extra numerical errors.
Thus we can conclude that when
$\lambda$ is small
 and it is required to work with  the efficient ridge solution based on the inverse Cholesky factor for the BLS on added nodes~\cite{CholBLSnodesSubmitted}
for the increment of inputs and nodes,
    the proposed square-root BLS  is more
    suitable than the proposed recursive BLS, to avoid the loss in testing accuracy caused by numerical errors.




    Lastly,  Table \Rmnum{7} gives the testing accuracies of the presented
BLS algorithms on the NORB dataset.
In the simulations,
we set  the network
as
 $k=15000$ nodes including
 $100 \times 10$
feature nodes and $14000$ enhancement nodes,
and add
 $p=1600<k$ training samples in each of the 5 updates,
  to increase the inputs from the initial 16300 to the final 24300.
 As shown in Table \Rmnum{7},
    both the proposed ridge solutions for BLS basically achieve the testing accuracy of the direct ridge solution,
and improve the testing accuracy of the existing BLS when $\lambda \ge 10^{-5}$.
The above-mentioned improvement becomes more significant as  $\lambda$ is bigger, as in Tables \Rmnum{2} and \Rmnum{3}.


\subsection{Analysis of the Maximum Testing Accuracies Achieved under Different Ridge Parameters}


Fig. 1 shows the maximum testing accuracies achieved by the BLS algorithms under different ridge parameters on the MNIST and NORB datasets, which come from
the accuracies in Tables \Rmnum{2}, \Rmnum{3} and \Rmnum{7}.
Since the ridge solutions \textbf{Sqrt.},
\textbf{Recur.}
and \textbf{D-Rdg}
always achieve the same maximum  testing accuracies,
only the accuracies for the existing generalized inverse solution and the ridge solutions
are listed in Fig. 1.


Tables \Rmnum{2},  \Rmnum{3} and Fig. 1.a show that
 on the MNIST dataset,
 the existing generalized inverse solution and the
  ridge solutions
 achieve nearly the same
      maximum testing accuracy when $\lambda = {{10}^{-8}}$.
    Specifically,
    Table \Rmnum{3}  shows that the
     ridge solutions and the existing BLS achieve the  maximum testing accuracies
     of $98.93\%$ and $98.92\%$, respectively, and the former is a little better than the latter.
 On the other hand,   Tables \Rmnum{7} and Fig. 1.b show
that on the NORB dataset,
the
  ridge solutions and the existing BLS achieve the  maximum testing accuracies
of $90.34\%$ and $90.24\%$, respectively, when $\lambda = {{10}^{-3}}$.
Thus it can be concluded that compared with  the existing generalized inverse solution,
the proposed two ridge solutions achieve better maximum testing accuracies  when the corresponding ridge parameter $\lambda$ is
 not too small, and achieve nearly the same maximum testing accuracy when the corresponding ridge parameter $\lambda$ is
 very small.

 \section{Conclusion}

  To
 improve the existing BLS on  added inputs,
 this paper proposes the recursive and square-root BLS algorithms,
 which
   utilize the inverse and inverse Cholesky factor of
  the
   Hermitian matrix in the ridge inverse, respectively,
  to update the ridge solution.
 The  recursive BLS  utilizes the matrix inversion lemma
  to
    update
  the inverse,
     while the  square-root BLS
 updates the upper-triangular  inverse Cholesky factor
 by multiplying
 it with an upper-triangular intermediate matrix.
  When
   the added $p$
   inputs
    are more than the total
$k$ nodes in the network,
i.e., $p>k$,
   the inverse of a sum of matrices
    is utilized to
    take
    a smaller
   matrix inversion
   or inverse Cholesky factorization.
For the distributed BLS with data-parallelism, we develop
  the parallel implementation of
the  square-root BLS, which is deduced by applying the parallel implementation of the inverse Cholesky factorization.


The existing BLS is based on  the generalized
inverse with the ridge regression,
which assumes
the ridge parameter $\lambda \to 0$ in the ridge inverse.
When $\lambda \to 0$ is not satisfied,
the   numerical experiments on the MNIST and NORB datasets show that
both the proposed ridge solutions
improve the testing accuracy of the existing generalized inverse solution,
and the improvement becomes more significant as  $\lambda$ is bigger.
Moreover, compared to the existing BLS,
the proposed two ridge solutions achieve better and nearly the same  maximum testing accuracies, respectively,  when the corresponding ridge parameter does not satisfies
and satisfies  $\lambda \to 0$.
On the MNIST dataset when
$\lambda = {{10}^{-8}}$, the maximum testing accuracies achieved by the proposed two ridge solutions  and the existing  BLS
are $98.64\%$  in the case of $p>k$,  and are $98.93\%$ and $98.92\%$ respectively in the case of $p<k$, while
they are $90.34\%$ and $90.24\%$  respectively on the NORB dataset when $\lambda = {{10}^{-3}}$.

With respect to the existing BLS, both the proposed BLS algorithms theoretically require less flops,
and are significantly faster in the numerical experiments on the MNIST dataset.
 When $p>k$,  the speedups  in  each additional  training time
  of the proposed  recursive and square-root BLS algorithms over the existing BLS are $1.99 \sim 7.40$
and $3.22 \sim 11.87$, respectively, and the speedups in total
training time  are $4.41$
and $6.92$, respectively.  When $p<k$,  the speedups  in  each additional  training time
  of  the  recursive and square-root BLS algorithms over the existing BLS are $3.28 \sim 4.37$
and $1.30 \sim 1.75$, respectively,
 and the speedups in total
training time  are $2.80$
and $1.59$, respectively.

  Compared to the  recursive BLS,
   the  square-root BLS requires less flops
   when
    $p > 0.82k$,
   and requires more flops when $p < 0.82k$.
If  nodes are inserted after each increment of inputs,
both the proposed BLS algorithms
 are applied with
the efficient ridge solution  for the BLS on
added nodes that is based on the inverse
Cholesky factor,
 to obtain the complete ridge solution.
In this case, the inverse of the Hermitian matrix
utilized in the  recursive BLS
is Cholesky-factorized
 and then obtained again by
  multiplying the updated Cholesky factor with its transpose.
  The corresponding extra computations
   make the  recursive BLS always require more flops than the  square-root BLS,
and cause the numerical errors. When $\lambda$ is small in the simulations, those numerical errors  reduce
the testing accuracy,
 or even make the  recursive BLS
unworkable.
  As a comparison,  the  square-root BLS does not need any extra computations and  basically achieves the
  testing accuracy of the direct ridge solution, since it is based on
  the Cholesky factor just like the efficient
  ridge solution  for the BLS on
added nodes.

\appendices

\section{The Derivation of (\ref{xA2QAt13413Deduce121a4})}

Substitute
(\ref{BmatrixFromMyQ322aForQ})
and
(\ref{AxInputIncrease31232})
into
(\ref{xA2QAt13413}) to obtain
\begin{multline}\label{xA2QAt13413Deduce121a2}
{{\bf{A}}_{\bar l + \bar p}^{\dagger }}  =({{\mathbf{Q}}_{l}}-{{\mathbf{\tilde B}}} \mathbf{A}_{ p}^{{}}{{\mathbf{Q}}_{l}}) \left[ \begin{matrix}
{\bf{A}}_{l}^T    &  {\bf{A}}_{ p}^{T}  \\
\end{matrix} \right]    \\
=  \left[ \begin{matrix}
 {{\mathbf{Q}}_{\bar l}}{\bf{A}}_{\bar l}^T-{{\mathbf{\tilde B}}} \mathbf{A}_{ \bar p}^{{}}{{\mathbf{Q}}_{\bar l}}{\bf{A}}_{\bar l}^T   &  {{\mathbf{Q}}_{\bar l}}{\bf{A}}_{ \bar p}^{T} -{{\mathbf{\tilde B}}} \mathbf{A}_{ \bar p}^{{}}{{\mathbf{Q}}_{\bar l}}{\bf{A}}_{ \bar p}^{T}  \\
\end{matrix} \right].
\end{multline}
Then substitute (\ref{BmatrixFromMyQ322a}) (i.e.,   ${{\mathbf{\tilde B}}}  = {{\mathbf{Q}}_{\bar l}}\mathbf{A}_{ \bar p}^{T}{{(\mathbf{I}+\mathbf{A}_{ \bar p}^{{}}{{\mathbf{Q}}_{\bar l}}\mathbf{A}_{ \bar p}^{T})}^{-1}}$) into the last entry in the right side of (\ref{xA2QAt13413Deduce121a2}) to simplify it into
\begin{align}
& {{\mathbf{Q}}_{\bar l}}{\bf{A}}_{ \bar p}^{T} -{{\mathbf{\tilde B}}} \mathbf{A}_{ \bar p}^{{}}{{\mathbf{Q}}_{\bar l}}{\bf{A}}_{ \bar p}^{T}  \notag \\
& = {{\mathbf{Q}}_{\bar l}}{\bf{A}}_{ \bar p}^{T} - {{\mathbf{Q}}_{\bar l}}\mathbf{A}_{ \bar p}^{T}{{(\mathbf{I}+\mathbf{A}_{ \bar p}^{{}}
{{\mathbf{Q}}_{\bar l}}\mathbf{A}_{ \bar p}^{T})}^{-1}}\mathbf{A}_{ \bar p}^{{}}{{\mathbf{Q}}_{\bar l}}{\bf{A}}_{ \bar p}^{T}   \notag \\
& ={{\mathbf{Q}}_{\bar l}}{\bf{A}}_{ \bar p}^{T} - {{\mathbf{Q}}_{\bar l}}\mathbf{A}_{ \bar p}^{T}(\mathbf{I}-{{(\mathbf{I}+\mathbf{A}_{ \bar p}^{{}}
{{\mathbf{Q}}_{\bar l}}\mathbf{A}_{ \bar p}^{T})}^{-1}})  \notag \\
& = {{\mathbf{Q}}_{\bar l}}\mathbf{A}_{ \bar p}^{T} {{(\mathbf{I}+\mathbf{A}_{ \bar p}^{{}}
{{\mathbf{Q}}_{\bar l}}\mathbf{A}_{ \bar p}^{T})}^{-1}} \notag \\
& = {{\mathbf{\tilde B}}},  \label{xA2QAt13413Deduce121a5}
\end{align}
where (\ref{BmatrixFromMyQ322a}) is applied to get the last row.
Finally,
we
 substitute (\ref{xA2QAt13413Deduce121a5}) and
(\ref{xA2QAt13413})
(i.e., ${{\bf{A}}_{\bar l }^{\dagger }}={{\mathbf{Q}}_{\bar l}}{{\bf{A}}_{\bar l }^{T}}$)
 into
 (\ref{xA2QAt13413Deduce121a2})
  to get  (\ref{xA2QAt13413Deduce121a4}).

\section{Parallel Implementation of Proposed Square-Root BLS Algorithm Based on Inverse
Cholesky Factor for the
Distributed BLS with Data-Parallelism}

In this appendix, we use $K$ instead of  $k$
to denote the total number of feature and enhancement nodes.
We will describe the parallel implementations 
to compute
${\mathbf{F}}_{\bar l} \in {\Re ^{K \times K}}$
by (\ref{Q2PiPiT9686954})
and compute
 ${{\mathbf{F}}_{{\bar{l}}+\bar{p}}} \in {\Re ^{K \times K}}$
 by  (\ref{K2AxLm94835}),
 (\ref{Lwave2IKK40425aBefore}),
and  (\ref{Lbig2LLwave59056}),
which include $K$ iterations. In the $k^{th}$ ($k=1,2,\cdots,K$) iteration,
the $k \times k$ leading principal sub-matrices in
 the upper-triangular ${\mathbf{F}}_{\bar l} $ and ${{\mathbf{F}}_{{\bar{l}}+\bar{p}}}$
 are computed.


\subsection{The Considered Distributed BLS with Data-Parallelism and the Corresponding Square-Root BLS}


In
 the considered distributed
BLS with data-parallelism,   training samples
are partitioned into $\tau$ workers. Worker $i$ ($i = 1,2,\cdots,\tau$)
possesses ${l _i}$ training samples that can be denoted as
${{\bf{A}}^{{l_i}}} \in {\Re ^{l_i \times K}} $.

In
 the above-mentioned distributed
BLS with data-parallelism,
we can apply the proposed  square-root BLS algorithm based on inverse Cholesky factor
introduced in \textbf{Algorithms 5} and
\textbf{6}.
Accordingly,
 worker $i$ ($i = 1,2,\cdots,\tau$)
needs to
 compute the upper-triangular  inverse Cholesky factor
${\mathbf{F}}^{{\tilde l}_i} \in {\Re ^{K \times K}} $ satisfying
\begin{equation}\label{FFt2AAI3290ds3sd23}
{\mathbf{F}}^{{\tilde l}_i} ({\mathbf{F}}^{{\tilde l}_i})^{T}={({{  {({{\bf{A}}^{{{\tilde l}_i}}})^{T}}{{\bf{A}}^{{{\tilde l}_i}}}+\lambda \mathbf{I}}})^{-1}}
 \end{equation}
 and the corresponding output weights ${\mathbf{W}}^{{\tilde l}_i}$ satisfying
 \begin{equation}\label{WfromFFFFFF93281329k3sd2Wli}
{\mathbf{W}}^{{\tilde l}_i}={\mathbf{F}}^{{\tilde l}_i} {({\mathbf{F}}^{{\tilde l}_i})^{T}} {({{\bf{A}}^{{{\tilde l}_i}}})^{T}} {{{\bf{Y}}^{{{\tilde l}_i}}}},
\end{equation}
 where
\begin{equation}\label{A2difWorkers3290sd23}
{{\bf{A}}^{{{\tilde l}_i}}} = \left[ {\begin{array}{*{20}{c}}
{{{\bf{A}}^{{l_1}}}}&{{{\bf{A}}^{{l_2}}}}& \cdots &{{{\bf{A}}^{{l_i}}}}
\end{array}} \right]
 \end{equation}
 and ${{{\bf{Y}}^{{{\tilde l}_i}}}} = \left[ {\begin{array}{*{20}{c}}
{{{\bf{Y}}^{{l_1}}}}&{{{\bf{Y}}^{{l_2}}}}& \cdots &{{{\bf{Y}}^{{l_i}}}}
\end{array}} \right] $
consist of
 \begin{equation}\label{EachCPUnColumns239sdGeNi}
{\tilde l}_i={l _1} + {l _2} + \cdots + {l _i}
 \end{equation}
 training samples and labels (in workers $1,2,\cdots,i$), respectively.
 Notice that we can simply write (\ref{Q2PiPiT9686954})
 and
(\ref{WfromFFFFFF93281329k3sd})
as the above
(\ref{FFt2AAI3290ds3sd23})
and
(\ref{WfromFFFFFF93281329k3sd2Wli}),
respectively.

To update the inverse Cholesky factor and the output weights incrementally
without retraining the whole network from the beginning,
 worker $1$ computes
 the upper-triangular  inverse Cholesky factor
${\mathbf{F}}^{{\tilde l}_1}$ and  the corresponding output weights ${\mathbf{W}}^{{\tilde l}_1}$
by (\ref{FFt2AAI3290ds3sd23})
and
(\ref{WfromFFFFFF93281329k3sd2Wli}), respectively,
and then transmits ${\mathbf{F}}^{{\tilde l}_1}$ and ${\mathbf{W}}^{{\tilde l}_1}$  to worker $2$.
%
 After receiving ${\mathbf{F}}^{{\tilde l}_{i-1}}$ and ${\mathbf{W}}^{{\tilde l}_{i-1}}$ from worker $i-1$,
  worker $i$ ($i = 2,3,\cdots,\tau-1$)
 uses the
$l_i$ training samples ${{\bf{A}}^{{l_i}}}$ in worker $i$
to
 obtain the upper-triangular  inverse Cholesky factor ${\mathbf{V}}^{{\tilde l}_{i}} \in {\Re ^{K \times K}} $ satisfying
\begin{equation}\label{Lwave2IKK40425a}
{\mathbf{V}}^{{\tilde l}_{i}}  ({\mathbf{V}}^{{\tilde l}_{i}})^{T} ={{\left(\mathbf{I}+ ({\bf{E}}^{{\tilde l}_{i}})^{T} {\bf{E}}^{{\tilde l}_{i}}\right)}^{-1}}
 \end{equation}
 with the $l_i \times K$ matrix
\begin{equation}\label{K2AxLm94835forE}
{\bf{E}}^{{\tilde l}_{i}} = {\bf{A}}^{{l}_{i}}{\mathbf{F}}^{{\tilde l}_{i-1}},
\end{equation}
and updates ${\mathbf{F}}^{{\tilde l}_{i-1}}$ and ${\mathbf{W}}^{{\tilde l}_{i-1}}$ into
\begin{equation}\label{Lbig2LLwave59056toli}
{\mathbf{F}}^{{\tilde l}_{i}}={\mathbf{F}}^{{\tilde l}_{i-1}}{\mathbf{V}}^{{\tilde l}_{i}}
\end{equation}
and
\begin{equation}\label{WfromF3209db}
{\mathbf{W}}^{{\tilde l}_{i}} = {\mathbf{W}}^{{\tilde l}_{i-1}} + {\mathbf{F}}^{{\tilde l}_{i}} ({\mathbf{F}}^{{\tilde l}_{i}})^T ({\bf{A}}^{{l}_{i}})^T
({  {\bf{Y}}^{{l}_{i}} - {\bf{A}}^{{l}_{i}} {\mathbf{W}}^{{\tilde l}_{i-1}} }),
 \end{equation}
 respectively.
 After computing  ${\mathbf{F}}^{{\tilde l}_{i}}$ and ${\mathbf{W}}^{{\tilde l}_{i}}$  by (\ref{Lbig2LLwave59056toli})
and
(\ref{WfromF3209db}), respectively,
    worker $i$  ($i = 2,3,\cdots,\tau-1$) transmits  ${\mathbf{F}}^{{\tilde l}_{i}}$ and ${\mathbf{W}}^{{\tilde l}_{i}}$ to worker $i+1$.
 Finally,  worker $\tau$
 uses the
$l_{\tau}$ training samples ${{\bf{A}}^{{l_{\tau}}}}$ in worker $\tau$
to
 obtain the upper-triangular  inverse Cholesky factor ${\mathbf{V}}^{{\tilde l}_{\tau}}$
 by (\ref{K2AxLm94835forE}) and  (\ref{Lwave2IKK40425a}),
and updates ${\mathbf{F}}^{{\tilde l}_{\tau-1}}$ and ${\mathbf{W}}^{{\tilde l}_{\tau-1}}$ into
${\mathbf{F}}^{{\tilde l}_{\tau}}$ and ${\mathbf{W}}^{{\tilde l}_{\tau}}$
by (\ref{Lbig2LLwave59056toli})
and
(\ref{WfromF3209db}),
 respectively.
 Notice that we can
  compare (\ref{K2AxLm94835}) and
(\ref{K2AxLm94835forE})
to deduce
\begin{equation}\label{E2St32034sd34ds32}
{\bf{E}}^{{\tilde l}_{2}} =\mathbf{S}^{T},
 \end{equation}
 and then
  write
 (\ref{Lwave2IKK40425aBefore}),
 (\ref{Lbig2LLwave59056})  
and
(\ref{WfromF3209dbBefore932})
as
the above
(\ref{Lwave2IKK40425a}),
(\ref{Lbig2LLwave59056toli})
and
(\ref{WfromF3209db}), respectively.

%
%
%
%
%
%
%
%
%
%

\subsection{Several Variables for the Proposed Parallel Implementation}

In this subsection, we introduce several variables that will be utilized in
the proposed parallel implementation.

The first $k$ ($k \le K$) columns of
${\bf{E}}^{{\tilde l}_{i}}$ defined by (\ref{K2AxLm94835forE}) can be denoted as
\begin{equation}\label{E2AF23420d}
{\bf{E}}_{|k}^{{\tilde l}_{i}}={\bf{E}}^{{\tilde l}_{i}}(:,1:k)={\bf{A}}^{{l}_{i}}(:,1:k) {\mathbf{F}}_k^{{\tilde l}_{i-1}},
 \end{equation}
 where ${\mathbf{F}}_k^{{\tilde l}_{i-1}}$ is the $k \times k$ leading principal sub-matrix in
 the upper-triangular ${\mathbf{F}}^{{\tilde l}_{i-1}}$,
 and the notations ${\bf{E}}^{{\tilde l}_{i}}(:,1:k)$ and ${\bf{A}}^{{l}_{i}}(:,1:k)$ follow the Matlab standard.  Then let us utilize
 ${\bf{E}}_{|k}^{{\tilde l}_{i}}$ to define the $k \times k$
 upper-triangular  inverse Cholesky factor ${\mathbf{V}}_k^{{\tilde l}_{i}}$ satisfying
 \begin{equation}\label{VV2IEEi32sd23}
{\mathbf{V}}_k^{{\tilde l}_{i}}  ({\mathbf{V}}_k^{{\tilde l}_{i}})^{T} ={{\left(\mathbf{I}+ ({\bf{E}}_{|k}^{{\tilde l}_{i}})^{T} {\bf{E}}_{|k}^{{\tilde l}_{i}}\right)}^{-1}},
 \end{equation}
which can be written as
\begin{equation}\label{VV2R320d3de3d}
{\mathbf{V}}_k^{{\tilde l}_{i}}  ({\mathbf{V}}_k^{{\tilde l}_{i}})^{T}=({\bf{R}}_k^{{\tilde l}_{i}})^{ - 1}
 \end{equation}
with
 \begin{equation}\label{R2EEI3ds23ds}
{\bf{R}}_k^{{\tilde l}_{i}} = ({\bf{E}}_{|k}^{{\tilde l}_{i}})^{T} {\bf{E}}_{|k}^{{\tilde l}_{i}} + {\bf{I}}.
 \end{equation}
The inverse Cholesky factorization~\cite{my_inv_chol_paper} can be applied
to update ${\mathbf{V}}_{k-1}^{{\tilde l}_{i}}$ and ${{{{\bf{F}}}_{k - 1}^{{\tilde l}_{i}}}}$  into ${\mathbf{V}}_k^{{\tilde l}_{i}}$
and  ${{\bf{F}}_k^{{\tilde l}_{i}}}$, respectively,
by                                    
 \begin{equation}\label{VmatrixIter2deduce3}
{\mathbf{V}}_k^{{\tilde l}_{i}} = \left[ {\begin{array}{*{20}{c}}
{\mathbf{V}}_{k-1}^{{\tilde l}_{i}} &{{\bf{\bar v}}_k^{{\tilde l}_{i}}}\\
{{\bf{0}}_{k - 1}^T}&{{v_{kk}^{{\tilde l}_{i}}}}
\end{array}} \right]=\left[ {\begin{array}{*{20}{c}}
{\left( {\begin{array}{*{20}{c}}
{\mathbf{V}}_{k-1}^{{\tilde l}_{i}} \\
{{\bf{0}}_{k - 1}^T}
\end{array}} \right)}& {{\bf{v}}_k^{{\tilde l}_{i}}}
\end{array}} \right]
 \end{equation}
 and 
 \begin{equation}\label{Fi2ffiiT3902originalF}
{{\bf{F}}_k^{{\tilde l}_{i}}} =
 \left[ {\begin{array}{*{20}{c}}
{{{{\bf{F}}}_{k - 1}^{{\tilde l}_{i}}}}& {{{\bf{\bar f}}}_k^{{\tilde l}_{i}}} \\
{{\bf{0}}_{k - 1}^T}& {f_{kk}^{{\tilde l}_{i}}}
\end{array}} \right]
=\left[ {\begin{array}{*{20}{c}}
{\left( {\begin{array}{*{20}{c}}
{{{{\bf{F}}}_{k - 1}^{{\tilde l}_{i}}}} \\
{{\bf{0}}_{k - 1}^T}
\end{array}} \right)}& {{{\bf{f}}}_k^{{\tilde l}_{i}}}
\end{array}} \right],
 \end{equation}
 where $k=2,3,\cdots,K$.  In (\ref{VmatrixIter2deduce3}),  ${{\bf{\bar v}}_k^{{\tilde l}_{i}}}$ can be computed by~\cite{my_inv_chol_paper}
\begin{equation}\label{BarVecV329sdk239dsew3ds}
{{\bf{\bar v}}_k^{{\tilde l}_{i}}} =  - {{v_{kk}^{{\tilde l}_{i}}}} {\mathbf{V}}_{k-1}^{{\tilde l}_{i}} ({\mathbf{V}}_{k-1}^{{\tilde l}_{i}})^T{\bf{R}}_k^{{\tilde l}_{i}}(1:k - 1,k).
 \end{equation}



  The parallel implementation of the inverse Cholesky factorization~\cite{CholBLSnodesSubmitted}
  can be applied to compute the $k^{th}$ column of ${{\bf{V}}_{k}^{{\tilde l}_{i}}}$
(i.e., ${{\bf{\bar v}}_k^{{\tilde l}_{i}}}$ and ${{v_{kk}^{{\tilde l}_{i}}}}$ in  (\ref{VmatrixIter2deduce3}))
by
\begin{subnumcases}{\label{fiifli2Pi3902sdBoth}}
{{v_{kk}^{{\tilde l}_{i}}}} = 1/\sqrt {{{{\bf{\Xi }}}_{K - k + 1}^{{\tilde l}_{i}}}(1,1)} &  \label{fii2sqrt2390sd23}\\
{{\bf{\bar v}}_k^{{\tilde l}_{i}}} =   {{v_{kk}^{{\tilde l}_{i}}}} {{\bf{\Pi }}_{k - 1}^{{\tilde l}_{i}}}(:,1),  &  \label{f1i1i2gPi39d234d3}
\end{subnumcases}
where
 ${{\mathbf{\Pi }}_{k}}$ and ${{\mathbf{\Xi }}_{K-k}}$ are defined by~\cite{CholBLSnodesSubmitted}
\begin{small}
 \begin{equation}\label{pi2FFAA2390923}
{{\mathbf{\Pi }}_{k}^{{\tilde l}_{i}}}=- {\mathbf{V}}_k^{{\tilde l}_{i}} ({\mathbf{V}}_k^{{\tilde l}_{i}})^T({\bf{E}}_{|k}^{{\tilde l}_{i}})^T{\bf{\underline{E}}}_{K-k}^{{\tilde l}_{i}}
 \end{equation}
 \end{small}
and
 \begin{small}
 \begin{multline}\label{Xi2AAFFAA23094d3}
{{\mathbf{\Xi }}_{K-k}^{{\tilde l}_{i}}}= ({\bf{\underline{E}}}_{K-k}^{{\tilde l}_{i}})^T {\bf{\underline{E}}}_{K-k}^{{\tilde l}_{i}}  + {\bf{I}}
 \\ - ({\bf{\underline{E}}}_{K-k}^{{\tilde l}_{i}})^T  {\bf{E}}_{|k}^{{\tilde l}_{i}} {\mathbf{V}}_k^{{\tilde l}_{i}} ({\mathbf{V}}_k^{{\tilde l}_{i}})^T
  ({\bf{E}}_{|k}^{{\tilde l}_{i}})^T  {\bf{\underline{E}}}_{K-k}^{{\tilde l}_{i}},
 \end{multline}
 \end{small}
respectively.
 In (\ref{pi2FFAA2390923}) and
(\ref{Xi2AAFFAA23094d3}), ${\bf{\underline{E}}}_{K-k}^{{\tilde l}_{i}}$ denotes
 the  last $K-k$ columns of
 the $l_i \times K$ matrix
  ${\bf{E}}^{{\tilde l}_{i}}$,
   and then from (\ref{K2AxLm94835forE}), we can deduce
\begin{equation}\label{Eunderline2AF320ds}
{\bf{\underline{E}}}_{K-k}^{{\tilde l}_{i}}={\bf{E}}^{{\tilde l}_{i}}(:,k + 1:K) = {\bf{A}}^{{l}_{i}}{\mathbf{F}}^{{\tilde l}_{i-1}}(:,k + 1:K).
 \end{equation}

 Substitute (\ref{E2AF23420d}) and
(\ref{Eunderline2AF320ds}) into
(\ref{pi2FFAA2390923})
and
(\ref{Xi2AAFFAA23094d3})
to obtain
\begin{multline}\label{PiDefin923sd32ds}
{{\mathbf{\Pi }}_{k}^{{\tilde l}_{i}}} =  - {\mathbf{V}}_k^{{\tilde l}_{i}} ({\mathbf{V}}_k^{{\tilde l}_{i}})^T  ({\mathbf{F}}_k^{{\tilde l}_{i-1}})^T{\bf{A}}^{{l}_{i}}(:,1:k)^T
 {\bf{A}}^{{l}_{i}}   \\
\times {\mathbf{F}}^{{\tilde l}_{i-1}}(:,k + 1:K)
 \end{multline}
and
\begin{footnotesize}
\begin{align}
{{\bf{\Xi }}_{K-k}^{{\tilde l}_{i}}}= &  {\mathbf{F}}^{{\tilde l}_{i-1}}(:,k + 1:K)^T ({\bf{A}}^{{l}_{i}})^T{\bf{A}}^{{l}_{i}}{\mathbf{F}}^{{\tilde l}_{i-1}}(:,k + 1:K) + {\bf{I}} -  \notag \\
&{\mathbf{F}}^{{\tilde l}_{i-1}}(:,k + 1:K)^T   ({\bf{A}}^{{l}_{i}})^T {\bf{A}}^{{l}_{i}}(:,1:k) {\mathbf{F}}_k^{{\tilde l}_{i-1}} {\mathbf{V}}_k^{{\tilde l}_{i}} ({\mathbf{V}}_k^{{\tilde l}_{i}})^T \times  \notag \\
&({\mathbf{F}}_k^{{\tilde l}_{i-1}})^T{\bf{A}}^{{l}_{i}}{(:,1:k)^T}{\bf{A}}^{{l}_{i}} {\mathbf{F}}^{{\tilde l}_{i-1}}(:,k + 1:K)  \notag \\
=&{\mathbf{F}}^{{\tilde l}_{i-1}}(:,k + 1:K)^T  \notag \\
& \left( \begin{array}{l}
{({{\bf{A}}^{{l_i}}})^T}{{\bf{A}}^{{l_i}}} - {({{\bf{A}}^{{l_i}}})^T}{{\bf{A}}^{{l_i}}}(:,1:k) {\mathbf{F}}_k^{{\tilde l}_{i-1}}  \\
  \times {\mathbf{V}}_k^{{\tilde l}_{i}} ({\mathbf{V}}_k^{{\tilde l}_{i}})^T ({\mathbf{F}}_k^{{\tilde l}_{i-1}})^T{\bf{A}}^{{l}_{i}}{(:,1:k)^T}{\bf{A}}^{{l}_{i}}
\end{array} \right)  \notag \\
&\times {\mathbf{F}}^{{\tilde l}_{i-1}}(:,k + 1:K)  +{\bf{I}},  \label{Xi2IFAAAAFVVeidsw493ds3}
\end{align}
\end{footnotesize}
respectively.
Then write (\ref{PiDefin923sd32ds}) as
 \begin{equation}\label{Pi2PiTilde239d3}
{{\mathbf{\Pi }}_{k}^{{\tilde l}_{i}}} = {{\mathbf{\tilde \Pi }}_{k}^{{\tilde l}_{i}}} {\mathbf{F}}^{{\tilde l}_{i-1}}(:,k + 1:K)
 \end{equation}
 where ${{\mathbf{\tilde \Pi }}_{k}^{{\tilde l}_{i}}} \in {\Re ^{k \times K}}$ is defined by
\begin{equation}\label{PiTildeDef320ds23}
{{\mathbf{\tilde \Pi }}_{k}^{{\tilde l}_{i}}} =  - {\mathbf{V}}_k^{{\tilde l}_{i}} ({\mathbf{V}}_k^{{\tilde l}_{i}})^T ({\mathbf{F}}_k^{{\tilde l}_{i-1}})^T{\bf{A}}^{{l}_{i}}(:,1:k)^T
 {\bf{A}}^{{l}_{i}},
 \end{equation}
 and
  write  (\ref{Xi2IFAAAAFVVeidsw493ds3}) as
 \begin{equation}\label{Xi2IFXiF3249ds}
{{\bf{\Xi }}_{K-k}^{{\tilde l}_{i}}} = {\bf{I}} + {\mathbf{F}}^{{\tilde l}_{i-1}}(:,k + 1:K)^T {{\bf{\tilde \Xi }}_k^{{\tilde l}_{i}}}{\mathbf{F}}^{{\tilde l}_{i-1}}(:,k + 1:K)
 \end{equation}
 where    ${{\bf{\tilde \Xi}}_{k}} \in {\Re ^{K \times K}}$  is defined by
 \begin{multline}\label{XiTildeDef320ds23}
{{\bf{\tilde \Xi}}_{k}^{{\tilde l}_{i}}}= {({{\bf{A}}^{{l_i}}})^T}{{\bf{A}}^{{l_i}}} - {({{\bf{A}}^{{l_i}}})^T}{{\bf{A}}^{{l_i}}}(:,1:k) {\mathbf{F}}_k^{{\tilde l}_{i-1}}
\times \\ {\mathbf{V}}_k^{{\tilde l}_{i}} ({\mathbf{V}}_k^{{\tilde l}_{i}})^T ({\mathbf{F}}_k^{{\tilde l}_{i-1}})^T{\bf{A}}^{{l}_{i}}{(:,1:k)^T}{\bf{A}}^{{l}_{i}}.
 \end{multline}
 In what follows, 
  ${{\mathbf{\tilde \Pi }}_{k}^{{\tilde l}_{i}}}$  defined by (\ref{PiTildeDef320ds23})
  and  ${{\bf{\tilde \Xi}}_{k}}$  defined by (\ref{XiTildeDef320ds23}) will be applied to develop the
 parallel implementation of the proposed square-root BLS algorithm based on inverse
Cholesky factor.

 \subsection{Parallel Implementation of Proposed Square-Root BLS Algorithm Based on Inverse
Cholesky Factor}


As described above,  worker $1$
possesses
 ${l _1}$ training samples, i.e.,
${{\bf{A}}^{{{\tilde l}_1}}}={{\bf{A}}^{{l_1}}}$,
which are utilized
 to
  compute
 the upper-triangular  inverse Cholesky factor
${\mathbf{F}}^{{\tilde l}_1}$ satisfying (\ref{FFt2AAI3290ds3sd23}) with $i=1$.
  Worker $1$  can use
  the inverse Cholesky factorization~\cite{my_inv_chol_paper}
  or the  corresponding parallel implementation~\cite{CholBLSnodesSubmitted}
  to compute
  ${\mathbf{F}}^{{\tilde l}_1}$ in $K$ iterations. In the $k^{th}$ ($k=1,2,\cdots,K$) iteration,
  worker $1$ updates the $(k-1) \times (k-1)$ upper-triangular ${{{{\bf{F}}}_{k - 1}^{{\tilde l}_{1}}}}$ into the $k \times k$ upper-triangular ${{\bf{F}}_k^{{\tilde l}_{1}}}$
  by
  (\ref{Fi2ffiiT3902originalF}) with $i=1$, i.e.,
  \begin{equation}\label{Fi2ffiiT3902originalFWITHi1}
{{\bf{F}}_k^{{\tilde l}_{1}}} =
 \left[ {\begin{array}{*{20}{c}}
{{{{\bf{F}}}_{k - 1}^{{\tilde l}_{1}}}}& {{{\bf{\bar f}}}_k^{{\tilde l}_{1}}} \\
{{\bf{0}}_{k - 1}^T}& {f_{kk}^{{\tilde l}_{1}}}
\end{array}} \right]
=\left[ {\begin{array}{*{20}{c}}
{\left( {\begin{array}{*{20}{c}}
{{{{\bf{F}}}_{k - 1}^{{\tilde l}_{1}}}} \\
{{\bf{0}}_{k - 1}^T}
\end{array}} \right)}& {{{\bf{f}}}_k^{{\tilde l}_{1}}}
\end{array}} \right].
 \end{equation}
 It can be seen from (\ref{Fi2ffiiT3902originalFWITHi1}) that   worker $1$
 only needs to compute ${\bf{f}}_k^{{\tilde l}_{1}}={\left[ {\begin{array}{*{20}{c}}
{{({{{\bf{\bar f}}}_k^{{\tilde l}_{1}}})^T}}&{f_{kk}^{{\tilde l}_{1}}}
\end{array}} \right]^T}$,
the $k^{th}$ column of ${{\bf{F}}_k^{{\tilde l}_{1}}}$ that contains the nonzero entries in the $k^{th}$ column of the  upper-triangular ${\mathbf{F}}^{{\tilde l}_1}$.
To improve the
parallelization, worker $1$ can transmit ${\bf{f}}_k^{{\tilde l}_{1}}$ to worker $2$, after
${\bf{f}}_k^{{\tilde l}_{1}}$ is computed in  the $k^{th}$ iteration.
Accordingly, when worker $1$ is computing some of the columns  $k+1,k+2,\cdots,K$
in
 ${\mathbf{F}}^{{\tilde l}_1}$,
worker $2$ can utilize ${\bf{f}}_k^{{\tilde l}_{1}}$ to compute
the $k^{th}$ column of the  upper-triangular ${\mathbf{F}}^{{\tilde l}_2}$
 in parallel, and does not need to wait till it receives the whole
  $K \times K$  matrix  ${\mathbf{F}}^{{\tilde l}_1}$ from  worker $1$.

%


Let us describe the general case for worker $i$  where  $i = 2,3,\cdots,\tau-1$.
After receiving ${\bf{f}}_k^{{\tilde l}_{i-1}}$ from worker $i-1$,
worker $i$ ($i = 2,3,\cdots,\tau-1$)
computes
\begin{subnumcases}{\label{equ5forWorkeriii2390sd23}}
{\mathbf{\theta }}_k^{{\tilde l}_{i}} = {\bf{\tilde \Xi }}_{k-1}^{{\tilde l}_{i}}(1:k,:)^T {\bf{f}}_k^{{\tilde l}_{i-1}} &  \label{ThetaDef30ds3}\\
{v_{kk}^{{\tilde l}_{i}}} = 1/\sqrt {1 + {\bf{\theta }}_k^{{\tilde l}_{i}}(1:k)^T{\bf{f}}_k^{{\tilde l}_{i-1}}} &  \label{vii2thetaPart12390dscx23}\\
{\bf{\bar v}}_k^{{\tilde l}_{i}} = {v_{kk}^{{\tilde l}_{i}}}{{\bf{\tilde \Pi }}_{k - 1}^{{\tilde l}_{i}}}(:,1:k){\bf{f}}_k^{{\tilde l}_{i-1}} &  \label{vBarIter2deduce2}\\
{{{\bf{\bar f}}}_k^{{\tilde l}_{i}}}={{{\bf{F}}_{k - 1}^{{\tilde l}_{i}}}{\bf{\bar v}}_k^{{\tilde l}_{i}} + {v_{kk}^{{\tilde l}_{i}}}{{{\bf{\bar f}}}_k^{{\tilde l}_{i-1}}}} &  \label{fkli2Flvvfl932sd}\\
{f_{kk}^{{\tilde l}_{i}}}={{f_{kk}^{{\tilde l}_{i-1}}}{v_{kk}^{{\tilde l}_{i}}}}  &  \label{fkk2fkkvkk239sd23sd} \\
{{\bf{F}}_k^{{\tilde l}_{i}}} = \left[ {\begin{array}{*{20}{c}}
{{{{\bf{F}}}_{k - 1}^{{\tilde l}_{i}}}}& {{{\bf{\bar f}}}_k^{{\tilde l}_{i}}} \\
{{\bf{0}}_{k - 1}^T}& {f_{kk}^{{\tilde l}_{i}}}
\end{array}} \right],  &  \label{FtildeIterDeduceEasy}
\end{subnumcases}
and then transmits ${\bf{f}}_k^{{\tilde l}_{i}}={\left[ {\begin{array}{*{20}{c}}
{{({{{\bf{\bar f}}}_k^{{\tilde l}_{i}}})^T}}&{f_{kk}^{{\tilde l}_{i}}}
\end{array}} \right]^T}$ to worker $i+1$.
Notice that
the above  ${v_{kk}^{{\tilde l}_{i}}}$ and ${\bf{\bar v}}_k^{{\tilde l}_{i}}$
  form the $k^{th}$ column of
  ${\bf{V}}_k^{{\tilde l}_{i}}$ by (\ref{VmatrixIter2deduce3}),
while
${\bf{f}}_k^{{\tilde l}_{i}}={\left[ {\begin{array}{*{20}{c}}
{{({{{\bf{\bar f}}}_k^{{\tilde l}_{i}}})^T}}&{f_{kk}^{{\tilde l}_{i}}}
\end{array}} \right]^T}$
forms
   the $k^{th}$ column of ${{\bf{F}}_k^{{\tilde l}_{i}}}$ by (\ref{FtildeIterDeduceEasy}), i.e.,  (\ref{Fi2ffiiT3902originalF}).
 Moreover, for the next
 iteration, 
  worker $i$ updates
   ${{\bf{\tilde \Pi }}_{k-1}^{{\tilde l}_{i}}}$ and
 ${{\bf{\tilde \Xi }}_{k-1}^{{\tilde l}_{i}}}$
  into
  ${{\bf{\tilde \Pi }}_k^{{\tilde l}_{i}}}$ and ${{\bf{\tilde \Xi }}_k^{{\tilde l}_{i}}}$, respectively,
  by
\begin{subnumcases}{\label{PiXiIter3209sd23sd23}}
 {{\bf{\tilde \Pi }}_k^{{\tilde l}_{i}}} = \left[ {\begin{array}{*{20}{c}}
{{{\bf{\tilde \Pi }}_{k - 1}^{{\tilde l}_{i}}}}\\
{{{\bf{0}}^T}}
\end{array}} \right] - {v_{kk}^{{\tilde l}_{i}}}  \left[ {\begin{array}{*{20}{c}}
 {\bf{\bar v}}_k^{{\tilde l}_{i}} \\
{v_{kk}^{{\tilde l}_{i}}}
\end{array}} \right]  {({\mathbf{\theta }}_k^{{\tilde l}_{i}})^T}&  \label{PiIter2deduce4}\\
 {{\bf{\tilde \Xi }}_k^{{\tilde l}_{i}}} = {{\bf{\tilde \Xi }}_{k - 1}^{{\tilde l}_{i}}} - ({v_{kk}^{{\tilde l}_{i}}})^2{\mathbf{\theta }}_k^{{\tilde l}_{i}} {({\mathbf{\theta }}_k^{{\tilde l}_{i}})^T}. &  \label{XiIter2deduce5}
\end{subnumcases}
When $i =\tau$,
 worker $\tau$ also computes (\ref{equ5forWorkeriii2390sd23}) and (\ref{PiXiIter3209sd23sd23}),
 while it is no longer necessary for worker $\tau$ to transmit  ${\bf{f}}_k^{{\tilde l}_{\tau}}$ to another worker.
 The above (\ref{equ5forWorkeriii2390sd23}) and (\ref{PiXiIter3209sd23sd23}) will be deduced in the following subsections D and E, respectively.


In the $k^{th}$ ($k=1,2,\cdots,K$) iteration,
  worker $1$ computes ${\bf{f}}_k^{{\tilde l}_{1}}$
 and transmits it to worker $2$,   worker $i$ ($i=2,3,\cdots,\tau-1$) then computes (\ref{equ5forWorkeriii2390sd23})
   to transmit
  ${\bf{f}}_k^{{\tilde l}_{i}}$ to worker $i+1$, and
  worker $\tau$ computes (\ref{equ5forWorkeriii2390sd23})
    with $i=\tau$ to obtain  ${\bf{f}}_k^{{\tilde l}_{\tau}}$ finally.
  It can be seen that the delay in worker $i$ ($i=2,3,\cdots,\tau$) mainly comes from~\footnote{Notice that ${{\bf{\tilde \Pi }}_{k-1}^{{\tilde l}_{i}}}$ and
 ${{\bf{\tilde \Xi }}_{k-1}^{{\tilde l}_{i}}}$
 are updated
  into
  ${{\bf{\tilde \Pi }}_k^{{\tilde l}_{i}}}$ and ${{\bf{\tilde \Xi }}_k^{{\tilde l}_{i}}}$ by (\ref{PiXiIter3209sd23sd23}) after worker $i$  transmits  ${\bf{f}}_k^{{\tilde l}_{i}}$ to worker $i+1$,
  and then usually the computation of (\ref{PiXiIter3209sd23sd23}) does not delay the transmission of ${\bf{f}}_k^{{\tilde l}_{i}}$.} the computation of (\ref{equ5forWorkeriii2390sd23}).
   The dominant computational complexity of  (\ref{equ5forWorkeriii2390sd23}) comes from (\ref{ThetaDef30ds3})~\footnote{In (\ref{ThetaDef30ds3}), the calculation of the $(k+1)^{th}, (k+2)^{th}, \cdots, K^{th}$ entries in ${\mathbf{\theta }}_k^{{\tilde l}_{i}}$ can be delayed
  until ${\bf{f}}_k^{{\tilde l}_{i}}$ is transmitted, since only ${\bf{\theta }}_k^{{\tilde l}_{i}}(1:k)$ (including the first $k$ entries
 of ${\mathbf{\theta }}_k^{{\tilde l}_{i}}$) is required to compute  ${\bf{f}}_k^{{\tilde l}_{i}}$ by (\ref{equ5forWorkeriii2390sd23}),
 and ${\bf{\theta }}_k^{{\tilde l}_{i}}(k+1:K)$ is just utilized in (\ref{PiXiIter3209sd23sd23}).},
(\ref{vBarIter2deduce2}) and (\ref{fkli2Flvvfl932sd}), which only require  the flops (floating point operations)
of $2k^2$, $2k^2$ and $k^2$, respectively.
Thus it can be expected that the computation of (\ref{equ5forWorkeriii2390sd23}) will not cause too long a delay.



The proposed parallel implementation of square-root BLS is summarized in \textbf{Algorithm 7},
where ${\mathbf{F}}_k^{{\tilde l}_{i}}$ ($i = 1,2,\cdots,\tau$) denotes the $k \times k$ leading principal sub-matrix in
 the upper-triangular ${\mathbf{F}}^{{\tilde l}_{i}}$, as mentioned above. 
Notice that in \textbf{Algorithm 7},
  ${\bf{f}}_k^{{\tilde l}_{i}}$ ($i = 1,2,\cdots,\tau$),  the $k^{th}$ column of ${{\bf{F}}_k^{{\tilde l}_{i}}}$, contains the nonzero entries in the $k^{th}$ column of the  upper-triangular ${\mathbf{F}}^{{\tilde l}_i}$.

\begin{algorithm}
\caption{:~\bf The Proposed Parallel Implementation of Square-Root BLS}
\begin{algorithmic}[1]
\Require ${{\bf{A}}^{{l_i}}} \in {\Re ^{l_i \times K}}$  ($i = 1,2,\cdots,\tau$) in worker $i$,
which denotes ${l _i}$ training samples stored in worker $i$;
\Ensure  The upper-triangular  inverse Cholesky factor
${\mathbf{F}}^{{\tilde l}_\tau} \in {\Re ^{K \times K}} $ satisfying
${\mathbf{F}}^{{\tilde l}_\tau} ({\mathbf{F}}^{{\tilde l}_\tau})^{T}={({{  {({{\bf{A}}^{{{\tilde l}_\tau}}})^{T}}{{\bf{A}}^{{{\tilde l}_\tau}}}+\lambda \mathbf{I}}})^{-1}}$ where ${{\bf{A}}^{{{\tilde l}_\tau}}} = \left[ {\begin{array}{*{20}{c}}
{{{\bf{A}}^{{l_1}}}}&{{{\bf{A}}^{{l_2}}}}& \cdots &{{{\bf{A}}^{{l_\tau}}}}
\end{array}} \right]$;
\For{$k=1:K$ ($K$ is the size of ${\bf{R}}$)}
\State   Worker $1$  transmits ${\bf{f}}_k^{{\tilde l}_{1}}$ to worker $2$, where ${\bf{f}}_k^{{\tilde l}_{1}}$ is  the $k^{th}$ column of ${{\bf{F}}_k^{{\tilde l}_{1}}}\in {\Re ^{k \times k}}$;
 Then worker $1$ computes some of the columns  $k+1,k+2,\cdots, K$ in
${\mathbf{F}}^{{\tilde l}_1}$;
\State  When worker $1$ is computing some of the columns  $k+1,k+2,\cdots, K$ in ${\mathbf{F}}^{{\tilde l}_1}$,
worker $2$  computes (\ref{equ5forWorkeriii2390sd23}) with $i=2$ in parallel
 to obtain ${\bf{f}}_k^{{\tilde l}_{2}}$
and form ${{\bf{F}}_k^{{\tilde l}_{2}}} \in {\Re ^{k \times k}}$
   after  receiving ${\bf{f}}_k^{{\tilde l}_{1}}$ from worker $1$, and  then  transmits ${\bf{f}}_k^{{\tilde l}_{2}}$ to worker $3$.
After ${\bf{f}}_k^{{\tilde l}_{2}}$ is computed,
 worker  $2$ also computes (\ref{PiXiIter3209sd23sd23}) to  update
   ${{\bf{\tilde \Pi }}_{k-1}^{{\tilde l}_{2}}}$ and
 ${{\bf{\tilde \Xi }}_{k-1}^{{\tilde l}_{2}}}$
  into
  ${{\bf{\tilde \Pi }}_k^{{\tilde l}_{2}}}$ and ${{\bf{\tilde \Xi }}_k^{{\tilde l}_{2}}}$, respectively, which is utilized in the $(k+1)^{th}$ iteration for worker  $2$;
\State  $\cdots \cdots$
\State When workers $i-1$ is  computing (\ref{PiXiIter3209sd23sd23}) to obtain ${{\bf{\tilde \Pi }}_k^{{\tilde l}_{i-1}}}$ and ${{\bf{\tilde \Xi }}_k^{{\tilde l}_{i-1}}}$, or computing some of the
columns  $k+1,k+2,\cdots, K$ in the  upper-triangular ${\mathbf{F}}^{{\tilde l}_{i-1}}$,
worker $i$ ($i=3,4,\cdots,\tau-1$) computes (\ref{equ5forWorkeriii2390sd23}) in parallel to obtain  ${\bf{f}}_k^{{\tilde l}_{i}}$
and form  ${{\bf{F}}_k^{{\tilde l}_{i}}} \in {\Re ^{k \times k}}$
  after receiving ${\bf{f}}_k^{{\tilde l}_{i-1}}$ from worker $i-1$, and then  transmits ${\bf{f}}_k^{{\tilde l}_{i}}$ to worker $i+1$. After ${\bf{f}}_k^{{\tilde l}_{i}}$ is computed,
 worker  $i$ also computes (\ref{PiXiIter3209sd23sd23}) to  update
   ${{\bf{\tilde \Pi }}_{k-1}^{{\tilde l}_{i}}}$ and
 ${{\bf{\tilde \Xi }}_{k-1}^{{\tilde l}_{i}}}$
  into
  ${{\bf{\tilde \Pi }}_k^{{\tilde l}_{i}}}$ and ${{\bf{\tilde \Xi }}_k^{{\tilde l}_{i}}}$, respectively, which is utilized in the $(k+1)^{th}$ iteration for worker  $i$.
\State  $\cdots \cdots$
\State  When workers $\tau-1$ is  computing (\ref{PiXiIter3209sd23sd23}) with $i=\tau-1$ to obtain ${{\bf{\tilde \Pi }}_k^{{\tilde l}_{\tau-1}}}$ and ${{\bf{\tilde \Xi }}_k^{{\tilde l}_{\tau-1}}}$, or computing
some of the columns  $k+1,k+2,\cdots, K$ in the  upper-triangular ${\mathbf{F}}^{{\tilde l}_{\tau-1}}$,
worker $\tau$ computes (\ref{equ5forWorkeriii2390sd23}) with $i = \tau$ in parallel to obtain  ${\bf{f}}_k^{{\tilde l}_{\tau}}$
and form  ${{\bf{F}}_k^{{\tilde l}_\tau}} \in {\Re ^{k \times k}}$
  after receiving ${\bf{f}}_k^{{\tilde l}_{\tau-1}}$ from worker $\tau-1$. After ${\bf{f}}_k^{{\tilde l}_{\tau}}$ is computed,
 worker  $\tau$  computes (\ref{PiXiIter3209sd23sd23})
 to  update
   ${{\bf{\tilde \Pi }}_{k-1}^{{\tilde l}_{\tau}}}$ and
 ${{\bf{\tilde \Xi }}_{k-1}^{{\tilde l}_{\tau}}}$
  into
  ${{\bf{\tilde \Pi }}_k^{{\tilde l}_{\tau}}}$ and ${{\bf{\tilde \Xi }}_k^{{\tilde l}_{\tau}}}$, respectively, which is utilized in the $(k+1)^{th}$ iteration for worker  $\tau$;
\EndFor
\State  ${{\bf{F}}_K^{{\tilde l}_\tau}}={\mathbf{F}}^{{\tilde l}_\tau} \in {\Re ^{K \times K}}$ is computed after $K$ iterations;
\end{algorithmic}
\end{algorithm}

\subsection{The Derivation of (\ref{equ5forWorkeriii2390sd23})}



From ${\mathbf{\theta }}_k^{{\tilde l}_{i}}$ defined by (\ref{ThetaDef30ds3}), we can deduce
\begin{equation}\label{theta2Xif239sd23ds}
{\mathbf{\theta }}_k^{{\tilde l}_{i}}(1:k) = {\bf{\tilde \Xi }}_{k-1}^{{\tilde l}_{i}}(1:k,1:k)^T {\bf{f}}_k^{{\tilde l}_{i-1}}.
 \end{equation}
To deduce
(\ref{vii2thetaPart12390dscx23}),
substitute (\ref{Xi2IFXiF3249ds})
 with $k=k-1$ into (\ref{fii2sqrt2390sd23}) to obtain
\begin{equation}\label{viiIter2deduce1}
{v_{kk}^{{\tilde l}_{i}}} = 1/\sqrt {1 + ({\bf{f}}_k^{{\tilde l}_{i-1}})^T{{{\bf{\tilde \Xi }}}_{k - 1}^{{\tilde l}_{i}}}(1:k,1:k){\bf{f}}_k^{{\tilde l}_{i-1}}},
 \end{equation}
into which
 substitute (\ref{theta2Xif239sd23ds}).


%
%
%
%
%



To deduce
(\ref{vBarIter2deduce2}),
let us write
 (\ref{R2EEI3ds23ds})
 as
\begin{equation}\label{R2EE390d32d32sd}
{\bf{R}}_k^{{\tilde l}_{i}}(1:k - 1,k) = {\bf{E}}_{|k}^{{\tilde l}_{i}}(:,1:k - 1)^T{\bf{E}}_{|k}^{{\tilde l}_{i}}(:,k).
 \end{equation}
From (\ref{E2AF23420d}), we can deduce
 \begin{equation}\label{Ei2Af2389sdkqwe3}
{\bf{E}}_{|k}^{{\tilde l}_{i}}(:,k) = {\bf{A}}^{{l}_{i}}(:,1:k) {\bf{f}}_k^{{\tilde l}_{i-1}}
 \end{equation}
 and
 \begin{align}
{\bf{E}}_{|k}^{{\tilde l}_{i}}(:,1:k - 1)&= {\bf{A}}^{{l}_{i}}(:,1:k){\mathbf{F}}_k^{{\tilde l}_{i-1}}(:,1:k - 1)  \notag \\
&= {\bf{A}}^{{l}_{i}}(:,1:k){\left[ {\begin{array}{*{20}{c}}
{({\mathbf{F}}_{k-1}^{{\tilde l}_{i-1}})^T}&{{\bf{0}}_{k - 1}^{}}
\end{array}} \right]^T}  \notag \\
&= {\bf{A}}^{{l}_{i}}(:,1:k - 1){\mathbf{F}}_{k-1}^{{\tilde l}_{i-1}},  \label{EiCols9023ksd23ds3}
\end{align}
which are substituted
into
(\ref{R2EE390d32d32sd})
to obtain
\begin{multline}\label{R2EE390d32d32sd2EE39s}
{\bf{R}}_k^{{\tilde l}_{i}}(1:k - 1,k) = \\
({\mathbf{F}}_{k-1}^{{\tilde l}_{i-1}})^T{\bf{A}}^{{l}_{i}}(:,1:k - 1)^T
 {\bf{A}}^{{l}_{i}}(:,1:k) {\bf{f}}_k^{{\tilde l}_{i-1}}.
 \end{multline}
Then substitute (\ref{R2EE390d32d32sd2EE39s}) into
(\ref{BarVecV329sdk239dsew3ds})
to obtain
\begin{multline}\label{v2viiVVFAAf930ds2deduce}
{{\bf{\bar v}}_k^{{\tilde l}_{i}}} =  - {{v_{kk}^{{\tilde l}_{i}}}} {\mathbf{V}}_{k-1}^{{\tilde l}_{i}} ({\mathbf{V}}_{k-1}^{{\tilde l}_{i}})^T \\
\times ({\mathbf{F}}_{k-1}^{{\tilde l}_{i-1}})^T{\bf{A}}^{{l}_{i}}(:,1:k - 1)^T
 {\bf{A}}^{{l}_{i}}(:,1:k) {\bf{f}}_k^{{\tilde l}_{i-1}},
 \end{multline}
and write (\ref{PiTildeDef320ds23}) as
 \begin{multline}\label{Pi2lVVFAA2399sd32d}
{{\bf{\tilde \Pi }}_{k - 1}^{{\tilde l}_{i}}}(:,1:k)=- {\mathbf{V}}_{k-1}^{{\tilde l}_{i}} ({\mathbf{V}}_{k-1}^{{\tilde l}_{i}})^T \\
\times ({\mathbf{F}}_{k-1}^{{\tilde l}_{i-1}})^T{\bf{A}}^{{l}_{i}}(:,1:k-1)^T
 {\bf{A}}^{{l}_{i}}(:,1:k).
 \end{multline}
 Finally, we substitute (\ref{Pi2lVVFAA2399sd32d})
 into
(\ref{v2viiVVFAAf930ds2deduce})
to obtain
(\ref{vBarIter2deduce2}).


To deduce (\ref{fkli2Flvvfl932sd}) and
(\ref{fkk2fkkvkk239sd23sd}),
substitute
 (\ref{Fi2ffiiT3902originalF})
 and
(\ref{VmatrixIter2deduce3})
into
(\ref{Lbig2LLwave59056toli})
to obtain
\begin{align}
{{{{\bf{F}}}_{k}^{{\tilde l}_{i}}}} &={\mathbf{F}}_k^{{\tilde l}_{i-1}}{\mathbf{V}}_k^{{\tilde l}_{i}}  \notag \\
&= \left[ {\begin{array}{*{20}{c}}
{{{{\bf{F}}}_{k - 1}^{{\tilde l}_{i-1}}}}& {{{\bf{\bar f}}}_k^{{\tilde l}_{i-1}}} \\
{{\bf{0}}_{k - 1}^T}& {f_{kk}^{{\tilde l}_{i-1}}}
\end{array}} \right] \left[ {\begin{array}{*{20}{c}}
{\mathbf{V}}_{k-1}^{{\tilde l}_{i}} &{{\bf{\bar v}}_k^{{\tilde l}_{i}}}\\
{{\bf{0}}_{k - 1}^T}&{{v_{kk}^{{\tilde l}_{i}}}}
\end{array}} \right]  \notag \\
&= \left[ {\begin{array}{*{20}{c}}
{{{{\bf{F}}}_{k - 1}^{{\tilde l}_{i-1}}}} {\mathbf{V}}_{k-1}^{{\tilde l}_{i}}    &  {{{\bf{F}}_{k - 1}^{{\tilde l}_{i-1}}}{\bf{\bar v}}_k^{{\tilde l}_{i}} + {v_{kk}^{{\tilde l}_{i}}}{{{\bf{\bar f}}}_k^{{\tilde l}_{i-1}}}}  \\
{{\bf{0}}_{k - 1}^T}& {{f_{kk}^{{\tilde l}_{i-1}}}{v_{kk}^{{\tilde l}_{i}}}}
\end{array}} \right],  \label{F2FV2BLK329sd23sd}
\end{align}
which can be compared with (\ref{FtildeIterDeduceEasy})  (i.e.,  (\ref{Fi2ffiiT3902originalF}))
to deduce
(\ref{fkli2Flvvfl932sd})
and
(\ref{fkk2fkkvkk239sd23sd}).
Notice that it is not required to deduce (\ref{FtildeIterDeduceEasy}), since (\ref{FtildeIterDeduceEasy}) is the same as   (\ref{Fi2ffiiT3902originalF}).





\subsection{The Derivation of (\ref{PiXiIter3209sd23sd23})}



%
%


Define
\begin{equation}\label{DefPhi3290d32ds3sd}
{{\bf{\Phi }}_k^{{\tilde l}_{i}}} = ({\bf{A}}^{{l}_{i}} )^T {\bf{A}}^{{l}_{i}}(:,1:k) {\mathbf{F}}_k^{{\tilde l}_{i-1}} {\mathbf{V}}_k^{{\tilde l}_{i}},
 \end{equation}
 which is substituted into
 (\ref{PiTildeDef320ds23})  and
(\ref{XiTildeDef320ds23})
   to obtain
 \begin{equation}\label{PiTildeDef320ds23ToVPhi}
{{\mathbf{\tilde \Pi }}_{k}^{{\tilde l}_{i}}} =  - {\mathbf{V}}_k^{{\tilde l}_{i}}({{\bf{\Phi }}_k^{{\tilde l}_{i}}})^T
 \end{equation}
 and
 \begin{equation}\label{XiTildeDef320ds23ToPhi39ds3}
{{\bf{\tilde \Xi}}_{k}^{{\tilde l}_{i}}} = ({\bf{A}}^{{l}_{i}})^T {\bf{A}}^{{l}_{i}} - {{\bf{\Phi }}_k^{{\tilde l}_{i}}} ({{\bf{\Phi }}_k^{{\tilde l}_{i}}})^T,
 \end{equation}
respectively.
Then write (\ref{F2FV2BLK329sd23sd})  as ${\mathbf{F}}_k^{{\tilde l}_{i-1}}{\mathbf{V}}_k^{{\tilde l}_{i}}=\left[ {\begin{array}{*{20}{c}}
{{{{\bf{F}}}_{k - 1}^{{\tilde l}_{i-1}}}} {\mathbf{V}}_{k-1}^{{\tilde l}_{i}}    &  {{{\bf{F}}_{k - 1}^{{\tilde l}_{i-1}}}{\bf{\bar v}}_k^{{\tilde l}_{i}} + {v_{kk}^{{\tilde l}_{i}}}{{{\bf{\bar f}}}_k^{{\tilde l}_{i-1}}}}  \\
{{\bf{0}}_{k - 1}^T}& {{f_{kk}^{{\tilde l}_{i-1}}}{v_{kk}^{{\tilde l}_{i}}}}
\end{array}} \right]$, which is substituted into (\ref{DefPhi3290d32ds3sd}) to obtain
\begin{small}
\begin{align}
{{\bf{\Phi }}_k^{{\tilde l}_{i}}} &= ({\bf{A}}^{{l}_{i}} )^T {\bf{A}}^{{l}_{i}}(:,1:k) \left[ {\begin{array}{*{20}{c}}
{{{{\bf{F}}}_{k - 1}^{{\tilde l}_{i-1}}}} {\mathbf{V}}_{k-1}^{{\tilde l}_{i}}    &  {{{\bf{F}}_{k - 1}^{{\tilde l}_{i-1}}}{\bf{\bar v}}_k^{{\tilde l}_{i}} + {v_{kk}^{{\tilde l}_{i}}}{{{\bf{\bar f}}}_k^{{\tilde l}_{i-1}}}}  \\
{{\bf{0}}_{k - 1}^T}& {{f_{kk}^{{\tilde l}_{i-1}}}{v_{kk}^{{\tilde l}_{i}}}}
\end{array}} \right]  \notag \\
& = \left[ {\begin{array}{*{20}{c}}
{ ({\bf{A}}^{{l}_{i}})^T {\bf{A}}^{{l}_{i}}(:,1:k-1) {{{{\bf{F}}}_{k - 1}^{{\tilde l}_{i-1}}}} {\mathbf{V}}_{k-1}^{{\tilde l}_{i}}  }&{{\phi _{:k}^{{\tilde l}_{i}}}}
\end{array}} \right]  \notag \\
&= \left[ {\begin{array}{*{20}{c}}
{{\bf{\Phi }}_{k-1}^{{\tilde l}_{i}}} &{{\phi _{:k}^{{\tilde l}_{i}}}}
\end{array}} \right]  \label{PhiIncrease32sdkl230ksdesf}
\end{align}
\end{small}
with
\begin{small}
\begin{align}
{\phi _{:k}^{{\tilde l}_{i}}}&= ({\bf{A}}^{{l}_{i}} )^T {\bf{A}}^{{l}_{i}}(:,1:k){
\left[{{{({{{\bf{F}}_{k - 1}^{{\tilde l}_{i-1}}}{\bf{\bar v}}_k^{{\tilde l}_{i}} + {v_{kk}^{{\tilde l}_{i}}}{{{\bf{\bar f}}}_k^{{\tilde l}_{i-1}}}} )}^T}} \;  {{f_{kk}^{{\tilde l}_{i-1}}}{v_{kk}^{{\tilde l}_{i}}}} \right]^T}  \notag \\
&= ({\bf{A}}^{{l}_{i}} )^T {\bf{A}}^{{l}_{i}}(:,1:k){v_{kk}^{{\tilde l}_{i}}}{\left[({{{\bf{\bar f}}}_k^{{\tilde l}_{i-1}}})^T \;  {f_{kk}^{{\tilde l}_{i-1}}}\right]^T}   \notag \\
&\quad \quad \quad \quad \quad \quad \quad   + ({\bf{A}}^{{l}_{i}} )^T {\bf{A}}^{{l}_{i}}(:,1:k - 1) {{\bf{F}}_{k - 1}^{{\tilde l}_{i-1}}}{\bf{\bar v}}_k^{{\tilde l}_{i}} \notag \\
&= {v_{kk}^{{\tilde l}_{i}}} ({\bf{A}}^{{l}_{i}} )^T  {\bf{A}}^{{l}_{i}}(:,1:k) {{{\bf{f}}}_k^{{\tilde l}_{i-1}}} +  \notag \\
&\quad \quad \quad \quad \quad \quad \quad   + ({\bf{A}}^{{l}_{i}} )^T {\bf{A}}^{{l}_{i}}(:,1:k - 1) {{\bf{F}}_{k - 1}^{{\tilde l}_{i-1}}}{\bf{\bar v}}_k^{{\tilde l}_{i}}. \label{phiCol239sd0d32sd3}
\end{align}
\end{small}

To deduce
(\ref{PiIter2deduce4}) that updates ${{\bf{\tilde \Pi }}_{k}^{{\tilde l}_{i}}}$,
substitute
(\ref{VmatrixIter2deduce3}) and  (\ref{PhiIncrease32sdkl230ksdesf})
into  (\ref{PiTildeDef320ds23ToVPhi}) to obtain
\begin{align}
{{\bf{\tilde \Pi }}_{k}^{{\tilde l}_{i}}} &=  - \left[ {\begin{array}{*{20}{c}}
{\mathbf{V}}_{k-1}^{{\tilde l}_{i}} &{{\bf{\bar v}}_k^{{\tilde l}_{i}}}\\
{{\bf{0}}_{k - 1}^T}&{{v_{kk}^{{\tilde l}_{i}}}}
\end{array}} \right] \left[ {\begin{array}{*{20}{c}}
{{\bf{\Phi }}_{k-1}^{{\tilde l}_{i}}} & {{\phi _{:k}^{{\tilde l}_{i}}}}
\end{array}} \right]_{}^T   \notag \\
&= \left[ {\begin{array}{*{20}{c}}
{ - {\mathbf{V}}_{k-1}^{{\tilde l}_{i}} ({{\bf{\Phi }}_{k-1}^{{\tilde l}_{i}}})^T - {{\bf{\bar v}}_k^{{\tilde l}_{i}}} ({{\phi _{:k}^{{\tilde l}_{i}}}})^T}\\
{ - {{v_{kk}^{{\tilde l}_{i}}}} ({{\phi _{:k}^{{\tilde l}_{i}}}})^T}
\end{array}} \right]  \notag \\
&= \left[ {\begin{array}{*{20}{c}}
{{\bf{\tilde \Pi }}_{k-1}^{{\tilde l}_{i}}} \\
{{\bf{0}}_k^T}
\end{array}} \right] - \left[ {\begin{array}{*{20}{c}}
{{\bf{\bar v}}_k^{{\tilde l}_{i}}}\\
{{v_{kk}^{{\tilde l}_{i}}}}
\end{array}} \right]({{\phi _{:k}^{{\tilde l}_{i}}}})^T.  \label{Pi2Phi238sdc23sd23asds}
\end{align}
To deduce (\ref{XiIter2deduce5})
 that updates ${{\bf{\tilde \Xi}}_{k}^{{\tilde l}_{i}}}$,
substitute  (\ref{PhiIncrease32sdkl230ksdesf}) into
(\ref{XiTildeDef320ds23ToPhi39ds3})
 to obtain
\begin{align}
{{\bf{\tilde \Xi}}_{k}^{{\tilde l}_{i}}} &= ({\bf{A}}^{{l}_{i}} )^T  {\bf{A}}^{{l}_{i}} - \left[ {\begin{array}{*{20}{c}}
{{\bf{\Phi }}_{k-1}^{{\tilde l}_{i}}} & {{\phi _{:k}^{{\tilde l}_{i}}}}
\end{array}} \right]\left[ {\begin{array}{*{20}{c}}
{{\bf{\Phi }}_{k-1}^{{\tilde l}_{i}}} & {{\phi _{:k}^{{\tilde l}_{i}}}}
\end{array}} \right]_{}^T  \notag \\
&= ({\bf{A}}^{{l}_{i}})^T {\bf{A}}^{{l}_{i}} - {{\bf{\Phi }}_{k-1}^{{\tilde l}_{i}}} ({{\bf{\Phi }}_{k-1}^{{\tilde l}_{i}}})^T - {{\phi _{:k}^{{\tilde l}_{i}}}} ({{\phi _{:k}^{{\tilde l}_{i}}}})^T  \notag \\
&= {{\bf{\tilde \Xi}}_{k-1}^{{\tilde l}_{i}}} - {{\phi _{:k}^{{\tilde l}_{i}}}} ({{\phi _{:k}^{{\tilde l}_{i}}}})^T.  \label{XiIter2phi3209sd32ds}
\end{align}
Then it can be seen that we only need to verify
\begin{equation}\label{vTheta2phi3290ssd32e}
{{v_{kk}^{{\tilde l}_{i}}}} {\mathbf{\theta }}_k^{{\tilde l}_{i}} = {{\phi _{:k}^{{\tilde l}_{i}}}},
 \end{equation}
 which is substituted
  into (\ref{Pi2Phi238sdc23sd23asds})
 and
(\ref{XiIter2phi3209sd32ds}) to obtain
(\ref{PiIter2deduce4})
and
(\ref{XiIter2deduce5}), respectively.

In the end of this subsection, let us verify (\ref{vTheta2phi3290ssd32e}).
Substitute (\ref{ThetaDef30ds3}) into
 (\ref{vTheta2phi3290ssd32e}) to write
\begin{equation}\label{vTheta2vXif320sd32}
{{v_{kk}^{{\tilde l}_{i}}}} {\mathbf{\theta }}_k^{{\tilde l}_{i}} = {{v_{kk}^{{\tilde l}_{i}}}} {\bf{\tilde \Xi }}_{k-1}^{{\tilde l}_{i}}(1:k,:)^T {\bf{f}}_k^{{\tilde l}_{i-1}},
 \end{equation}
 where ${\bf{\tilde \Xi }}_{k-1}^{{\tilde l}_{i}}(1:k,:)$ (i.e., the first $k$ rows of  ${\bf{\tilde \Xi }}_{k-1}^{{\tilde l}_{i}}$)
 can be written as
\begin{small}
\begin{multline}\label{XiFirstRows329sd23sd23}
{\bf{\tilde \Xi }}_{k-1}^{{\tilde l}_{i}}(1:k,:) = {{{\bf{A}}^{{l_i}}}(:,1:k)^T}{{\bf{A}}^{{l_i}}} -  {\bf{A}}^{{l}_{i}}(:,1:k)^T {\bf{A}}^{{l}_{i}}(:,1:k - 1)   \\
\times {\mathbf{F}}_{k-1}^{{\tilde l}_{i-1}} {\mathbf{V}}_{k-1}^{{\tilde l}_{i}} ({\mathbf{V}}_{k-1}^{{\tilde l}_{i}})^T  ({\mathbf{F}}_{k-1}^{{\tilde l}_{i-1}})^T{\bf{A}}^{{l}_{i}}{(:,1:k - 1)^T}{{\bf{A}}^{{l}_{i}}}
\end{multline}
\end{small}
by using (\ref{XiTildeDef320ds23}).
Then
substitute (\ref{XiFirstRows329sd23sd23}) into (\ref{vTheta2vXif320sd32})
  to obtain
\begin{footnotesize}
\begin{multline}\label{viiTheta2viiApAp329sd233}
{{v_{kk}^{{\tilde l}_{i}}}} {\mathbf{\theta }}_k^{{\tilde l}_{i}} = {{v_{kk}^{{\tilde l}_{i}}}}
({{\bf{A}}^{{l_i}}})^T{{{\bf{A}}^{{l_i}}}(:,1:k)} {\bf{f}}_k^{{\tilde l}_{i-1}} -  ({{\bf{A}}^{{l_i}}})^T{\bf{A}}^{{l}_{i}}{(:,1:k - 1)} {\mathbf{F}}_{k-1}^{{\tilde l}_{i-1}}  \\
\times {{v_{kk}^{{\tilde l}_{i}}}}
{\mathbf{V}}_{k-1}^{{\tilde l}_{i}}
({\mathbf{V}}_{k-1}^{{\tilde l}_{i}})^T ({\mathbf{F}}_{k-1}^{{\tilde l}_{i-1}})^T
\\ \times {\bf{A}}^{{l}_{i}}(:,1:k - 1)^T  {\bf{A}}^{{l}_{i}}(:,1:k) {\bf{f}}_k^{{\tilde l}_{i-1}}.
\end{multline}
\end{footnotesize}
Finally,
we  substitute (\ref{v2viiVVFAAf930ds2deduce})
  into (\ref{viiTheta2viiApAp329sd233})
to obtain
\begin{footnotesize}
 \begin{displaymath}
{{v_{kk}^{{\tilde l}_{i}}}} {\mathbf{\theta }}_k^{{\tilde l}_{i}} = {{v_{kk}^{{\tilde l}_{i}}}}
({{\bf{A}}^{{l_i}}})^T{{{\bf{A}}^{{l_i}}}(:,1:k)} {\bf{f}}_k^{{\tilde l}_{i-1}} +  ({{\bf{A}}^{{l_i}}})^T{\bf{A}}^{{l}_{i}}{(:,1:k - 1)} {\mathbf{F}}_{k-1}^{{\tilde l}_{i-1}} {{\bf{\bar v}}_k^{{\tilde l}_{i}}},
 \end{displaymath}
 \end{footnotesize}
 into which
substitute
  (\ref{phiCol239sd0d32sd3})
 to obtain  (\ref{vTheta2phi3290ssd32e}).

\ifCLASSOPTIONcaptionsoff
  \newpage
\fi

\end{document}